\documentclass[11pt]{article}
\usepackage{opendrivelab}
\usepackage{pifont}
\usepackage[table]{xcolor}
\usepackage{float}
\usepackage{multirow}
\usepackage{array}
\usepackage{tabularx}
\usepackage{ltablex}
\keepXColumns
\newcolumntype{L}[1]{>{\raggedright\arraybackslash}p{#1}}
\newlength{\orglogoheight}
\newcommand{\redx}{\textcolor{red}{\ding{55}}}
\newcommand{\greencheck}{\textcolor{green!60!black}{\ding{51}}}

\title{Bench2Dex: Benchmarking Visuo-Tactile Bimanual Dexterous Manipulation Across Dexterous Hands}

\orglabel{}
\orglogo{%
  \makebox[\linewidth][l]{%
    \includegraphics[height=1.2cm,keepaspectratio]{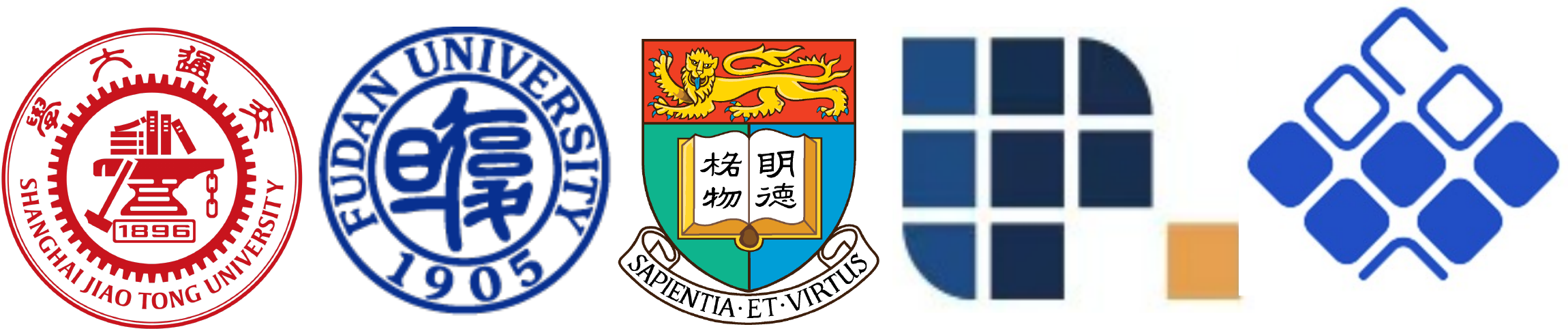}%
  }%
}

\contributors{%
  \small
  Zhenjie Yang\textsuperscript{3*}\quad
  Yideng Zhang\textsuperscript{1*}\quad
  Dongjie Zhang\textsuperscript{2,5*}\quad
  Chenyu Jiang\textsuperscript{2,5*}\quad
  Xianshuai Liu\textsuperscript{1}\quad \\
  Yufeng Li\textsuperscript{1,5}\quad
  Zuhao Ge\textsuperscript{2}\quad 
  Xingyu Jiao\textsuperscript{2,5}\quad
  Zheng Zhang\textsuperscript{1}\quad 
  Kaiyu He\textsuperscript{1}\quad
  He Wang\textsuperscript{1}\quad 
  Yuwen Zhong\textsuperscript{1}\quad \\
  Yi Deng\textsuperscript{1}\quad
  Muyun Jiang\textsuperscript{7}\quad
  Xianliang Huang\textsuperscript{2}\quad
  Haisheng Su\textsuperscript{1}\quad 
  Donghang Zhang\textsuperscript{4}\quad
  Jian Zhang\textsuperscript{4}\quad \\
  Xue Yang\textsuperscript{1,6}\quad 
  Hongyang Li\textsuperscript{3}\quad
  Zuxuan Wu\textsuperscript{2}\quad 
  Yu-Gang Jiang\textsuperscript{2}\quad 
  Xiaosong Jia\textsuperscript{2\dag}\quad
  Junchi Yan\textsuperscript{1\dag}\quad
}

\contriblegend{
$^{1}$ HKU \quad
}

\contriblegend{
    \textsuperscript{1} Shanghai Jiao Tong University \quad
    \textsuperscript{2} Fudan University \quad
    \textsuperscript{3} The University of Hong Kong  \quad \\
    \textsuperscript{4} Inspire Robots \quad
    \textsuperscript{5} Zhongguancun Academy \quad 
    \textsuperscript{6} COWARobot Co. Ltd\quad
    \textsuperscript{7} Nanyang Technological University \quad \\
    \textsuperscript{*}\,Core contribution
    \textsuperscript{\dag}\,Corresponding authors \quad \\
    Contact: \href{mailto:yangzj@hku.hk}{\texttt{yangzj@hku.hk}}, \href{mailto:jiaxiaosong@fudan.edu.cn}{\texttt{jiaxiaosong@fudan.edu.cn}}
}

\projectlinks{%
  \link{Project Page}{https://bench2dex.github.io}\quad
  \link{Full Code}{https://github.com/Bench2Dex/Bench2Dex}\quad
  \link{Huggingface}{https://huggingface.co/Bench2Dex}\quad
  \link{ModelScope}{https://www.modelscope.cn/organization/Bench2Dex}\quad
  \link{Documentation}{  https://bench2dex.github.io/doc/}\quad
  }

\begin{document}
\maketitle

\begin{figure}[htbp]
    \centering
    \includegraphics[width=1\linewidth]{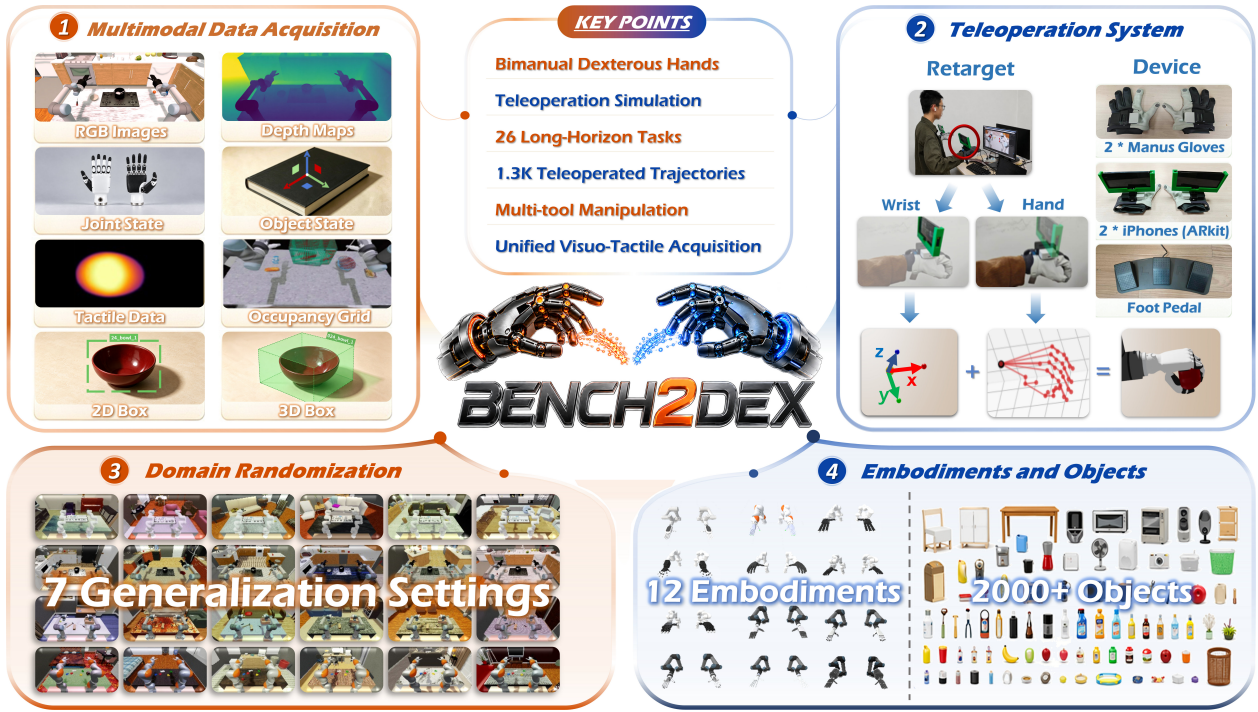}
    \caption{\textbf{Overview of Bench2Dex.} Bench2Dex is a simulation benchmark for bimanual dexterous manipulation with 26 tasks, 12 robot embodiments, and about 1.3K human-teleoperated demonstration trajectories, collected across a range of objects and scenes. It supports teleoperated demonstration collection and 8 data modalities: RGB images, depth maps, joint states, object states, a shared visuo-tactile representation, 2D/3D bounding boxes, and occupancy grids. With this multimodal data and controlled domain randomization, Bench2Dex supports evaluation of policy performance, robustness, and generalization on everyday bimanual manipulation tasks.}
    \label{fig:system_overview}
\end{figure}

\begin{abstract}
Tactile sensing provides contact information that can be difficult to infer from vision alone, but tactile hardware for dexterous hands has not converged to a common design. Dexterous hands differ in finger structure, contact surfaces, and sensor layouts, while simulated tactile signals still differ from measurements produced by physical sensors. These factors make it difficult to study visuo-tactile manipulation across diverse dexterous hands within a consistent experimental setting. We present \textbf{Bench2Dex}, a simulation benchmark for visuo-tactile bimanual manipulation across 12 dexterous hands. We adapt existing robot models with a shared simulated tactile interface that converts local contact geometry into image-like tactile observations. The interface provides a consistent observation format across different hand morphologies without attempting to reproduce the output of a specific physical tactile sensor. Bench2Dex includes 26 bimanual manipulation tasks that involve tool use, articulated-object interaction, and multi-stage manipulation, together with about 1.3K human-teleoperated demonstrations. The benchmark provides synchronized visual, tactile, proprioceptive, action, and object-state observations, together with executable task metrics. For robustness, we group seven perturbation types into invariance axis, where the correct action does not change, and equivariance axis, where the correct action changes together with the perturbation. We evaluate ACT, Diffusion Policy, $\pi_{0.5}$, and GR00T N1.5 on Bench2Dex and report their performance and failure modes. Bench2Dex is meant as a platform for studying visuo-tactile learning across dexterous hands. It does not assume that simulated tactile observations can replace real tactile sensing; it offers a shared setting for algorithm development while tactile hardware and simulation models are still evolving. All code for training, inference, and teleoperation is open-sourced.
\end{abstract}

\section{Introduction}
\label{sec:introduction}
Dexterous manipulation is a core capability for embodied agents: it lets robots interact with tools, articulated objects, and everyday environments through direct physical contact~\cite{dexgraspnet,unidexgrasp,dexart,robocasa,behavior1k}. Driven by large-scale teleoperation datasets~\cite{droid,bridgedatav2,aloha,mobilealoha}, simulation benchmarks~\cite{metaworld,rlbench,calvin,libero,maniskill2,robocasa,triworldbench2026,molmospaces2026}, cross-embodiment data efforts~\cite{openx,robomind}, and vision-language-action policies~\cite{rt1,rt2,octo,openvla}, robot learning has made steady progress in scale, task coverage, and reproducibility.
 
However, most manipulation benchmarks and datasets are still built around one fixed robot platform, one gripper or hand design, or a narrow set of sensing modalities~\cite{metaworld,rlbench,calvin,libero,maniskill2,robocasa,droid,bridgedatav2}. Some large datasets now cover multiple embodiments, but their evaluation protocols are not built to compare dexterous manipulation policies across different arm-hand designs and contact surfaces~\cite{openx,robomind,rh20t}.
 
This gap matters most for visuo-tactile learning. Vision-based tactile sensors output image-like contact signals that depend closely on fingertip shape and sensor layout~\cite{threedvitac,gelsight2017,digit2020,tacto2022,taxim2021}. As a result, tactile signals recorded on one hand do not transfer directly to another hand, and, to our knowledge, no existing benchmark supports visuo-tactile data collection across multiple dexterous hands under one setting~\cite{threedvitac,anyteleop,fromonehand,robotwin2}.
 
A benchmark that spans many embodiments is useful for more than coverage. Different hands have different kinematic limits, finger designs, and contact patterns, and these differences affect grasp strategy, tool use, and long-horizon task execution~\cite{dexgraspnet,unidexgrasp,bidexhands,dexmimicgen,anyteleop,fromonehand}. A policy that works on one hand and fails on another is not necessarily a weaker policy -- it may simply not have been exposed to that hand's shape during training. In the same way, a shared visuo-tactile setting helps check whether a policy makes general use of contact information, rather than fitting to one hand's tactile layout~\cite{threedvitac,gelsight2017,digit2020}.
 
Large-scale robot data~\cite{openx,droid,bridgedatav2,robomind}, dexterous grasping and bimanual manipulation~\cite{dexgraspnet,unidexgrasp,bidexhands,dexmimicgen,DexJoCo}, and teleoperation systems~\cite{aloha,mobilealoha,anyteleop,fromonehand} have each advanced on their own, but largely as separate lines of work. \textbf{To our knowledge, there is still no benchmark that combines diverse bimanual dexterous embodiments, teleoperated simulation, cross-embodiment visuo-tactile data collection, long-horizon tool-use tasks, executable progress evaluation, and generalization testing in one setting}~\cite{bidexhands,dexmimicgen,threedvitac,anyteleop,fromonehand,tactidex,robotwin2,DexJoCo,Dexverse}.
 
To address this gap, we introduce \textbf{Bench2Dex}, a simulation benchmark for teleoperated bimanual dexterous manipulation built on \textbf{Isaac Lab}~\cite{isaaclab,isaacgym}. Bench2Dex covers 12 robot embodiments and 26 long-horizon tasks built around tool use, multi-stage interaction, and daily manipulation scenarios. Human operators collect 1.3K teleoperated trajectories in simulation, and the benchmark records synchronized observations across eight modalities, including visual, geometric, proprioceptive, action, and visuo-tactile signals. Each task has a structured scene description and is scored with executable success and progress conditions. Together, these parts connect teleoperated data collection, multimodal observation, and evaluation within one framework.
 
We group the seven robustness perturbation types into two kinds. Tabletop texture, lighting conditions, scene background, camera pose, and distractor objects change the input but not the task: the object and the goal are unchanged, so the correct action should stay the same, and a drop in success rate reflects sensitivity to nuisance factors rather than a harder task. Object pose and table height change the task geometry: the correct action should change together with the perturbation, so success here instead reflects whether the policy adapts correctly. We refer to the first group as invariance axis and the second as equivariance axis, following how these properties are defined for learned policies more generally. Reporting the two groups separately, rather than as one aggregate robustness score, lets a drop in success rate be traced to nuisance sensitivity or to genuine task generalization, instead of being folded into a single number.
 
Dexterous-hand hardware has not converged on a common finger or sensor design, and the tactile interface in Bench2Dex does not reproduce the output of any specific physical sensor. We do not treat this as a reason to wait. A shared simulation setting lets the community study visuo-tactile perception and cross-embodiment dexterous manipulation under matched tasks and conditions now, and the interface can be revised as tactile hardware and simulation models mature. \textbf{In sum, our main contributions are as follows:}

\begin{table*}[t]
\centering
\caption{Comparison of Bench2Dex with Representative Manipulation Benchmarks.}
\label{tab:benchmark_comparison}
\resizebox{\linewidth}{!}{
\begin{tabular}{l|ccccccc}
\toprule
\textbf{Benchmark}
& \textbf{Task Num}
& \textbf{Embodiment}
& \textbf{Tool/Device Use}
& \textbf{Dexterous Hand}
& \textbf{Teleoperation}
& \textbf{Vision-Based Tactile}
& \textbf{Articulated} \\
\midrule

LIBERO~\cite{libero}
& 130
& 1
& \redx
& \redx
& \greencheck
& \redx
& \greencheck \\

RLBench2~\cite{rlbench2}
& 13
& 1
& \greencheck
& \redx
& \redx
& \redx
& \greencheck \\

RoboCasa~\cite{robocasa}
& 100
& 1
& \greencheck
& \redx
& \greencheck
& \redx
& \greencheck \\

DROID~\cite{droid}
& 86
& 1
& \redx
& \redx
& \greencheck
& \redx
& \redx \\

DexMimicGen~\cite{dexmimicgen}
& 9
& 3
& \redx
& \greencheck
& \greencheck
& \redx
& \greencheck \\

RealMirror~\cite{realmirror}
& 5
& 1
& \redx
& \greencheck
& \greencheck
& \redx
& \greencheck \\

RoboTwin 2.0~\cite{robotwin2}
& 50
& 5
& \greencheck
& \redx
& \redx
& \redx
& \greencheck \\

RoboMIND 2.0~\cite{robomind2}
& 739
& 6
& \redx
& \redx
& \redx
& \greencheck
& \redx \\

MuJoCo Manipulus~\cite{mujocomani}
& 16
& 1
& \greencheck
& \redx
& \redx
& \redx
& \redx \\

RoboCasa365~\cite{robocasa365}
& 365
& 1
& \greencheck
& \redx
& \greencheck
& \redx
& \greencheck \\

BiCoord~\cite{BiCoord}
& 18
& 1
& \redx
& \redx
& \redx
& \redx
& \redx \\

DexJoCo~\cite{DexJoCo}
& 11
& 2
& \greencheck
& \greencheck
& \greencheck
& \redx
& \greencheck \\

DexVerse~\cite{Dexverse}
& 19
& 1
& \greencheck
& \greencheck
& \greencheck
& \redx
& \greencheck \\

\midrule
\textbf{Bench2Dex (Ours)}
& \textbf{26}
& \textbf{12}
& \textbf{\greencheck}
& \textbf{\greencheck}
& \textbf{\greencheck}
& \textbf{\greencheck}
& \textbf{\greencheck} \\

\bottomrule
\end{tabular}}
\end{table*}

\begin{itemize}[noitemsep,topsep=0pt,leftmargin=*,itemsep=2pt]
    \item We introduce \textbf{Bench2Dex}, a simulation benchmark for bimanual dexterous manipulation featuring 12 robot embodiments, 26 long-horizon tasks, and 1.3K human-teleoperated demonstrations. \textbf{We open-source the teleoperation systems for all embodiments.}

    \item We develop a \textbf{\emph{unified visuo-tactile interface}} that maps local contact geometry to a common image-like tactile representation across 12 dexterous hands with diverse finger structures and sensor layouts. Together with vision, proprioception, and action signals, it provides synchronized data across eight modalities.

    \item We establish an evaluation suite covering task completion, stage-level progress, and execution quality, with robustness tests spanning seven perturbation types along two axes: \textbf{\emph{invariance}}, where correct actions remain unchanged, and \textbf{\emph{equivariance}}, where they transform with the perturbation.

    \item We benchmark ACT, Diffusion Policy, $\pi_{0.5}$, and GR00T N1.5 on Bench2Dex, highlighting the gap between matched scenes and controlled scene perturbations in bimanual dexterous manipulation.
\end{itemize}

\section{Related Work}
\label{sec:related_work}
\subsection{Manipulation Benchmarks and Datasets}
\label{subsec:rw_benchmarks}
 
Simulation benchmarks such as Meta-World~\cite{metaworld}, RLBench~\cite{rlbench}, robosuite~\cite{zhu2020robosuite}, CALVIN~\cite{calvin}, LIBERO~\cite{libero}, and ManiSkill3~\cite{tao2024maniskill3} provide standardized evaluation for multi-task manipulation, while MimicGen~\cite{mandlekar2023mimicgen} and RoboTwin~\cite{robotwin} reduce demonstration cost through data generation. Large-scale real-world datasets---BridgeData V2~\cite{bridgedatav2}, Open X-Embodiment~\cite{openx}, DROID~\cite{droid}, RoboMIND 2.0~\cite{robomind2}, and AgiBot World~\cite{bu2025agibotworld}---have fueled generalist policies such as Octo~\cite{octo}, OpenVLA~\cite{openvla}, and $\pi_0$~\cite{black2024pi0}. Recent robustness benchmarks---RoboCasa~\cite{robocasa}, RoboCasa365~\cite{robocasa365}, GemBench~\cite{garcia2024gembench}, THE COLOSSEUM~\cite{pumacay2024colosseum}, RoboTwin 2.0~\cite{robotwin2}, RLBench2~\cite{rlbench2}, and MuJoCo Manipulus~\cite{mujocomani}---test controlled distribution shifts. This line of work relies on parallel-jaw grippers, so it provides no multi-finger dexterous hand or vision-based tactile support, and its evaluation protocols do not compare dexterous hand morphologies or visuo-tactile sensing configurations.
 
\subsection{Dexterous and Bimanual Manipulation Benchmarks}
\label{subsec:rw_dexterous_benchmarks}
 
A separate line of benchmarks targets dexterous or bimanual manipulation specifically, but each covers only part of the properties in Table~\ref{tab:benchmark_comparison}. Bi-DexHands~\cite{bidexhands} focuses on reinforcement learning with dual Shadow Hands but provides no visual observations, and BiCoord~\cite{BiCoord} studies bimanual coordination without dexterous hands. DexMimicGen~\cite{dexmimicgen} generates demonstrations for 3 dexterous embodiments across 9 tasks, but does not include tool use or tactile sensing. RealMirror~\cite{realmirror} supports a single dexterous embodiment across 5 tool-use tasks, without tactile sensing. DexJoCo~\cite{DexJoCo} supports 2 embodiments and 11 tasks with tool use, but omits tactile data. DexVerse~\cite{Dexverse} covers 19 tool-use tasks on a single dexterous embodiment with teleoperated demonstrations, but likewise does not include tactile sensing. None of these benchmarks combines multiple dexterous embodiments with tactile sensing.
 
\subsection{Dexterous Data Collection and Teleoperation}
\label{subsec:rw_dexterous_teleop}
 
Foundational work on Adroit~\cite{rajeswaran2018dexterous}, OpenAI in-hand manipulation~\cite{andrychowicz2020dexterous}, and DexYCB~\cite{chao2021dexycb} established dexterous hands as challenging platforms for robot learning. 
Recent efforts scale dexterous data through synthetic grasps~\cite{zhang2024dexgraspnet2,ye2025dex1b}, generative demonstrations~\cite{dexmimicgen}, hand-motion reconstruction from egocentric videos~\cite{zhu2026mint} and human-to-robot transfer~\cite{zhao2024dexh2r,hoque2025egodex,yang2026handedit}. Teleoperation systems such as Mobile ALOHA~\cite{mobilealoha}, ALOHA 2~\cite{aloha2team2024aloha2}, UMI~\cite{chi2024umi}, and FastUMI~\cite{zhaxizhuoma2024fastumi} advance bimanual data collection but use parallel-jaw grippers, while AnyTeleop~\cite{anyteleop} and From One Hand to Multiple~\cite{fromonehand} address cross-hand retargeting at the algorithmic level. These methods improve how dexterous demonstrations are generated or collected, but each is evaluated on its own custom setup rather than a shared benchmark spanning multiple embodiments.
 
\subsection{Tactile Sensing and Benchmarking for Manipulation}
\label{subsec:rw_tactile}
 
Vision-based tactile sensors, such as GelSight~\cite{yuan2017gelsight,gelsight2017}, DIGIT~\cite{digit2020}, TACTO~\cite{tacto2022}, and Taxim~\cite{taxim2021}, convert local contact deformation into image-like tactile observations, enabling robots to reason about contact states that are difficult to infer from external vision alone. Recent studies have used tactile feedback for insertion, grasp adjustment, and contact-rich manipulation~\cite{manipulation_by_feel,tactile_rl_insertion,letac_mpc}, while newer work further explores reusable tactile skins~\cite{bhirangi2024anyskin}, self-supervised touch representations~\cite{higuera2024sparsh}, visuo-tactile pretraining and policy learning~\cite{m3l_power_senses,vital_pretraining,threedvitac}, and tactile-conditioned diffusion or vision-language-action policies~\cite{reactive_diffusion_policy,tacdiffusion,tla2025,tactile_vla2025,tacvla2026,taf_vla2026}.
 
Complementary benchmark efforts have begun to standardize tactile evaluation at different levels. EgoTactile pairs egocentric video with full-hand pressure supervision, RCT evaluates contact-sequence-aware generalization across materials and sensors, HT-Bench targets full-hand tactile representation learning over 226 tasks, and HRDexDB aligns human and robot grasp sequences for cross-embodiment study~\cite{egotactile2026,rct2026,htbench2026,hrdexdb2026}. At the closed-loop policy level, roto 2.0 evaluates tactile-only reinforcement learning across four dexterous morphologies, TactiDex measures physically grounded contact in real-world dexterous manipulation, and SoftVTBench introduces goal- and safety-aware evaluation for visuo-tactile deformable-object manipulation~\cite{roto2_2026,tactidex,softvtbench2026}. Among the benchmarks in Table~\ref{tab:benchmark_comparison}, RoboMIND 2.0~\cite{robomind2} is the only other one with vision-based tactile support, but it uses parallel-jaw grippers across six embodiments, so its tactile signal is not organized around a shared representation for heterogeneous dexterous hand morphologies. These efforts address complementary slices of tactile learning, but do not jointly target a common tactile representation spanning heterogeneous bimanual dexterous embodiments and long-horizon tasks.
 
Bench2Dex instead focuses on a unified cross-embodiment tactile representation by reconstructing tactile contact surfaces from different hand meshes and converting contact depth into a common surface-aligned tactile-map format, building on geometry-consistent penetration-depth encoding~\cite{tacmap2026}. This enables consistent tactile data acquisition and evaluation across diverse dexterous embodiments.
 
As summarized in Table~\ref{tab:benchmark_comparison}, no existing benchmark jointly supports diverse bimanual dexterous embodiments, teleoperated demonstration collection, vision-based tactile sensing, tool use, and articulated-object interaction. Bench2Dex fills this gap with a single simulation pipeline that unifies multiple bimanual dexterous embodiments under one teleoperation interface, records synchronized multimodal observations including tactile data, and evaluates policies under controlled distribution shifts.

\section{Bench2Dex Benchmark}
\label{sec:benchmark}

Bench2Dex is designed as a full benchmark pipeline rather than a task collection alone. As shown in Figure~\ref{fig:system_overview}, it connects task construction, human teleoperation, multimodal data collection and executable evaluation into a unified simulation framework. This section describes the benchmark components used to generate data and evaluate policies. 

\subsection{Online Teleoperation Setup}
\label{subsec:teleoperation_setup}
Bench2Dex uses human teleoperation to collect demonstrations for long-horizon bimanual dexterous manipulation. The teleoperation system is integrated directly into the Isaac Lab simulation loop, so each episode is recorded together with the commanded action, robot state, object state, camera observations, task metadata, and success or metric signals. The recording pipeline is designed to introduce minimal per-step overhead, preserving the responsiveness required for real-time human teleoperation. Teleoperation therefore serves not only as a data collection tool, but also as a practical feasibility check for whether a task can be executed under a given robot embodiment and scene configuration.

The operator controls the robot through a Manus glove and an ARKit wrist-tracking stream. The Manus runtime provides a 25-node hand skeleton through shared memory; the system converts it to 21 MediaPipe-style hand keypoints and uses DexPilot retargeting to solve target joint angles for the active robot hand. The same retargeting interface is configured for twelve dexterous hand embodiments, giving all supported hands a unified teleoperation and data-collection protocol. Arm motion is controlled separately: ARKit provides wrist translation and orientation cues, while a Pinocchio-based closed-loop inverse-kinematics controller maps the desired wrist pose to the corresponding arm joint targets. The hand and arm targets are then merged into a single absolute joint-position command for the full bimanual robot.

During recording, Bench2Dex stores the action to be executed at the next simulation step, followed by the resulting post-step observations and evaluator state. Before starting or saving a trajectory, the robot is moved to a home configuration, which reduces discontinuities between demonstrations and resets the teleoperation filters and wrist anchors.

\subsection{Offline Multimodal Data Acquisition}
\label{subsec:offline_multimodal}

The lightweight online recording described above captures only the essential motion stream. The remaining modalities---multi-view RGB-D images, object bounding boxes, occupancy labels, and surface-aligned tactile observations---are generated by an offline replay pipeline that reconstructs each episode in simulation and renders the full sensor suite. This decoupling of teleoperation from expensive rendering is a key design choice: the human operator session is kept short, while the replay step can be parallelized, re-run with updated sensor configurations, or selectively applied to a subset of episodes.

\paragraph{Replay pipeline.}
For each recorded episode, the replay pipeline loads the saved object initial states and robot joint trajectory, then replays the simulation with the same random seed to ensure deterministic reproduction. Multi-view RGB-D images are rendered from the six calibrated camera viewpoints (chest, overhead, stereo left, stereo right, left wrist, right wrist). Semantic and instance segmentation labels are extracted from the rendered outputs, occupancy grids are voxelized from the scene geometry, and object bounding boxes in both 2D image coordinates and 3D world coordinates are generated from the projected object meshes. The HDF5 writer keeps camera calibration, action metadata, frame validity, and task metadata together with these observations, making replay, evaluation, and cross-embodiment comparison consistent across tasks and robot hands. Tactile observations are generated during the same replay pass using the surface-aligned ray-casting pipeline described in Section~\ref{subsec:tactile_data_collection}.

\begin{figure}[htbp]
    \centering
    \includegraphics[width=1\linewidth]{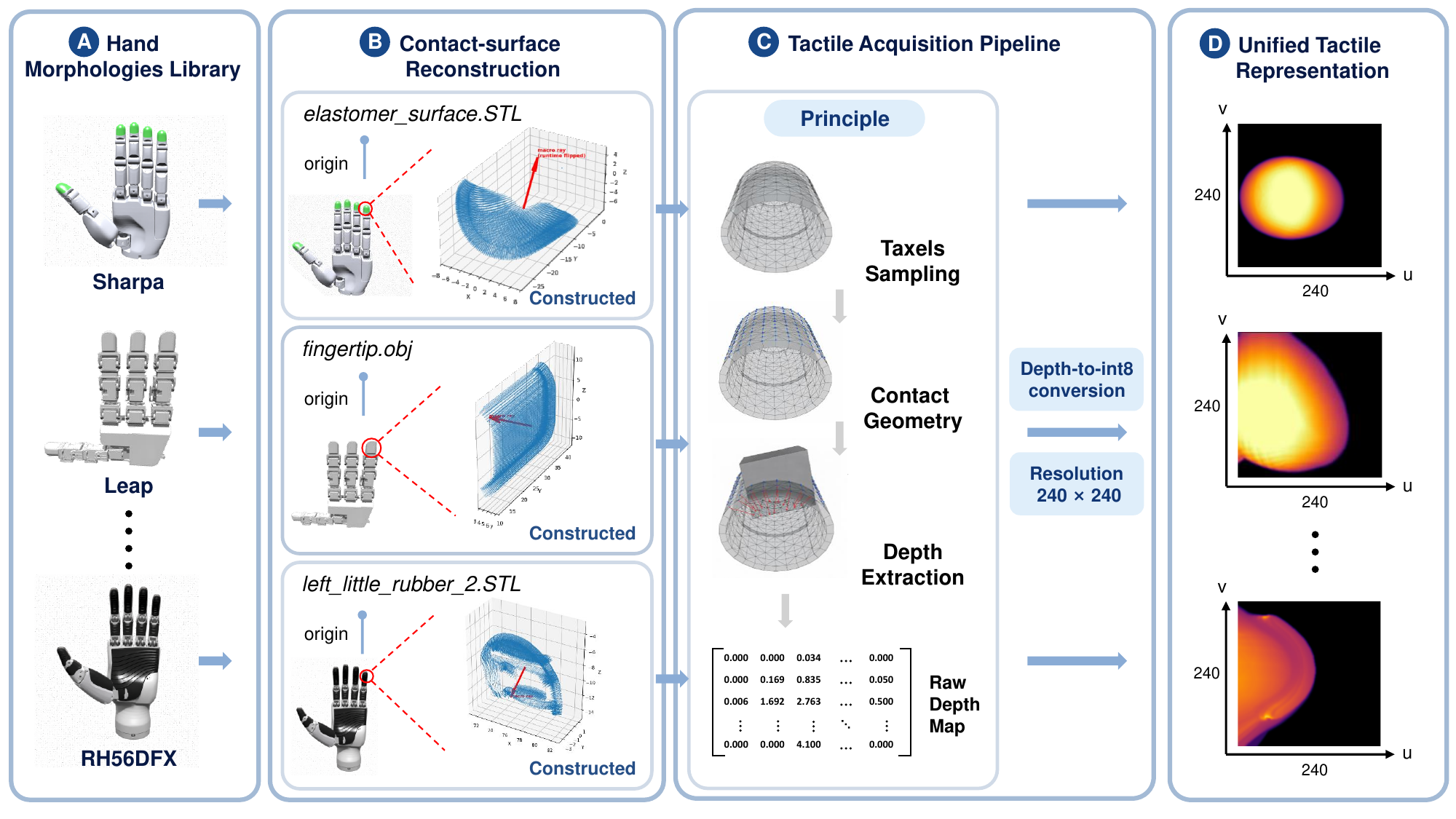}
    \caption{\textbf{Unified surface-aligned tactile acquisition for diverse robotic hands.} Contact surfaces are reconstructed for twelve robot-hand morphologies, and a shared surface-aligned ray-casting pipeline maps the heterogeneous hand geometries into a common image-like tactile representation.}
    \label{fig:tacmap}
\end{figure}

\subsection{Unified Visuo-Tactile Data Acquisition}
\label{subsec:tactile_data_collection}

Contact is central to bimanual dexterous manipulation. After a robot grasps a tool, pushes an articulated part, or stabilizes an object with the other hand, \textbf{\textit{the most informative interaction is often hidden from external cameras.}} Prior systems such as TACTO~\cite{tacto2022} and Taxim~\cite{taxim2021} established image-based tactile simulation, while TacMap~\cite{tacmap2026} introduced a geometry-consistent penetration-depth representation over contact surfaces. Drawing on these paradigms, we develop a unified surface-aligned tactile acquisition layer for Bench2Dex. Our contribution is a cross-embodiment interface that combines reconstructed hand contact surfaces, a shared site registry, consistent encoding, and offline replay to expose heterogeneous robot hands through one tactile data representation.

As illustrated in Figure~\ref{fig:tacmap}, for each tactile site $s$, our acquisition layer uses a precomputed local surface map derived from the hand contact mesh. Each grid cell corresponds to a fixed surface point with an associated inward-facing normal. During replay, rays are cast from these surface points along the inward normals against task-object meshes, so the first-hit depth measures how far an object surface has penetrated past the nominal contact surface; the contact-consistency-filtered depth is then encoded as an 8-bit tactile image $\mathbf{T}_{s,t}\in\{0,\ldots,255\}^{H\times W}$ for site $s$ at time $t$. Bench2Dex also retains the metric ray depth and a binary contact-validity mask, while the compact tactile stream applies a piecewise depth encoding and spatial smoothing. The complete signal semantics and quantization procedure are provided in Appendix~\ref{app:tacmap_quantization}. This representation converts sparse geometric contact into an image-like signal that can be processed by visual encoders or fused with RGB-D, proprioception, and object-state observations.

\textbf{A key design goal is embodiment consistency.} Bench2Dex uses one tactile acquisition paradigm for the twelve bimanual robot-hand embodiments included in the benchmark, while preserving the local contact geometry of each hand. To achieve this, we rebuilt the tactile contact-surface meshes for each benchmark hand and converted them into surface point-and-normal assets, provided in Appendix~\ref{app:tacmap_surface_reconstruction}. A tactile registry then maps each robot embodiment to its tactile site names, surface assets, resolution, normal convention, and site-body attachment rules, avoiding hand-specific branching in the data collection code.

\subsection{Datasets and Policy}
\label{sec:datasets and policy}

\noindent\hspace*{2em}\textbf{Dataset.} Bench2Dex contains about 1.3K human-teleoperated demonstrations across 26 long-horizon tasks and 12 bimanual dexterous embodiments. Each trajectory is replayed into a unified HDF5 record with synchronized RGB-D observations, joint states and actions, object states, surface-aligned tactile maps, 2D/3D bounding boxes, and occupancy grids.

\noindent\hspace*{2em}\textbf{Policy.} We evaluate ACT~\cite{aloha}, Diffusion Policy (DP)~\cite{dp}, $\pi_{0.5}$~\cite{pi05}, and GR00T N1.5~\cite{gr00t}. ACT and DP are trained from scratch using multi-view RGB and joint-state observations, whereas $\pi_{0.5}$ and GR00T N1.5 are fully fine-tuned from pretrained checkpoints. For the pretrained policies, state/action projections are adapted to each embodiment where required, with shape-incompatible or newly added parameters initialized randomly. Complete training configurations are provided in Appendix~\ref{app:policy_training}.

\subsection{Evaluation Metrics}
\label{sec:metric}

Bench2Dex uses a unified evaluation suite that separates primary benchmark scores from task-specific diagnostics. Unless stated otherwise, we use a reach-and-stop protocol: each task provides an executable terminal predicate, and an episode terminates successfully only after that predicate remains true for its configured dwell time (0.5~s by default). Rollouts otherwise terminate at the evaluation budget or when evaluation cannot proceed. The protocol, step budget, dwell time, resolved seed, and sampled generalization parameters are retained for each episode.

The primary completion metric is stable success rate,
\begin{equation}
    \mathrm{SR}=\frac{1}{N}\sum_{e=1}^{N} S_e,
\end{equation}
where $S_e=1$ only when the terminal predicate stably holds, rather than at a transient frame. To characterize partial progress on long-horizon tasks, the benchmark also provides the \emph{latched stage completion rate} (LSCR),
\begin{equation}
    \mathrm{LSCR}_e =
    \frac{1}{K_e}
    \sum_{k=1}^{K_e}
    L_{e,k},
\end{equation}
where $K_e$ is the number of stages and $L_{e,k}=1$ if stage $k$ is reached at any time while its declared dependencies are satisfied by previously latched or concurrently reached stages. Latching prevents progress from being erased when an early predicate is intentionally reversed by a later action, such as closing door after opening it.

For efficiency analysis, the benchmark provides mean time to stable success over successful episodes, together with SR so that this conditional quantity is not interpreted independently of completion. Supported safety diagnostics include safe success rate, hard-violation rate, drop rate, tracked-object high-speed violation rate, and the fraction of executed steps containing a task-level violation. A hard violation is a configured drop or high-speed event; high speed is a severe-motion proxy rather than a contact-force or collision measurement, while joint-limit observations remain auxiliary diagnostics. For generalization analysis, episodes are grouped into the \texttt{None}, \texttt{Equi.}, \texttt{Inv.}, and \texttt{Full} channels, for which the benchmark can compute SR, sample counts, and, when the baseline SR is nonzero, the success-rate ratio relative to \texttt{None}. Additional completion, progress, efficiency, safety, grasp, tool-use, motion, and reproducibility diagnostics are described in Appendix~\ref{app:evaluation_metrics}.

\subsection{Generalization Strategy}
\label{subsec:generalization_strategy}

Bench2Dex evaluates whether a policy learns task-level manipulation concepts rather than memorizing a fixed simulator instance. Each evaluation keeps the semantic goal, object set, and success conditions unchanged while controlling two groups of scene factors. Scene background, tabletop texture, lighting conditions, distractor objects, and camera pose are treated as \textbf{Invariance} factors because they alter task-irrelevant visual conditions without changing the intended task behavior. Object pose and table height are treated as \textbf{Equivariance} factors because they alter task-relevant geometry and therefore require corresponding changes in reaching, grasping, and contact trajectories.

The protocol defines four evaluation channels. \textbf{No generalization (\texttt{None})} exactly restores a matched anchor episode and serves as the baseline. \textbf{Equivariance-only generalization (\texttt{Equi.})} resamples only equivariance factors while preserving the anchor's invariant context, testing geometric adaptation. \textbf{Invariance-only generalization (\texttt{Inv.})} resamples only invariance factors while preserving the anchor's task geometry, testing robustness to nuisance variation. \textbf{Combined generalization (\texttt{Full})} independently resamples both groups without an anchor, testing whether a policy can simultaneously ignore contextual shifts and adapt to new task geometry.

For each episode $e$, Bench2Dex samples a generalization configuration $\mathbf{g}_e$ and instantiates the executable scene as
\begin{equation}
    \mathcal{S}_e = \operatorname{Build}(\mathcal{T}, \mathbf{g}_e),
\end{equation}
where $\mathcal{T}$ denotes the task specification and $\mathbf{g}_e$ stores the resolved parameters for scene background, tabletop texture, lighting conditions, distractor objects, object pose, table height, and camera pose. The resolved configuration is saved with the episode metadata, making every channel replayable and auditable. Appendix~\ref{app:generalization_config} specifies the sampling ranges and channel composition.

\section{Experiments}
\label{sec:experiments}

\subsection{Experimental Setup}
\label{subsec:experimental_setup}

We evaluate ACT, an image-conditioned 1D U-Net Diffusion Policy
variant (DP), $\pi_{0.5}$, and GR00T N1.5 on all 26 Bench2Dex
task--embodiment settings spanning 12 embodiments
(Appendix~\ref{app:task_catalog}). Each policy is evaluated with 50 rollouts
under four generalization channels, yielding
$26\times4\times4\times50=20{,}800$ evaluation episodes. Evaluation uses deterministic task-partitioned episode seeds and a task-specific horizon set to $1.5\times$ the recorded expert horizon. Channel semantics follow Section~\ref{subsec:generalization_strategy}: \texttt{None}, \texttt{Equi.}, and \texttt{Inv.} are aligned by episode index to matched anchors, whereas \texttt{Full} is sampled independently without an anchor. Cross-channel comparisons below therefore use marginal success counts and do not assume a per-instance difficulty ordering.

The primary outcome is reach-and-stop stable success
(Section~\ref{sec:metric}), and each table entry is a success count out of 50
rollouts. For a fixed policy and channel, we report the equal-weight task-macro
success rate over the 26 settings. Since every setting contributes 50 episodes,
this rate is numerically identical to the pooled success rate over 1,300
episodes. The \textbf{SR} column is the equal-weight mean of the four
channel-specific success rates and introduces no additional evaluation samples.

\subsection{Main Results}
\label{subsec:evaluation_results}

\begin{table*}[t]
  \centering
  \caption{\textbf{Stable-success counts across generalization channels on all
26 Bench2Dex task--embodiment settings spanning 12 embodiments.} Each
task entry is the number of reach-and-stop successes among 50 rollouts.
None, Equi., Inv., and Full correspond to the \texttt{none},
\texttt{equi\_only}, \texttt{inv\_only}, and \texttt{inv\_equi} scene profiles,
respectively. \textbf{SR} and \textbf{LSCR} are the four-channel mean stable
success rate and latched stage completion rate, respectively, for each
task--policy pair. The bottom row reports the equal-weight
task-macro SR (\%) and the all-task mean LSCR.}
  \label{tab:policy_generalization_results}
  \begingroup
  \setlength{\tabcolsep}{3.0pt}
  \renewcommand{\arraystretch}{1.08}
  \scriptsize
  \resizebox{\textwidth}{!}{%
    \begin{tabular}{@{}ll*{6}{c}@{\hspace{2.5pt}\vrule width 0.35pt\hspace{2.5pt}}*{6}{c}@{\hspace{2.5pt}\vrule width 0.35pt\hspace{2.5pt}}*{6}{c}@{\hspace{2.5pt}\vrule width 0.35pt\hspace{2.5pt}}*{6}{c}@{}}
      \toprule
      \multicolumn{1}{c}{\multirow{2}{*}{\textbf{Task}}}
      & \multicolumn{1}{c}{\multirow{2}{*}{\textbf{Embodiment}}}
      & \multicolumn{6}{c}{\textbf{ACT}}
      & \multicolumn{6}{c}{\textbf{DP}}
      & \multicolumn{6}{c}{\boldmath$\pi_{0.5}$}
      & \multicolumn{6}{c}{\textbf{GR00T N1.5}} \\
      \cmidrule(lr){3-8}
      \cmidrule(lr){9-14}
      \cmidrule(lr){15-20}
      \cmidrule(lr){21-26}
      & & \textbf{None} & \textbf{Equi.} & \textbf{Inv.} & \textbf{Full} & \textbf{SR} & \textbf{LSCR}
      & \textbf{None} & \textbf{Equi.} & \textbf{Inv.} & \textbf{Full} & \textbf{SR} & \textbf{LSCR}
      & \textbf{None} & \textbf{Equi.} & \textbf{Inv.} & \textbf{Full} & \textbf{SR} & \textbf{LSCR}
      & \textbf{None} & \textbf{Equi.} & \textbf{Inv.} & \textbf{Full} & \textbf{SR} & \textbf{LSCR} \\
      \midrule

      Baking Tray Prep & \multirow[c]{3}{*}{IIWA7+Sharpa}
      & 9 & 2 & 1 & 0 & 6.0\% & 11.6\%
      & 2 & 2 & 0 & 0 & 2.0\% & 7.9\%
      & 7 & 1 & 3 & 4 & 7.5\% & 11.0\%
      & 12 & 7 & 1 & 0 & \textbf{10.0\%} & \textbf{29.3\%} \\

      Canned Food Tray Arrangement &
      & 17 & 2 & 0 & 0 & 9.5\% & 34.6\%
      & 0 & 0 & 0 & 0 & 0.0\% & 14.8\%
      & 7 & 4 & 6 & 7 & 12.0\% & 33.9\%
      & 17 & 5 & 3 & 0 & \textbf{12.5\%} & \textbf{37.3\%} \\

      Jigsaw Puzzle Assembly &
      & 0 & 0 & 0 & 0 & 0.0\% & 12.5\%
      & 0 & 0 & 0 & 0 & 0.0\% & 6.3\%
      & 0 & 0 & 0 & 0 & 0.0\% & 7.4\%
      & 10 & 0 & 0 & 0 & \textbf{5.0\%} & \textbf{30.3\%} \\

      \cmidrule(lr){1-26}

      Gaming Desk Setup & JAKA ZU7+DexHand021
      & 25 & 10 & 17 & 10 & 31.0\% & 46.0\%
      & 7 & 6 & 4 & 1 & 9.0\% & 30.7\%
      & 10 & 5 & 6 & 6 & 13.5\% & 32.3\%
      & 41 & 17 & 34 & 27 & \textbf{59.5\%} & \textbf{72.2\%} \\

      \cmidrule(lr){1-26}

      Medicine Shoebox Pack & \multirow[c]{2}{*}{Panda+Orca}
      & 5 & 1 & 5 & 4 & 7.5\% & 21.5\%
      & 0 & 0 & 0 & 0 & 0.0\% & 9.3\%
      & 3 & 0 & 3 & 3 & 4.5\% & 21.8\%
      & 18 & 3 & 3 & 2 & \textbf{13.0\%} & \textbf{26.2\%} \\

      Tool Box Loading &
      & 17 & 10 & 8 & 5 & 20.0\% & 37.5\%
      & 14 & 6 & 7 & 6 & 16.5\% & 39.8\%
      & 29 & 18 & 20 & 23 & \textbf{45.0\%} & \textbf{60.6\%}
      & 27 & 15 & 21 & 17 & 40.0\% & 51.0\% \\

      \cmidrule(lr){1-26}

      Sports Ball Cup Sort & Panda+Allegro
      & 4 & 0 & 0 & 0 & 2.0\% & 2.5\%
      & 1 & 0 & 0 & 0 & 0.5\% & 4.4\%
      & 4 & 3 & 2 & 4 & 6.5\% & 20.2\%
      & 15 & 0 & 2 & 1 & \textbf{9.0\%} & \textbf{21.5\%} \\

      \cmidrule(lr){1-26}

      Faucet Cup Water Fill & \multirow[c]{2}{*}{RM65+Revo2}
      & 6 & 5 & 2 & 2 & 7.5\% & 35.4\%
      & 3 & 2 & 1 & 2 & 4.0\% & 14.0\%
      & 5 & 1 & 3 & 5 & 7.0\% & 23.8\%
      & 24 & 11 & 18 & 13 & \textbf{33.0\%} & \textbf{55.0\%} \\

      Stationery Category Sorting &
      & 3 & 0 & 0 & 0 & 1.5\% & \textbf{23.1\%}
      & 3 & 0 & 1 & 0 & 2.0\% & 18.5\%
      & 1 & 1 & 1 & 1 & 2.0\% & 8.1\%
      & 5 & 1 & 0 & 0 & \textbf{3.0\%} & 16.9\% \\

      \cmidrule(lr){1-26}

      Bimanual Piano Melody & \multirow[c]{2}{*}{xArm7+Ability}
      & 6 & 0 & 0 & 2 & 4.0\% & 1.0\%
      & 1 & 0 & 0 & 0 & 0.5\% & 0.5\%
      & 6 & 1 & 1 & 5 & 6.5\% & 6.5\%
      & 30 & 8 & 2 & 6 & \textbf{23.0\%} & \textbf{23.5\%} \\

      Wine Glass Plate Balance &
      & 15 & 11 & 13 & 13 & 26.0\% & 72.0\%
      & 3 & 2 & 1 & 1 & 3.5\% & 39.8\%
      & 11 & 7 & 9 & 9 & 18.0\% & 59.2\%
      & 19 & 9 & 16 & 13 & \textbf{28.5\% }& \textbf{72.5\%} \\

      \cmidrule(lr){1-26}

      Cleaner Box Loading & \multirow[c]{3}{*}{xArm7+LEAP}
      & 34 & 20 & 7 & 5 & 33.0\% & 76.2\%
      & 11 & 3 & 2 & 2 & 9.0\% & 32.8\%
      & 30 & 13 & 29 & 24 & \textbf{48.0\%} & 73.8\%
      & 35 & 18 & 24 & 17 & 47.0\% & \textbf{82.3\%} \\

      Shoebox Accessory Pack &
      & 19 & 18 & 11 & 12 & 30.0\% & 56.3\%
      & 6 & 5 & 5 & 3 & 9.5\% & 35.2\%
      & 23 & 13 & 23 & 21 & 40.0\% & 55.5\%
      & 32 & 15 & 18 & 16 & \textbf{40.5\%} & \textbf{60.3\%} \\

      Toilet Lid Cleaner Pour &
      & 16 & 19 & 20 & 13 & 34.0\% & 35.5\%
      & 7 & 5 & 2 & 3 & 8.5\% & 23.5\%
      & 26 & 19 & 15 & 18 & 39.0\% & 47.0\%
      & 32 & 12 & 25 & 21 & \textbf{45.0\%} & \textbf{63.3\%} \\

      \cmidrule(lr){1-26}

      Breadbasket Fast-Food Loading & \multirow[c]{3}{*}{UR5+RH5DG2}
      & 11 & 6 & 3 & 1 & 10.5\% & \textbf{15.8\%}
      & 5 & 1 & 2 & 1 & 4.5\% & 7.3\%
      & 14 & 4 & 5 & 6 & \textbf{14.5\%} & 15.6\%
      & 15 & 2 & 1 & 0 & 9.0\% & 13.5\% \\

      Citrus Plate Loading &
      & 43 & 20 & 14 & 26 & 51.5\% & 69.9\%
      & 23 & 7 & 9 & 6 & 22.5\% & 44.4\%
      & 6 & 2 & 5 & 4 & 8.5\% & 18.9\%
      & 48 & 22 & 24 & 24 & \textbf{59.0\%} & \textbf{73.1\%} \\

      Fridge Wine Interhand Pour &
      & 8 & 4 & 4 & 6 & 11.0\% & 26.4\%
      & 4 & 0 & 1 & 0 & 2.5\% & 41.8\%
      & 9 & 6 & 9 & 6 & 15.0\% & 54.3\%
      & 14 & 6 & 14 & 8 & \textbf{21.0\%} & \textbf{58.6\%} \\

      \cmidrule(lr){1-26}

      Fruit Bowl Loading & \multirow[c]{3}{*}{UR5+RH56DFX}
      & 34 & 15 & 13 & 12 & 37.0\% & 62.0\%
      & 15 & 6 & 9 & 6 & 18.0\% & 47.8\%
      & 32 & 9 & 26 & 20 & \textbf{43.5\%} & \textbf{71.4\%}
      & 41 & 8 & 11 & 5 & 32.5\% & 68.1\% \\

      Screwdriver Box \& Hammer &
      & 16 & 5 & 18 & 10 & 24.5\% & 56.2\%
      & 18 & 5 & 9 & 9 & 20.5\% & 49.0\%
      & 23 & 12 & 17 & 12 & 32.0\% & 53.5\%
      & 35 & 18 & 22 & 21 & \textbf{48.0\%} & \textbf{73.7\%} \\

      Trash Disposal &
      & 15 & 6 & 9 & 7 & 18.5\% & 30.3\%
      & 10 & 8 & 4 & 6 & 14.0\% & 21.5\%
      & 31 & 14 & 21 & 27 & \textbf{46.5\%} & \textbf{63.7\%}
      & 26 & 10 & 19 & 14 & 34.5\% & 47.8\% \\

      \cmidrule(lr){1-26}

      Fridge Fruit Shelf Sorting & \multirow[c]{2}{*}{UR5+Shadow}
      & 1 & 0 & 0 & 0 & 0.5\% & 2.6\%
      & 0 & 0 & 0 & 0 & 0.0\% & 15.1\%
      & 0 & 0 & 0 & 0 & 0.0\% & \textbf{21.8\%}
      & 2 & 0 & 0 & 0 & \textbf{1.0\%} & 19.6\% \\

      Soup Serving &
      & 2 & 0 & 0 & 0 & 1.0\% & 11.5\%
      & 0 & 0 & 0 & 0 & 0.0\% & 3.1\%
      & 0 & 0 & 0 & 0 & 0.0\% & 10.0\%
      & 5 & 1 & 0 & 0 & \textbf{3.0\%} & \textbf{13.9\%} \\

      \cmidrule(lr){1-26}

      Frypan Stand \& Pour & \multirow[c]{2}{*}{UR5+Schunk}
      & 20 & 8 & 3 & 7 & 19.0\% & 51.2\%
      & 16 & 9 & 4 & 0 & 14.5\% & 30.0\%
      & 31 & 19 & 16 & 14 & 40.0\% & 60.0\%
      & 39 & 22 & 30 & 18 & \textbf{54.5\%} & \textbf{72.3\%} \\

      Microwave Bowl Loading &
      & 38 & 29 & 29 & 29 & 62.5\% & 37.4\%
      & 17 & 15 & 6 & 4 & 21.0\% & 48.6\%
      & 27 & 14 & 27 & 21 & 44.5\% & 61.1\%
      & 45 & 27 & 36 & 33 & \textbf{70.5\%} & \textbf{82.0\%} \\

      \cmidrule(lr){1-26}

      Ball Box Loading & \multirow[c]{2}{*}{UR5+Wuji}
      & 7 & 6 & 2 & 2 & 8.5\% & 10.4\%
      & 0 & 0 & 0 & 0 & 0.0\% & 7.1\%
      & 8 & 3 & 14 & 5 & 15.0\% & 19.5\%
      & 31 & 3 & 11 & 1 & \textbf{23.0\%} & \textbf{37.1\%} \\

      Condiment Box Loading &
      & 12 & 3 & 3 & 3 & 10.5\% & 31.8\%
      & 2 & 0 & 0 & 0 & 1.0\% & 4.2\%
      & 12 & 2 & 12 & 11 & \textbf{18.5\%} & \textbf{27.7\%}
      & 13 & 4 & 3 & 1 & 10.5\% & 27.7\% \\

      \midrule
      \multicolumn{2}{c}{\textbf{Task mean (\%)}}
      & 29.5\% & 15.4\% & 14.0\% & 13.0\% & 18.0\% & 33.5\%
      & 12.9\% & 6.3\% & 5.2\% & 3.8\% & 7.1\% & 23.0\%
      & 27.3\% & 13.2\% & 21.0\% & 19.7\% & 20.3\% & 36.1\%
      & 48.5\% & 18.8\% & 26.0\% & 19.8\% & \textbf{28.3\%} & \textbf{47.3\%} \\
      \bottomrule
    \end{tabular}%
  }
  \endgroup
\end{table*}

Under the matched \texttt{None} condition, GR00T achieves 631 successes
among 1,300 rollouts (48.5\%), followed by ACT with 383 (29.5\%),
$\pi_{0.5}$ with 355 (27.3\%), and DP with 168 (12.9\%). GR00T is the
strict observed task-level leader on 23 of the 26 settings, while $\pi_{0.5}$ leads
on two and ACT ties GR00T on the remaining setting. Its matched-condition
advantage is therefore broad across the evaluated tasks.

The ordering becomes less concentrated under the combined \texttt{Full} shift. GR00T and $\pi_{0.5}$ are nearly tied in aggregate success, with 258 (19.8\%) and 256 (19.7\%) successes, respectively, followed by ACT with 169 (13.0\%) and DP with 50 (3.8\%). Despite GR00T's two‑success (0.2 percentage‑point) aggregate advantage, $\pi_{0.5}$ has the largest number of strict observed task‑level leads: 12, compared with eight for GR00T and two for ACT, with four ties. Aggregate success and the distribution of task-level
leaders therefore provide complementary views of performance under the
combined shift. Averaged equally over the four channels, GR00T attains
28.3\% success, followed by $\pi_{0.5}$ at 20.3\%, ACT at 18.0\%,
and DP at 7.1\%.

The LSCR results sharpen this comparison. GR00T has the highest all-task mean
LSCR at 47.3\%, followed by $\pi_{0.5}$ at 36.1\%, ACT at 33.5\%, and DP at
23.0\%, preserving their ordering by mean SR. Relative to their mean SRs, these
LSCR values are higher by 19.0, 15.8, 15.5, and 15.9 percentage points,
respectively. These gaps show that LSCR records dependency-valid stage-level progress not represented by
binary stable success. This distinction is visible near the SR floor: on Jigsaw Puzzle
Assembly, DP and $\pi_{0.5}$ both have 0.0\% mean SR but nonzero LSCR (6.3\%
and 7.4\%), while GR00T reaches 30.3\% LSCR with 5.0\% mean SR. LSCR therefore
distinguishes observed dependency-valid stage progress among policy--task cases
that binary SR places at or near the floor.

\subsection{Generalization Patterns}
\label{subsec:generalization_results}

In the observed policy aggregates, \texttt{None} yields the highest SR for all
four policies: 29.5\% for ACT, 12.9\% for DP, 27.3\% for $\pi_{0.5}$, and
48.5\% for GR00T. Under \texttt{Full}, the corresponding rates are 13.0\%,
3.8\%, 19.7\%, and 19.8\%. These are descriptive marginal differences
between the sampled \texttt{None} and \texttt{Full} distributions.

These values summarize marginal performance under the sampled channel
distributions rather than a guaranteed difficulty ordering. \texttt{None}
restores a matched scene, each isolated channel varies one factor group while
retaining the other from its anchor, and \texttt{Full} independently varies
both groups. Because perturbation realizations vary in magnitude and composition, a \texttt{Full} episode is not paired with an episode from either isolated channel, nor can it be assumed to be harder on a per-instance basis. Channel differences therefore characterize empirical sensitivity to the sampled distributions;
they do not establish per-instance monotonicity, additive penalties, or
interactions between the two factor groups.

The realized four-channel ordering is policy dependent. For ACT and DP,
aggregate SR decreases from \texttt{None} through \texttt{Equi.} and
\texttt{Inv.} to \texttt{Full}. For $\pi_{0.5}$ and GR00T, \texttt{None}
remains highest and \texttt{Inv.} is second, but \texttt{Full} exceeds
\texttt{Equi.}. Thus, combined randomization does not force \texttt{Full} to be the lowest empirical aggregate, and relative sensitivity to the isolated shifts is policy dependent in the realized evaluation.

Using \texttt{None} as the matched-scene reference, the observed marginal
Full-to-None success-rate ratios are 44.1\% for ACT, 29.8\% for DP, 72.1\%
for $\pi_{0.5}$, and 40.9\% for GR00T. GR00T has the highest absolute success
under both \texttt{None} and \texttt{Full}, whereas $\pi_{0.5}$ has the
largest ratio. Although a lower \texttt{Full} count is not enforced by the
sampling design, it occurs in 89 of the 104 task--policy comparisons; the
remaining 15 are tied in this realized sample. Broken down by policy,
\texttt{Full} is lower on 25 and equal on one ACT setting, lower on 20 and
equal on six DP settings, lower on 18 and equal on eight $\pi_{0.5}$ settings,
and lower on all 26 GR00T settings.

The isolated axes reveal policy-dependent sensitivity. ACT and DP have
higher aggregate success under \texttt{Equi.} than under \texttt{Inv.}
(15.4\% versus 14.0\%, and 6.3\% versus 5.2\%), whereas
$\pi_{0.5}$ and GR00T show the reverse ordering (13.2\% versus 21.0\%,
and 18.8\% versus 26.0\%). The numbers of settings on which
\texttt{Inv.} is higher than, equal to, or lower than \texttt{Equi.} are
$6/9/11$ for ACT, $7/10/9$ for DP, $18/5/3$ for $\pi_{0.5}$, and
$16/3/7$ for GR00T. Separating the two axes therefore exposes differences
that would be obscured by a single aggregate robustness score.

\subsection{Task-Level Heterogeneity}
\label{subsec:task_failure_analysis}

\texttt{Full}-condition outcomes separate the 26 settings into 12 on which all four policies record nonzero success, three on which every policy records zero success, and 11 with mixed policy outcomes.  Microwave Bowl Loading has the
largest across-policy \texttt{Full} total, with $(29,4,21,33)$ successes for
ACT, DP, $\pi_{0.5}$, and GR00T, respectively. In contrast, Jigsaw Puzzle
Assembly, Fridge Fruit Shelf Sorting, and Soup Serving yield zero
\texttt{Full} successes for every policy. The mixed group exposes
policy-specific strengths: ACT leads Citrus Plate Loading with 26 successes,
$\pi_{0.5}$ leads Tool Box Loading with 23, and GR00T leads Gaming Desk Setup
with 27. On the two non-jigsaw IIWA7+Sharpa settings, only $\pi_{0.5}$ records nonzero \texttt{Full}-condition success. These contrasts show that aggregate robustness reflects heterogeneous task--policy outcomes, not a uniform ordering.

These results characterize stable success for the evaluated settings. The 50
rollouts per cell are evaluation episodes rather than independent retraining
replicates, so the table does not estimate between-training variability. Since
tasks and embodiments are not factorially crossed, task-level contrasts do not
isolate embodiment effects. Stable SR alone does not distinguish partial progress, safety, or execution time.

\section{Conclusion}
\label{sec:conclusion}

We introduced \textbf{Bench2Dex}, a simulation benchmark for bimanual dexterous manipulation spanning 12 robot embodiments, 26 long-horizon tasks, and approximately 1.3K human-teleoperated demonstrations. Bench2Dex combines teleoperation, synchronized multimodal data acquisition, a shared contact-geometry-based tactile interface, and
executable evaluation of task completion and stage-level progress. Evaluations of ACT, Diffusion Policy, $\pi_{0.5}$, and GR00T N1.5 using RGB and proprioceptive observations show lower aggregate success under combined scene perturbations than under matched scenes, with outcomes varying across tasks and policies. These results provide reference points for the evaluated training and execution configurations. By bringing data collection and evaluation into a common framework, Bench2Dex provides reusable components for studying bimanual dexterous manipulation and extending evaluation to tactile-conditioned policies and transfer across dexterous hands.

\textbf{Limitation and Outlook.} Dexterous-hand and tactile-sensor designs continue to evolve, and the hand and tactile models in Bench2Dex have not been calibrated
against matching physical hardware. Bench2Dex focuses on a shared simulation interface for studying visuo-tactile manipulation across diverse dexterous hands, rather than reproducing a specific physical hand--sensor system. It provides a common setting for algorithm development under explicit modeling and sensing assumptions. As hardware and simulation models advance, future extensions could incorporate refined hand and tactile models, hardware-specific calibration, and physical validation to assess which findings carry over to real-world manipulation.

% \section*{Acknowledgments}

\clearpage
\bibliographystyle{unsrtnat}
\bibliography{bibliography_short,references}

\clearpage
\bookmarksetup{startatroot}
\phantomsection
\pdfbookmark[1]{Supplementary Materials}{supplementary-materials}
{\noindent\Large\itshape Supplementary Materials\par}

\setcounter{section}{0}
\setcounter{subsection}{0}
\setcounter{figure}{0}
\setcounter{table}{0}

\renewcommand{\thesection}{\Alph{section}}
\renewcommand{\thesubsection}{\Alph{section}.\arabic{subsection}}
\renewcommand{\thefigure}{S\arabic{figure}}
\renewcommand{\thetable}{S\arabic{table}}

% Very important: unique hyperref anchors for Supplement
\renewcommand{\theHsection}{supp.\Alph{section}}
\renewcommand{\theHsubsection}{supp.\Alph{section}.\arabic{subsection}}
\renewcommand{\theHfigure}{suppfig.\arabic{figure}}
\renewcommand{\theHtable}{supptab.\arabic{table}}

\section{Full Task Catalog}
\label{app:task_catalog}

Table~\ref{tab:app_full_task_catalog} lists the tasks. For each task we
report the task identifier and name, a short English description, the success
condition used by the evaluator, the robot embodiment, and a scene snapshot.

{\footnotesize
\setlength{\tabcolsep}{2.5pt}
\renewcommand{\arraystretch}{1.06}

\begin{longtable}{@{}>{\raggedright\arraybackslash}m{1.55cm}>{\raggedright\arraybackslash}m{3.85cm}>{\raggedright\arraybackslash}m{4.40cm}>{\raggedright\arraybackslash}m{1.80cm}>{\centering\arraybackslash}m{1.45cm}@{}}
\caption{Full Bench2Dex task catalog: identifier, description, success
condition, robot embodiment, and scene snapshot.}
\label{tab:app_full_task_catalog}\\
\toprule
\textbf{Task} & \textbf{Description} & \textbf{Success Condition} & \textbf{Embodiment} & \textbf{Scene} \\
\midrule
\endfirsthead
\caption[]{Full Bench2Dex task catalog (continued).}\\
\toprule
\textbf{Task} & \textbf{Description} & \textbf{Success Condition} & \textbf{Embodiment} & \textbf{Scene} \\
\midrule
\endhead
\midrule
\multicolumn{5}{r}{\footnotesize Continued on next page} \\
\endfoot
\bottomrule
\endlastfoot
Wine Glass Plate Balance
& Carry three filled wine glasses to the plates of three diners without spilling or knocking over any objects.
& All three wine glasses are kept upright (within $30^\circ$ of vertical), each placed within $0.10$~m of one of the target positions, and fully at rest.
& xArm7+\allowbreak{}Ability & \includegraphics[width=\linewidth]{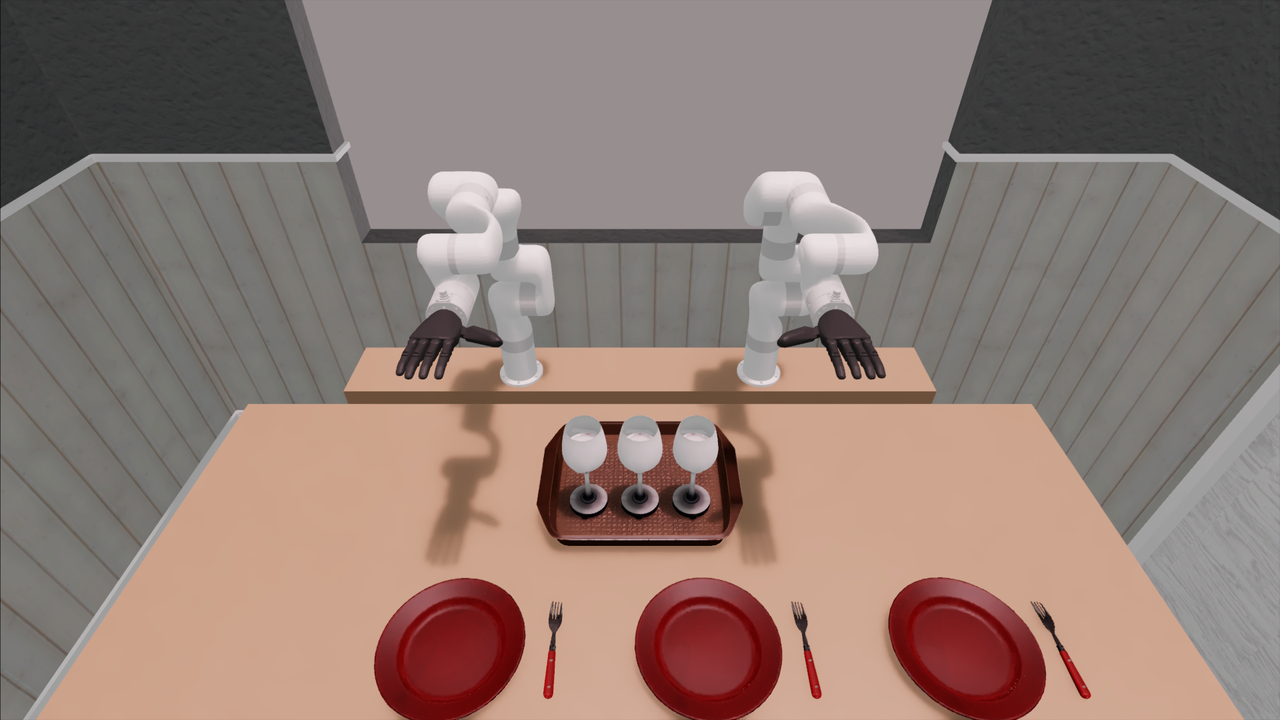} \\[2pt]
Fruit Bowl Loading
& Move the bowl to the center of the table and place two apples and a banana into it.
& The bowl is moved into the central zone of the table and kept upright, and two apples and one banana are placed inside it, all at rest.
& UR5+\allowbreak{}RH56DFX & \includegraphics[width=\linewidth]{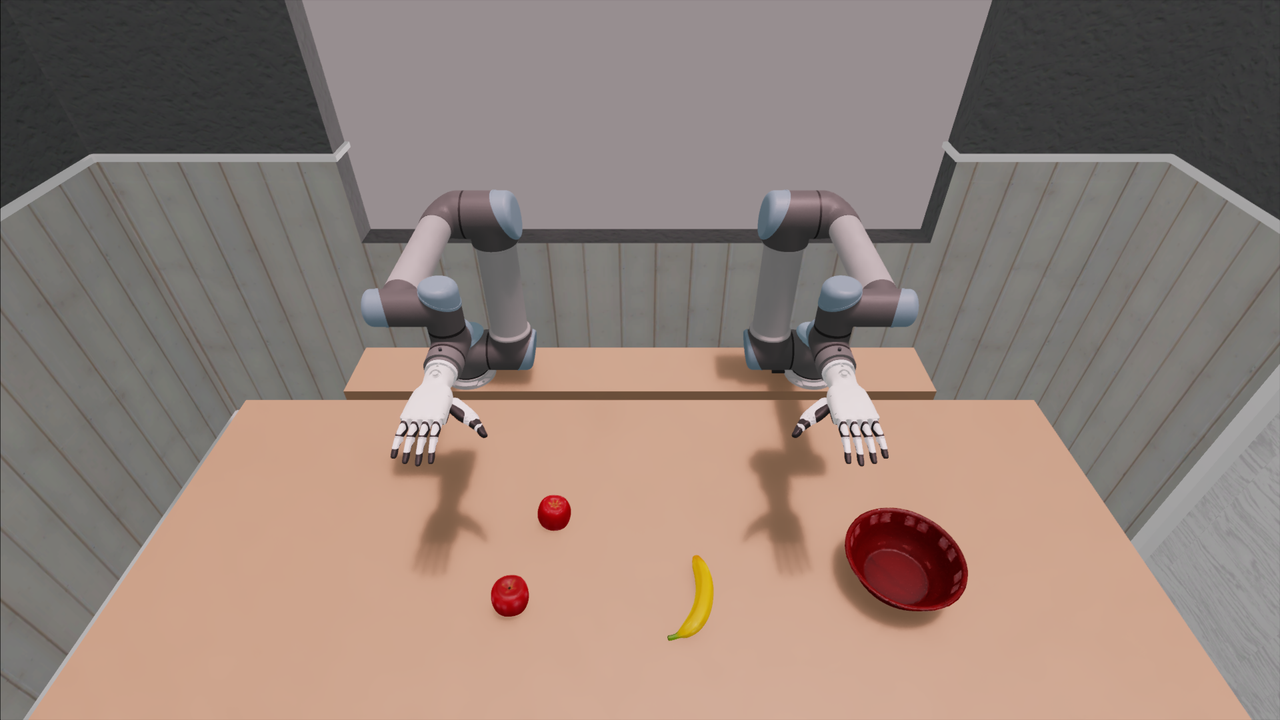} \\[2pt]
Citrus Plate Loading
& Place two lemons and two oranges onto the plate.
& Two lemons and two oranges are placed on the plate, with each object's center within $0.15$~m of the plate interior.
& UR5+\allowbreak{}RH5DG2 & \includegraphics[width=\linewidth]{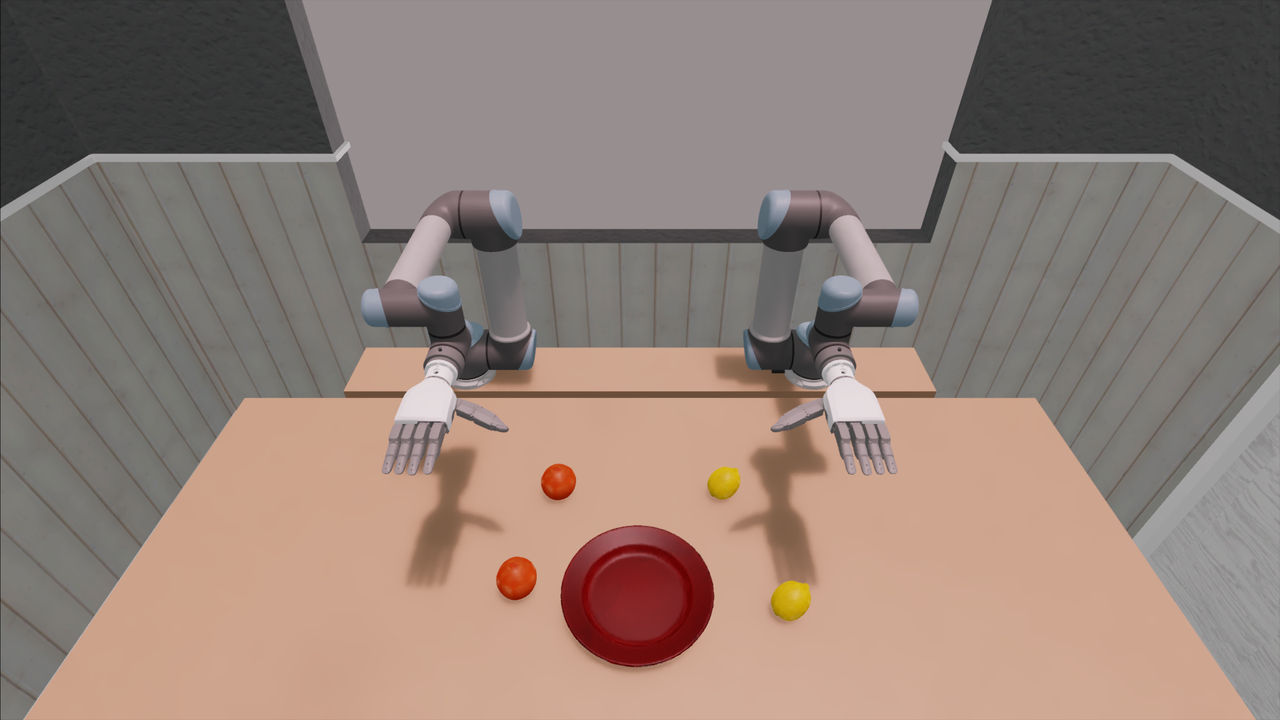} \\[2pt]
Frypan Stand \& Pour
& Place the frypan onto the display stand, then pour the soy sauce and olive oil into the frypan one by one.
& The frypan is placed on the stand; the soy sauce and olive oil are each tilted at least $50^\circ$ with their mouth over the frypan to pour, then both bottles are returned upright and at rest on the table while the bread remains in the pan.
& UR5+\allowbreak{}Schunk & \includegraphics[width=\linewidth]{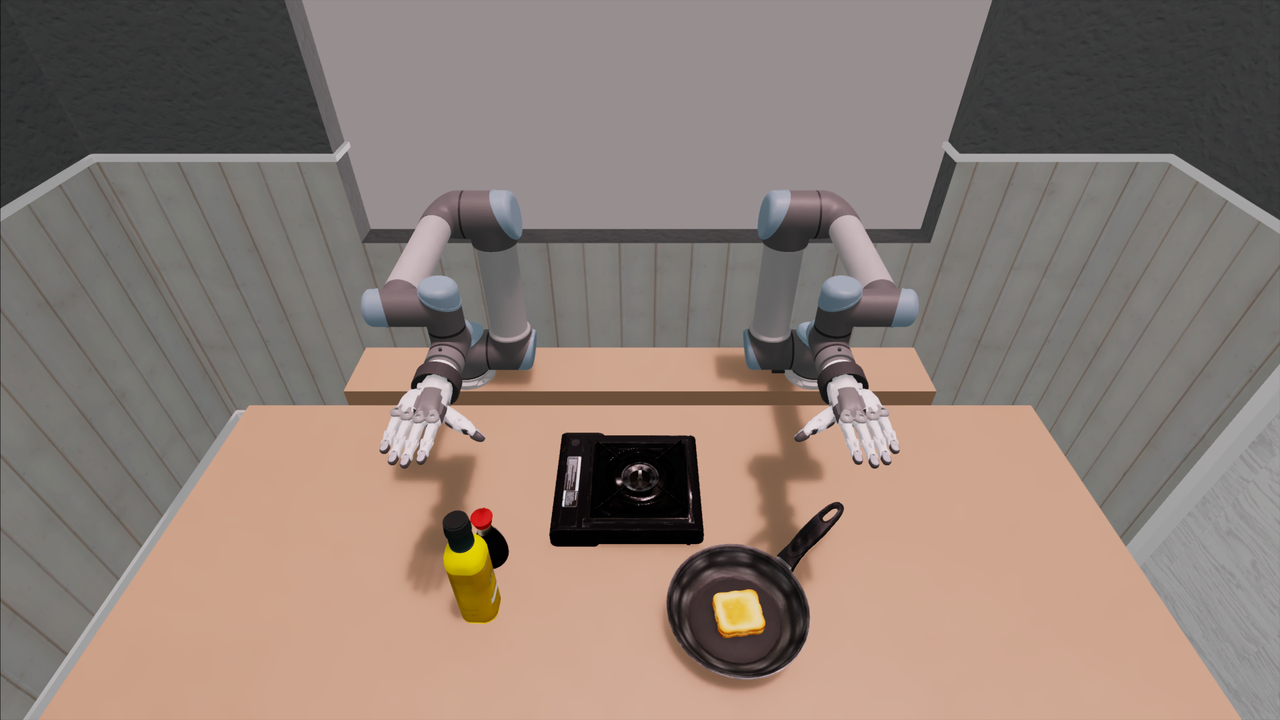} \\[2pt]
Cleaner Box Loading
& Use both hands to lift the cleaner upright into the wooden box, then place the soap into the box.
& The wooden box is upright, the cleaner is inside the box and upright (not tilted), and the soap is inside the box.
& xArm7+\allowbreak{}LEAP & \includegraphics[width=\linewidth]{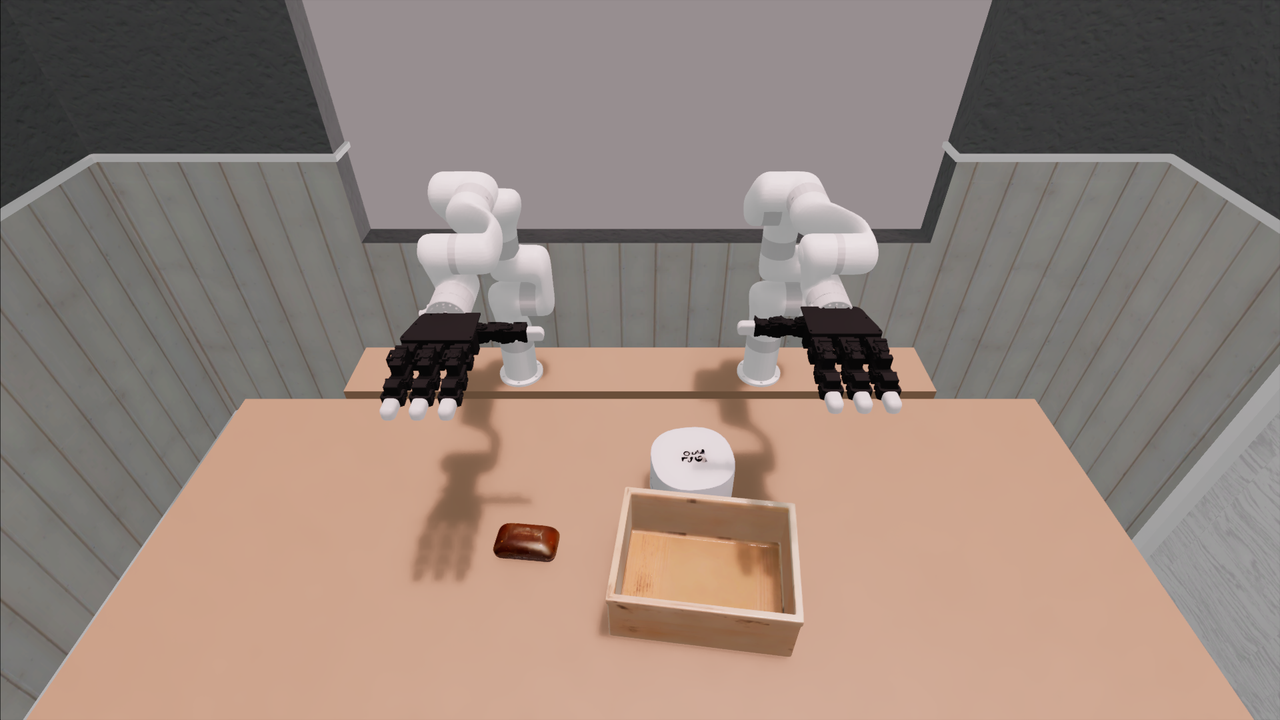} \\[2pt]
Screwdriver Box \& Hammer
& Place the right screwdriver into the box and move the box left, place the left screwdriver into the box, then strike the wooden block once with the hammer and place the hammer into the box.
& Both screwdrivers are inside the upright box, the hammer strikes the wooden block once (swing near the block with a rigid‑body response), and the hammer is then placed inside the box.
& UR5+\allowbreak{}RH56DFX & \includegraphics[width=\linewidth]{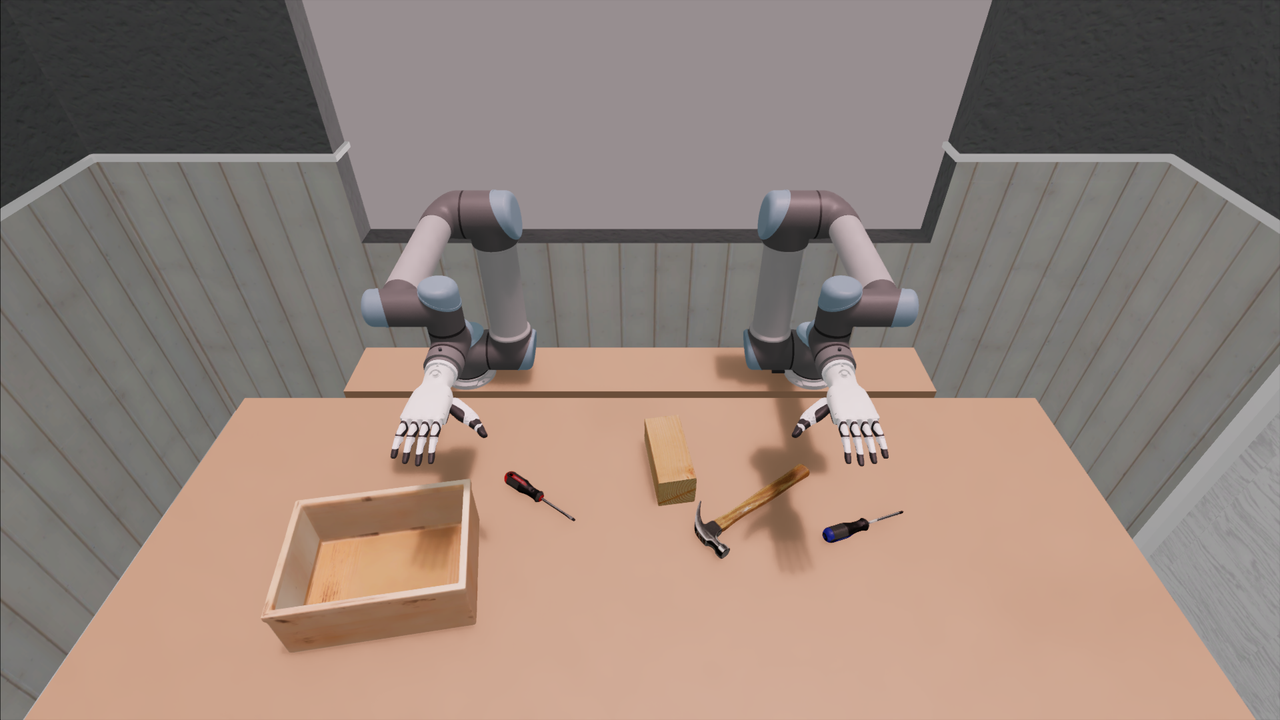} \\[2pt]
Condiment Box Loading
& Place the monosodium glutamate, soy sauce, and vinegar into the wooden box.
& The box is upright; the MSG, soy sauce, and vinegar are all inside the box, with the soy sauce and vinegar kept upright, and all objects at rest.
& UR5+\allowbreak{}Wuji & \includegraphics[width=\linewidth]{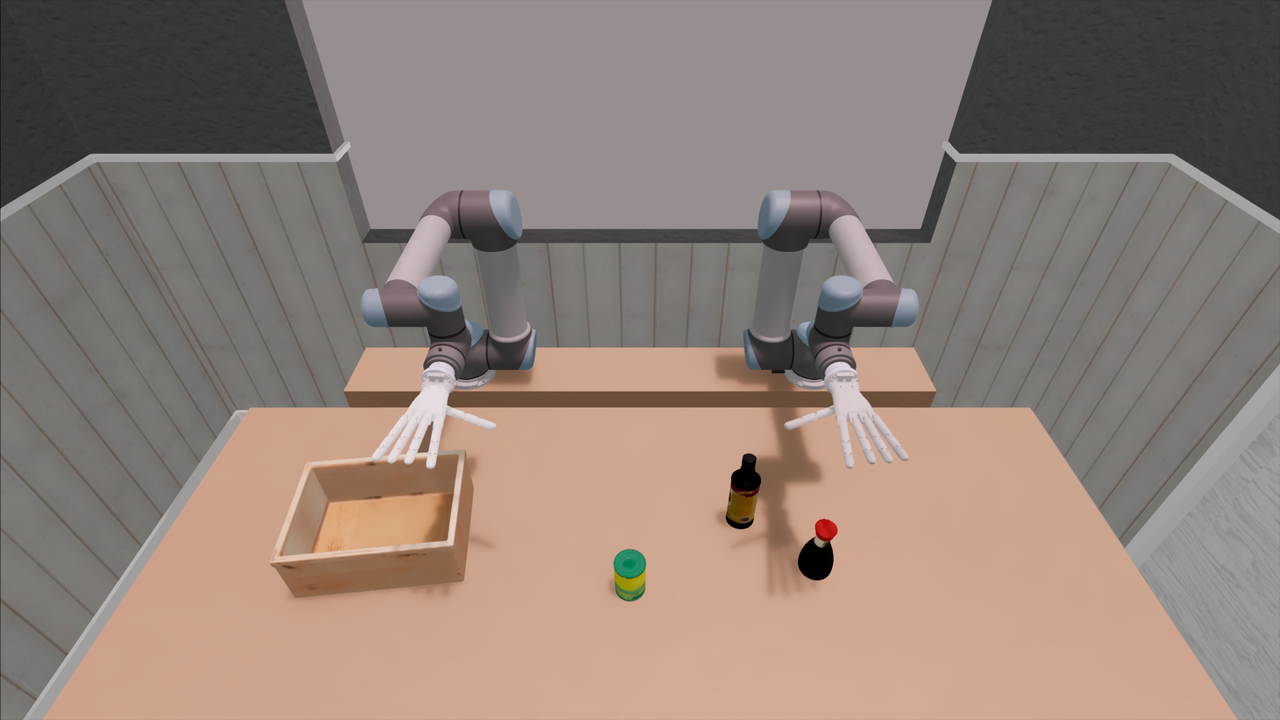} \\[2pt]
Tool Box Loading
& Use the right hand to place the drill and flat screwdriver and the left hand to place the wrench and Phillips screwdriver into the wooden box.
& All four tools (drill, flat screwdriver, adjustable wrench, Phillips screwdriver) are placed inside the upright wooden box.
& Panda+\allowbreak{}Orca & \includegraphics[width=\linewidth]{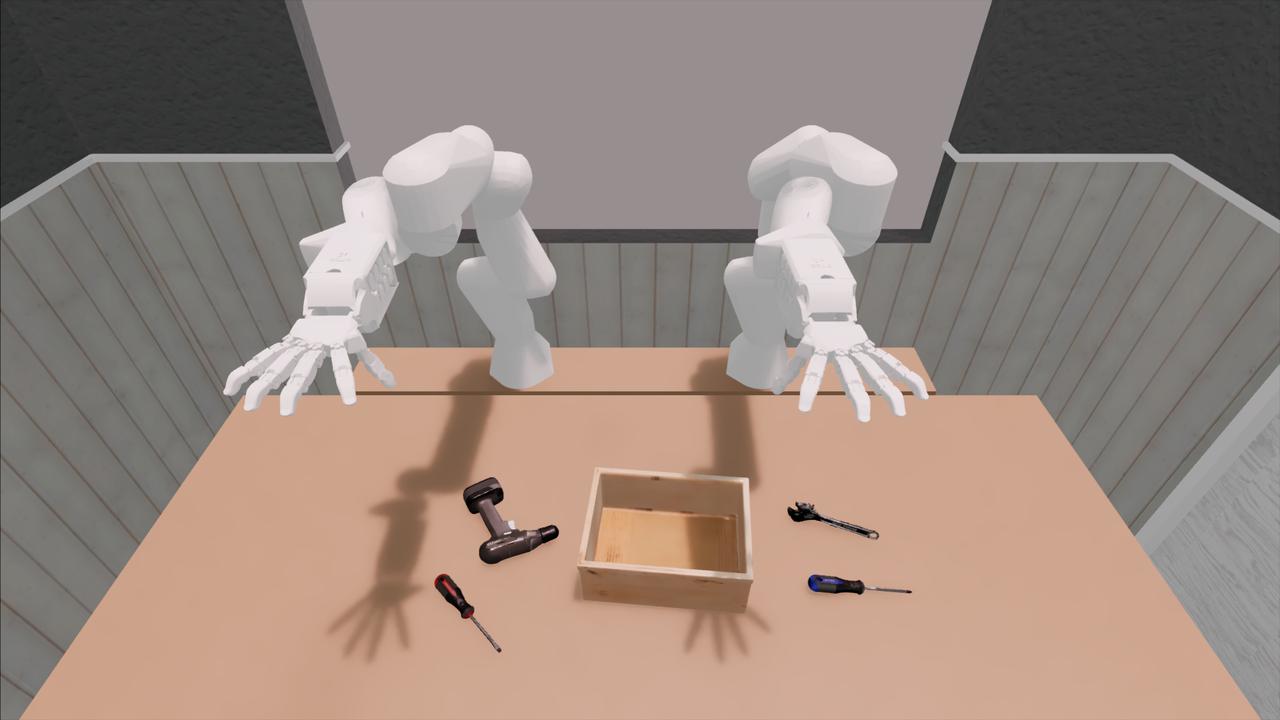} \\[2pt]
Stationery Category Sorting
& Use the right hand to sort the marker and battery into the pen cup and plastic box, and the left hand to sort the gray pen and glue into the respective containers.
& The pen and marker are placed upright in the pen cup and the glue and battery are placed in the plastic box, with both containers upright and all objects at rest.
& RM65+\allowbreak{}Revo2 & \includegraphics[width=\linewidth]{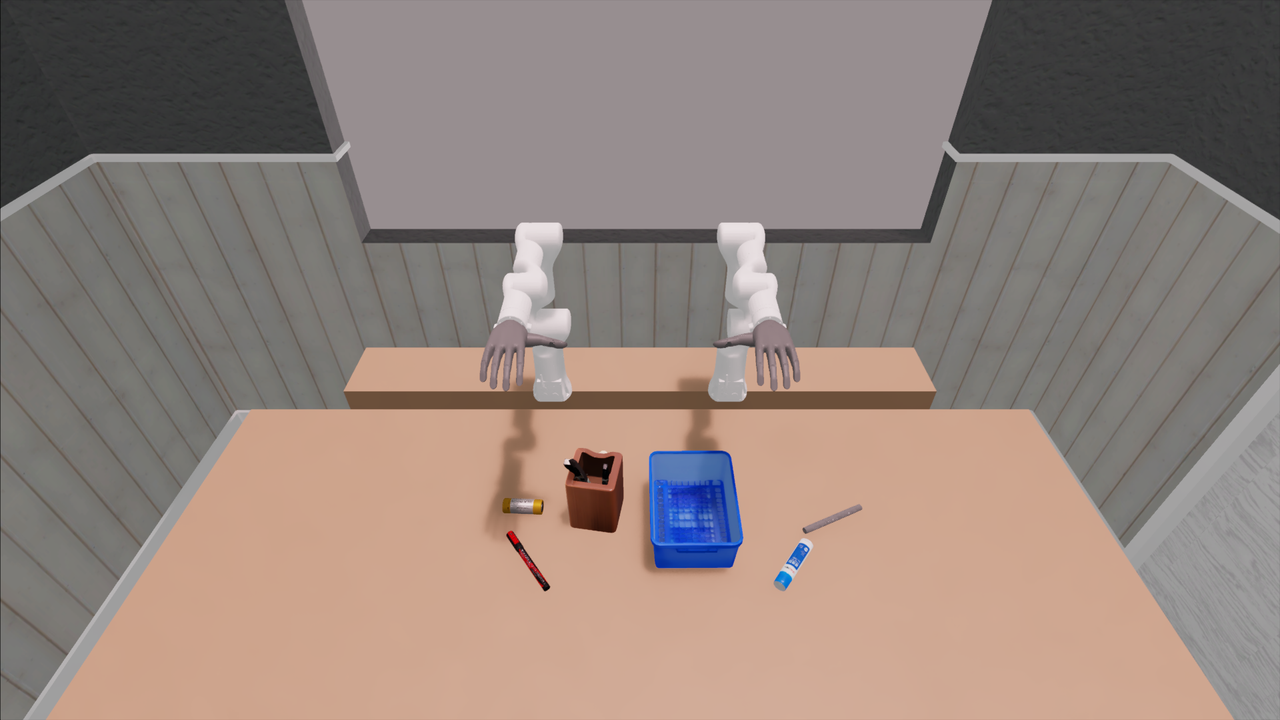} \\[2pt]
Canned Food Tray Arrangement
& Use the left hand to place the master chef can and MSG and the right hand to place the milk box and potted meat can onto the tray.
& The master chef can, MSG, milk box, and potted meat can are all placed on the tray and kept upright, with the tray upright and all objects at rest.
& IIWA7+\allowbreak{}Sharpa & \includegraphics[width=\linewidth]{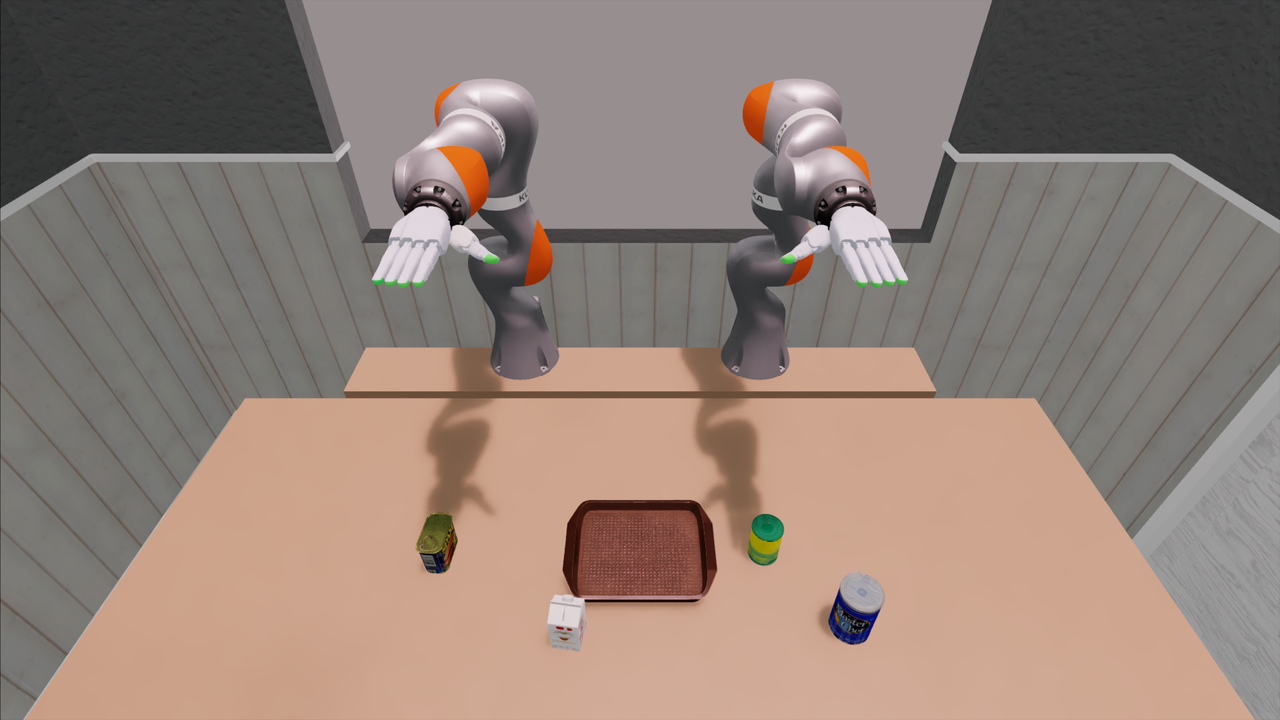} \\[2pt]
Ball Box Loading
& Place the mini soccer ball, tennis ball, golf ball, and ping‑pong ball into the box.
& The box is upright and all four balls are placed inside it, each within $0.22$~m of the box interior.
& UR5+\allowbreak{}Wuji & \includegraphics[width=\linewidth]{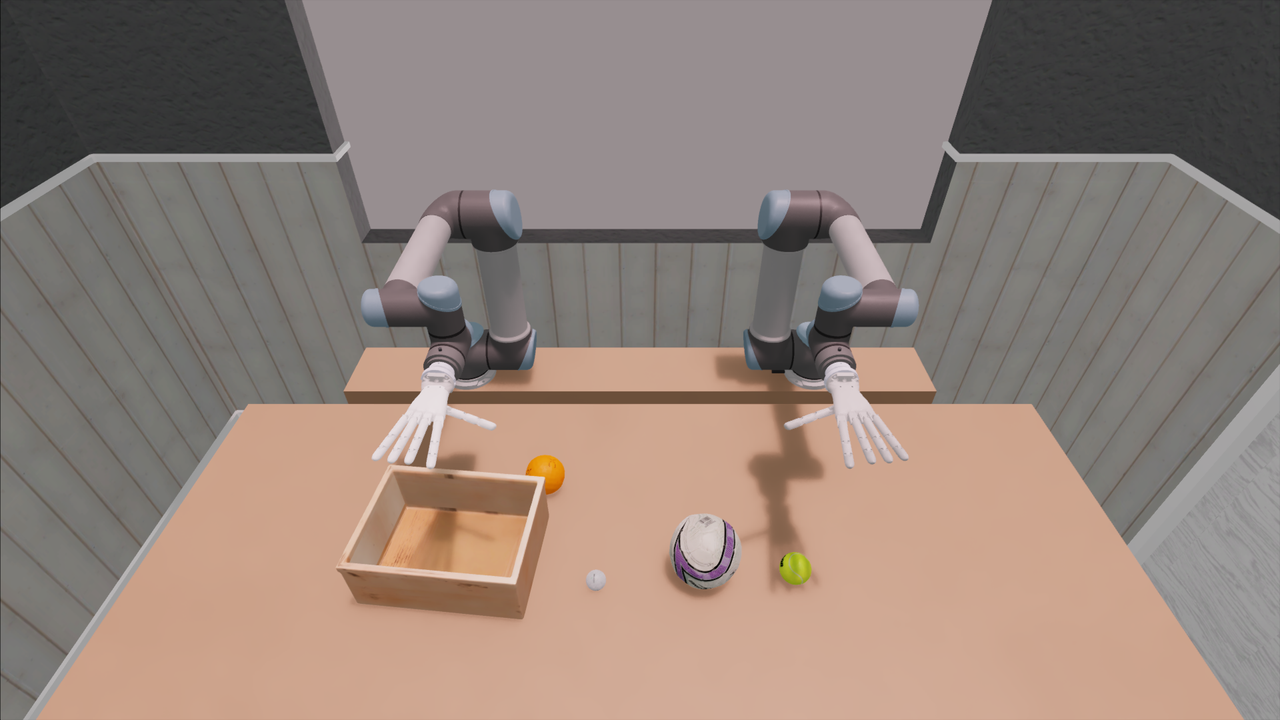} \\[2pt]
Baking Tray Prep
& Use the left hand to place the brush and spatula and the right hand to place the small pudding box and large gelatin box onto the tray.
& The brush, spatula, pudding box, and gelatin box are all placed on the tray, with the tray kept upright and all objects at rest.
& IIWA7+\allowbreak{}Sharpa & \includegraphics[width=\linewidth]{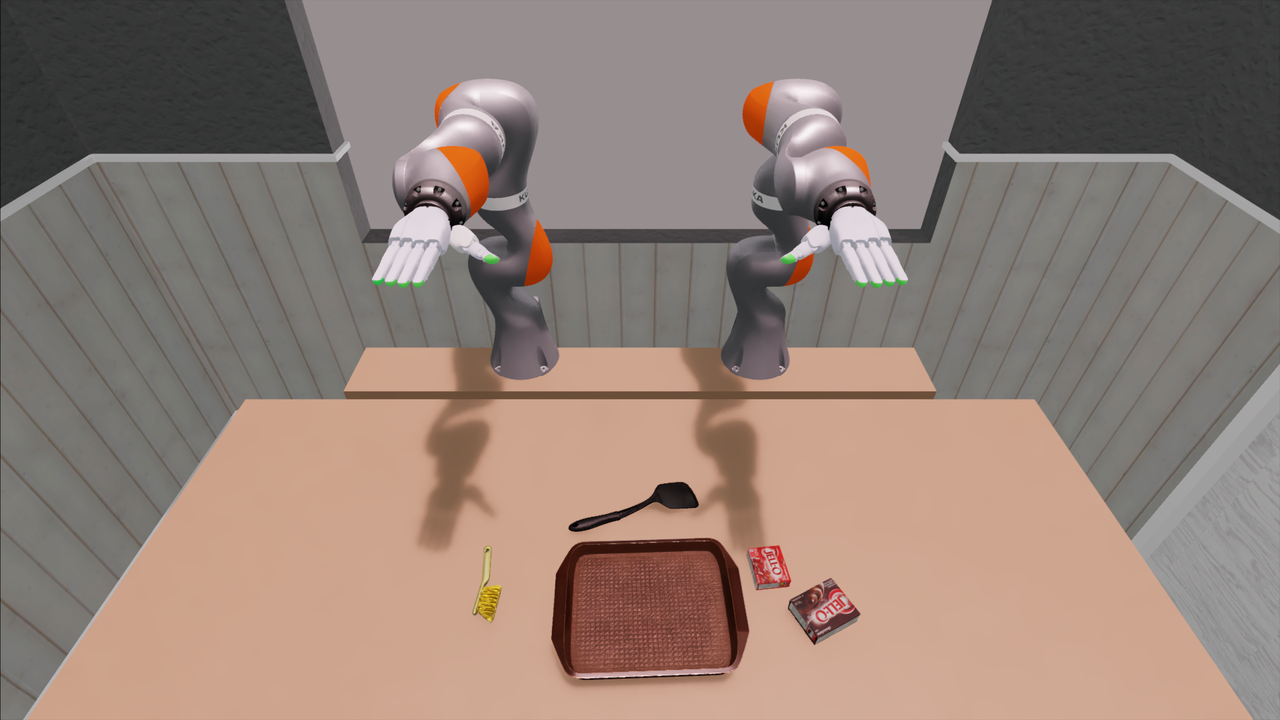} \\[2pt]
Fridge Wine Interhand Pour
& Open the refrigerator, take out the wine bottle, hand it to the right hand in the air to pour a glass of wine, return it to the left hand to put back into the fridge, and close the door.
& The fridge door is opened, the wine bottle is lifted out, tilted at least $50^\circ$ toward the glass to pour, then returned upright inside the fridge, and the door is closed.
& UR5+\allowbreak{}RH5DG2 & \includegraphics[width=\linewidth]{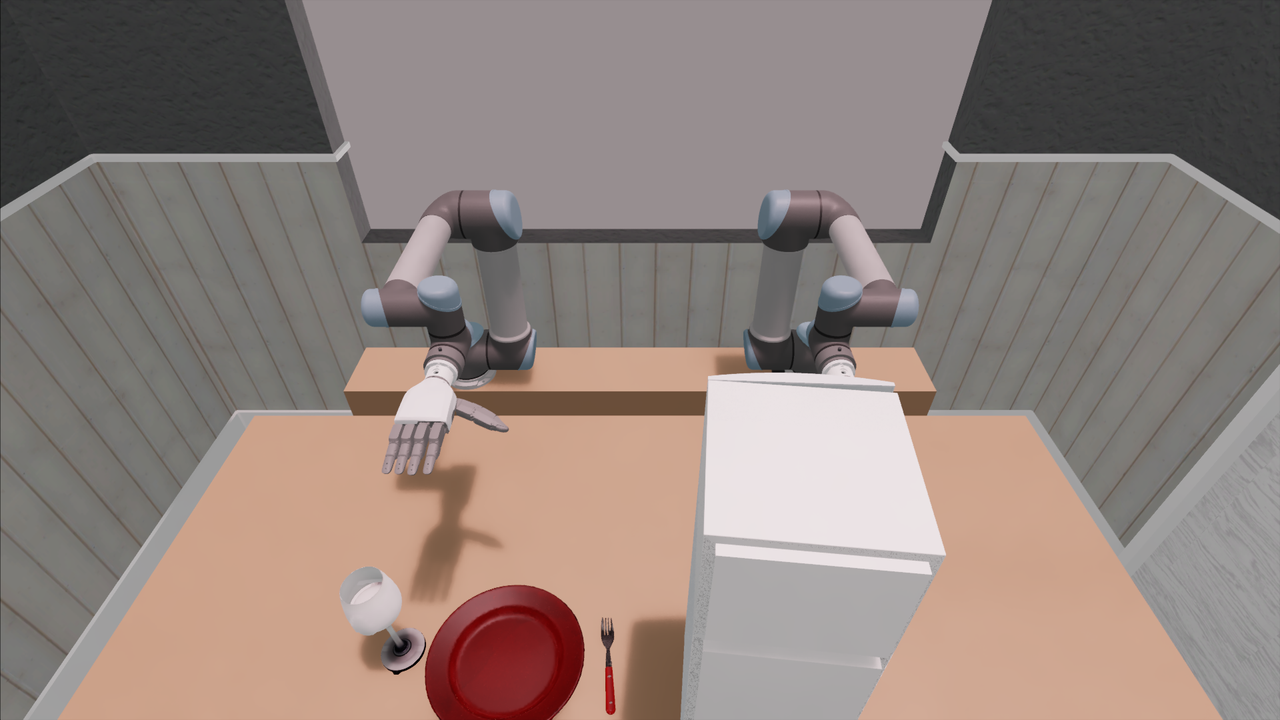} \\[2pt]
Trash Disposal
& Open the trash can lid, throw the crumpled paper into the trash can, use the left hand to throw the bottle and banana peel into the trash can, then close the lid.
& The trash can is upright and the crumpled paper, bottle, and banana peel are all placed inside it, at rest.
& UR5+\allowbreak{}RH56DFX & \includegraphics[width=\linewidth]{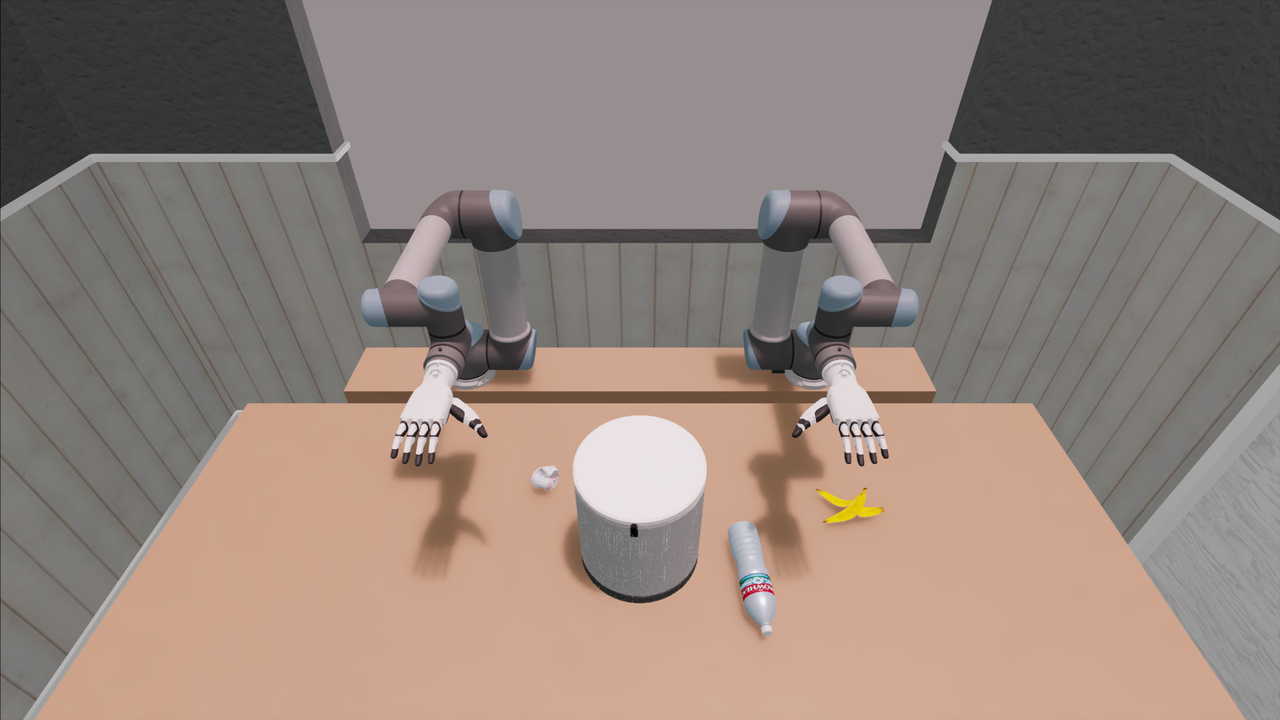} \\[2pt]
Fridge Fruit Shelf Sorting
& Move the banana to the left side of the desk, open both fridge doors, place the lemon on the upper shelf and the banana on the lower shelf, then close the door.
& The banana and lemon are placed inside the fridge, both fridge doors are closed, and both objects are at rest.
& UR5+\allowbreak{}Shadow & \includegraphics[width=\linewidth]{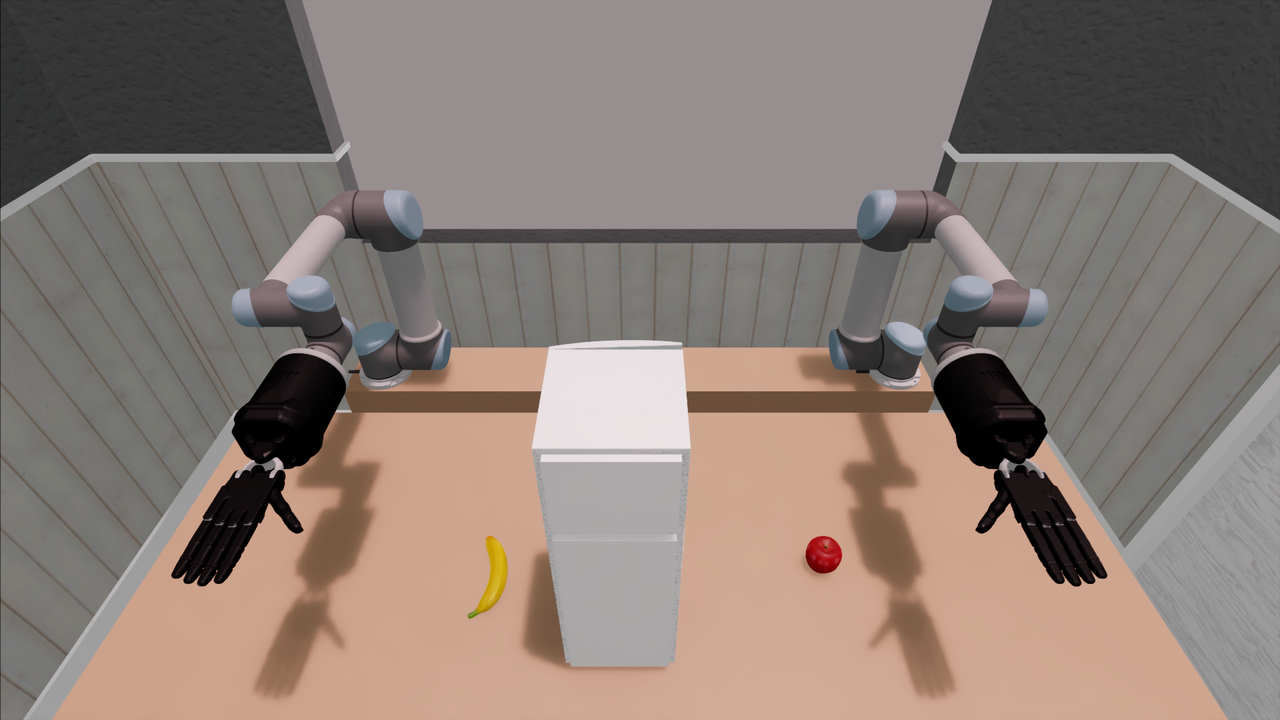} \\[2pt]
Microwave Bowl Loading
& Open the microwave door, bring the bowl to the microwave, place the baguette into the bowl, place the bowl into the microwave, and close the door.
& The bowl is inside the microwave cavity and upright, the baguette is placed in the bowl, and the microwave door is closed.
& UR5+\allowbreak{}Schunk & \includegraphics[width=\linewidth]{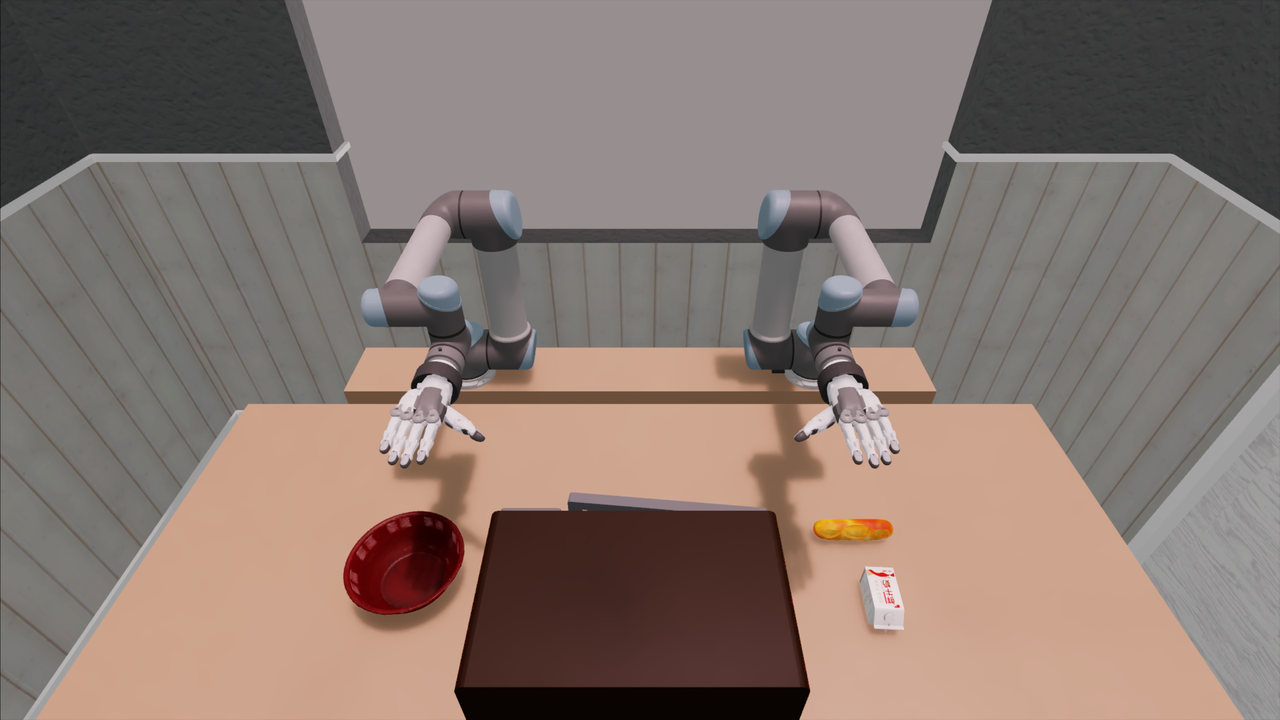} \\[2pt]
Toilet Lid Cleaner Pour
& Open the toilet lid, pour the cleaner into the toilet, and close the toilet lid.
& The toilet lid is opened, the cleaner is lifted and tilted at least $50^\circ$ toward the toilet bowl to pour while the lid stays open, and the lid is then closed.
& xArm7+\allowbreak{}LEAP & \includegraphics[width=\linewidth]{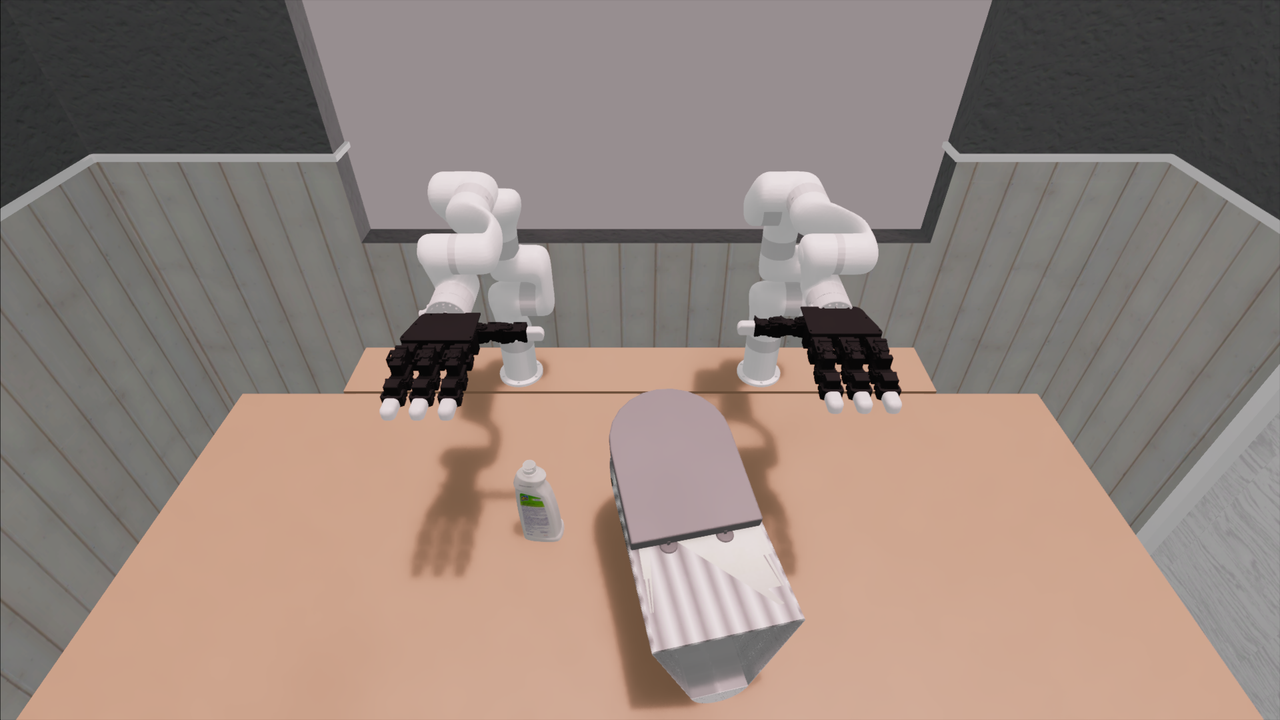} \\[2pt]
Breadbasket Fast‑Food Loading
& Place the baguette and bread into the bread basket, move the basket to the center of the table, then place the hamburger and french fries into the basket.
& The bread basket is upright and the french fries, hamburger, bread, and baguette are all placed inside it.
& UR5+\allowbreak{}RH5DG2 & \includegraphics[width=\linewidth]{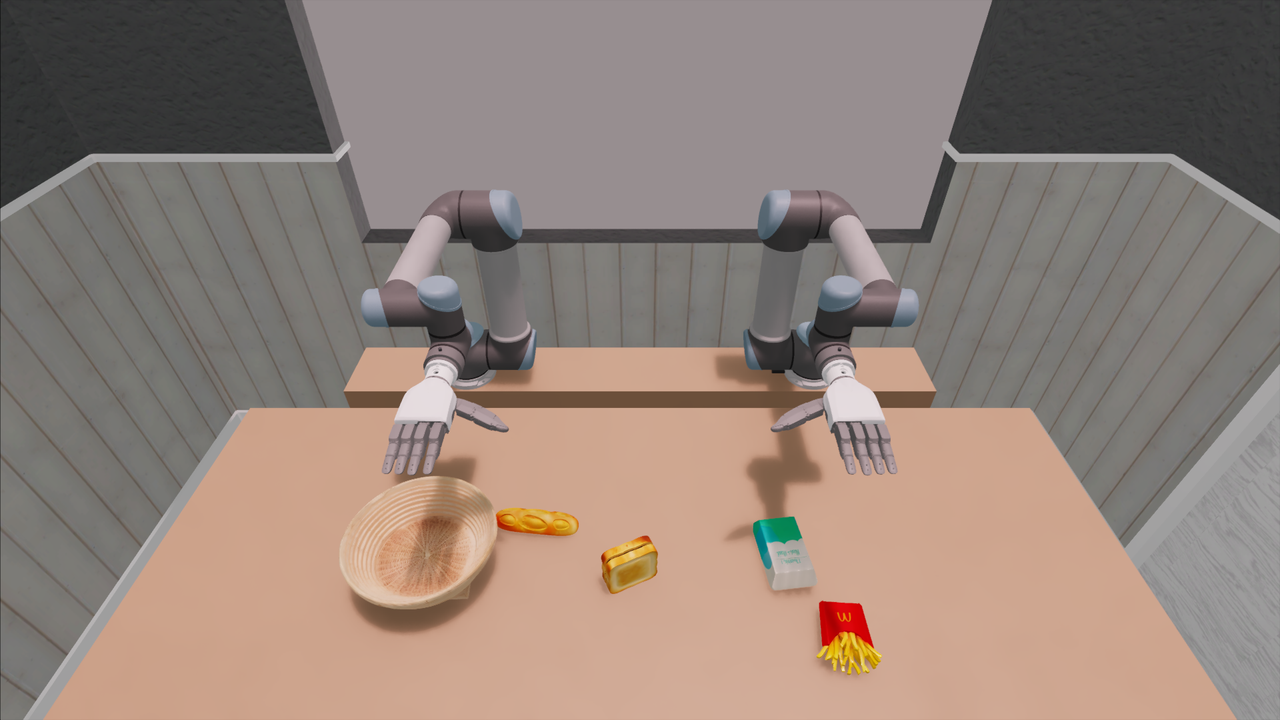} \\[2pt]
Medicine Shoebox Pack
& Move the shoe box to the center of the table, then place the pill bottle, toothpaste, and hydrating oil into the box.
& The shoe box is upright and the pill bottle, toothpaste, and hydrating oil are all placed inside it.
& Panda+\allowbreak{}Orca & \includegraphics[width=\linewidth]{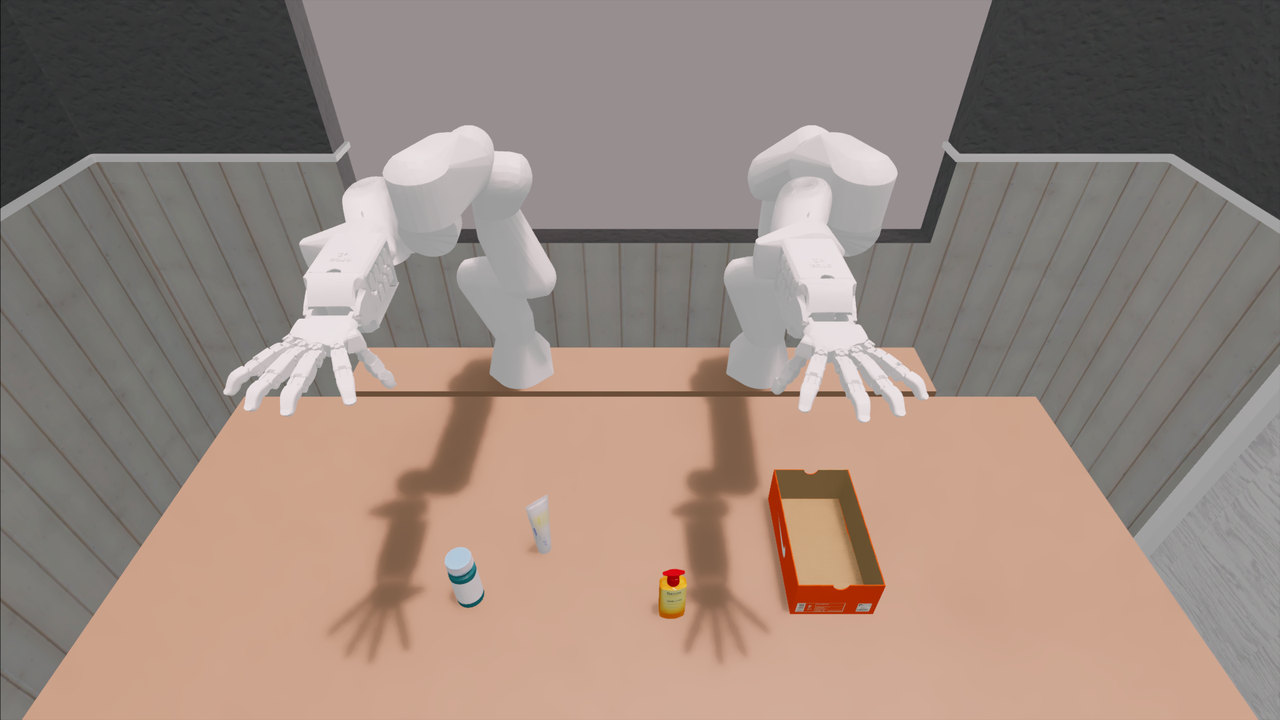} \\[2pt]
Shoebox Accessory Pack
& Place the seal into the shoe box and move the box to the center of the table, then place the shoe and pet collar into the box.
& The shoe box is upright and the seal, shoe, and pet collar are all placed inside it, at rest.
& xArm7+\allowbreak{}LEAP & \includegraphics[width=\linewidth]{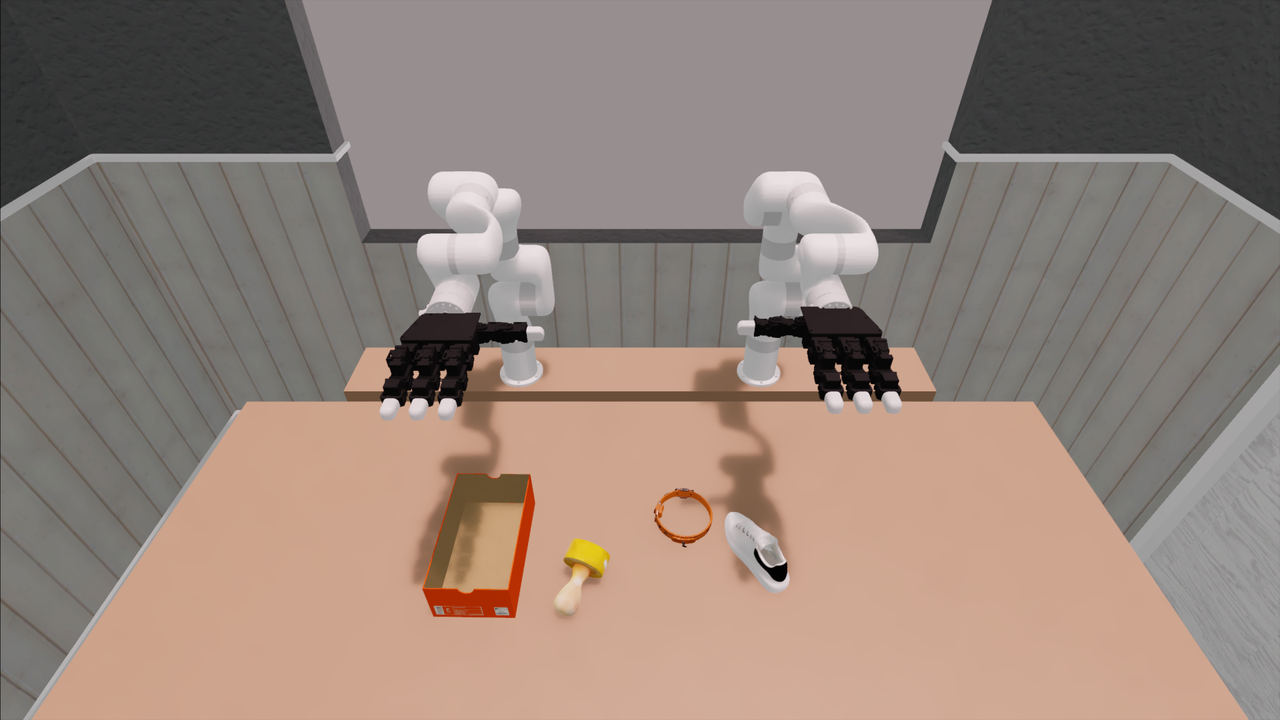} \\[2pt]
Sports Ball Cup Sort
& Place the tennis ball and baseball into the large cup and the racquetball and golf ball into the small cup.
& Both cups are upright; the tennis ball and baseball are inside the large cup and the racquetball and golf ball are inside the small cup, all at rest.
& Panda+\allowbreak{}Allegro & \includegraphics[width=\linewidth]{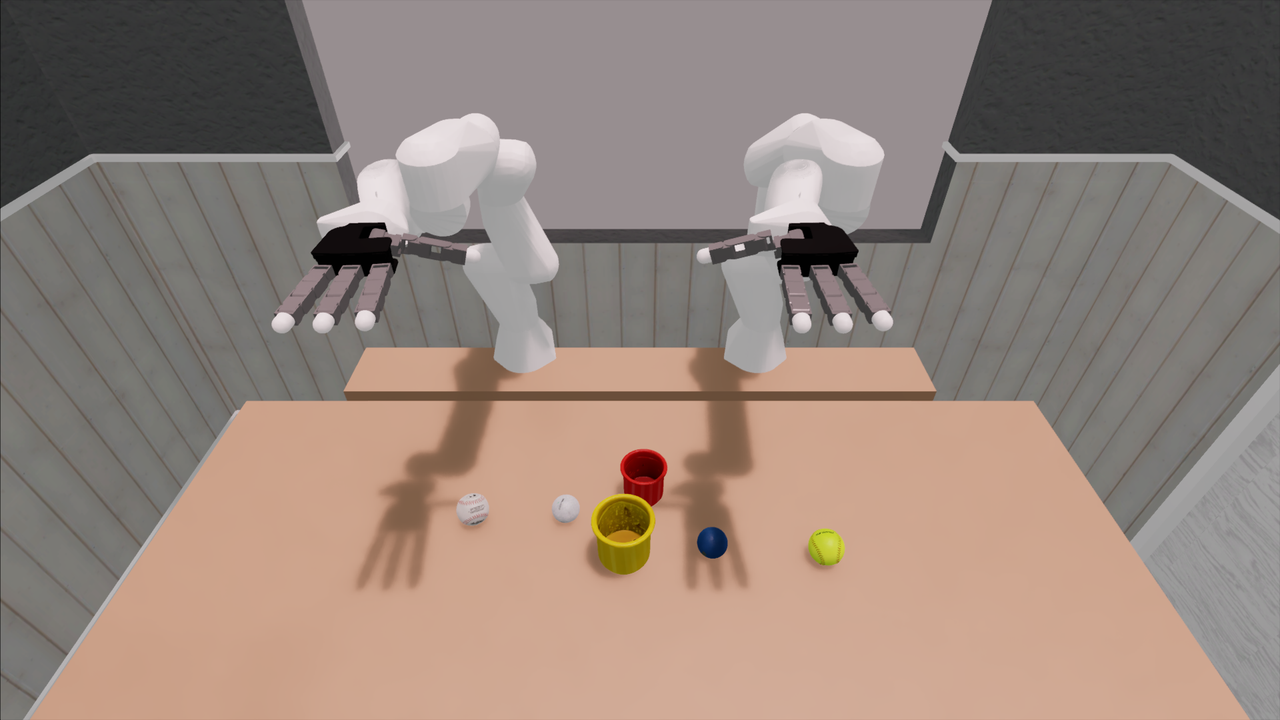} \\[2pt]
Faucet Cup Water Fill
& Place the spoon into the mug, place the mug under the faucet, open the faucet to fill the mug and then close it, and place the mug onto the tray.
& The spoon is placed in the upright mug; while the mug remains under the faucet opening, the faucet is opened and then closed again; and the mug is placed at rest on the tray.
& RM65+\allowbreak{}Revo2 & \includegraphics[width=\linewidth]{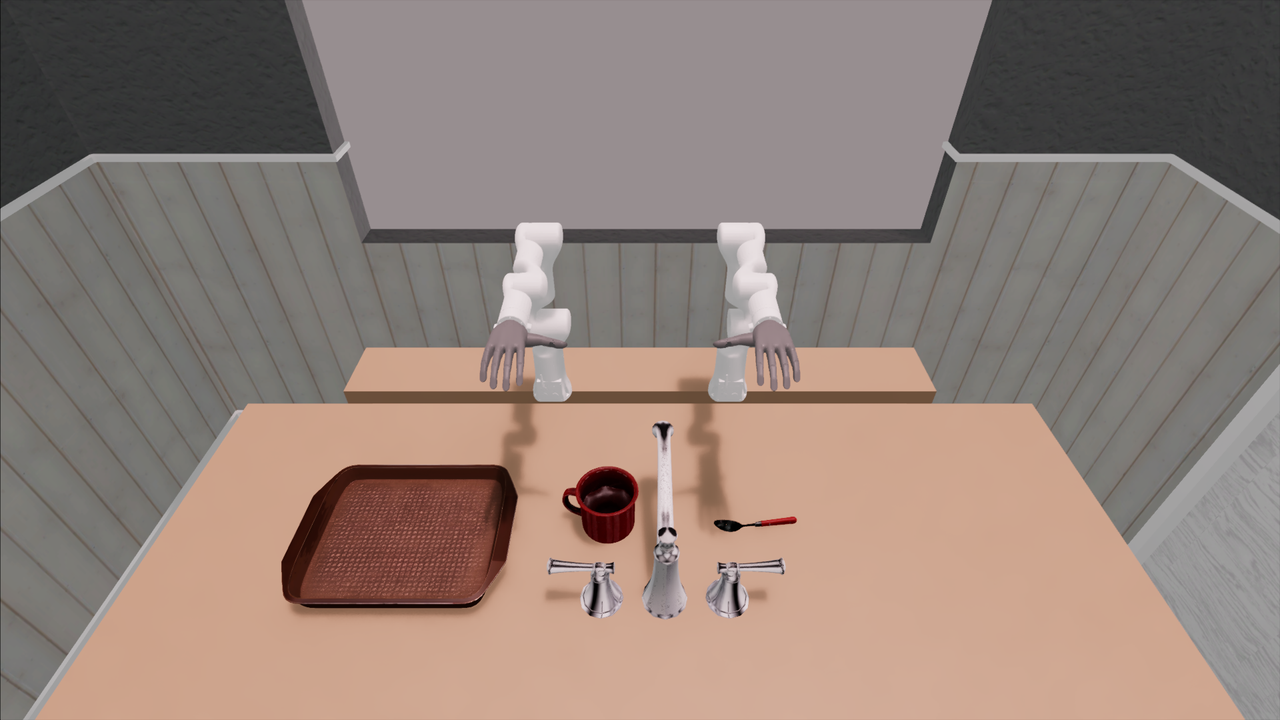} \\[2pt]
Jigsaw Puzzle Assembly
& Use the right and left hands in turn to assemble the green, red, blue, and yellow puzzle pieces onto the fixed white puzzle piece in the center of the table, forming a rectangle.
& All four colored puzzle pieces are placed at their target positions around the fixed white piece (within $0.02$~m in $xy$ and $0.01$~m in height of the reference), and all pieces are at rest.
& IIWA7+\allowbreak{}Sharpa & \includegraphics[width=\linewidth]{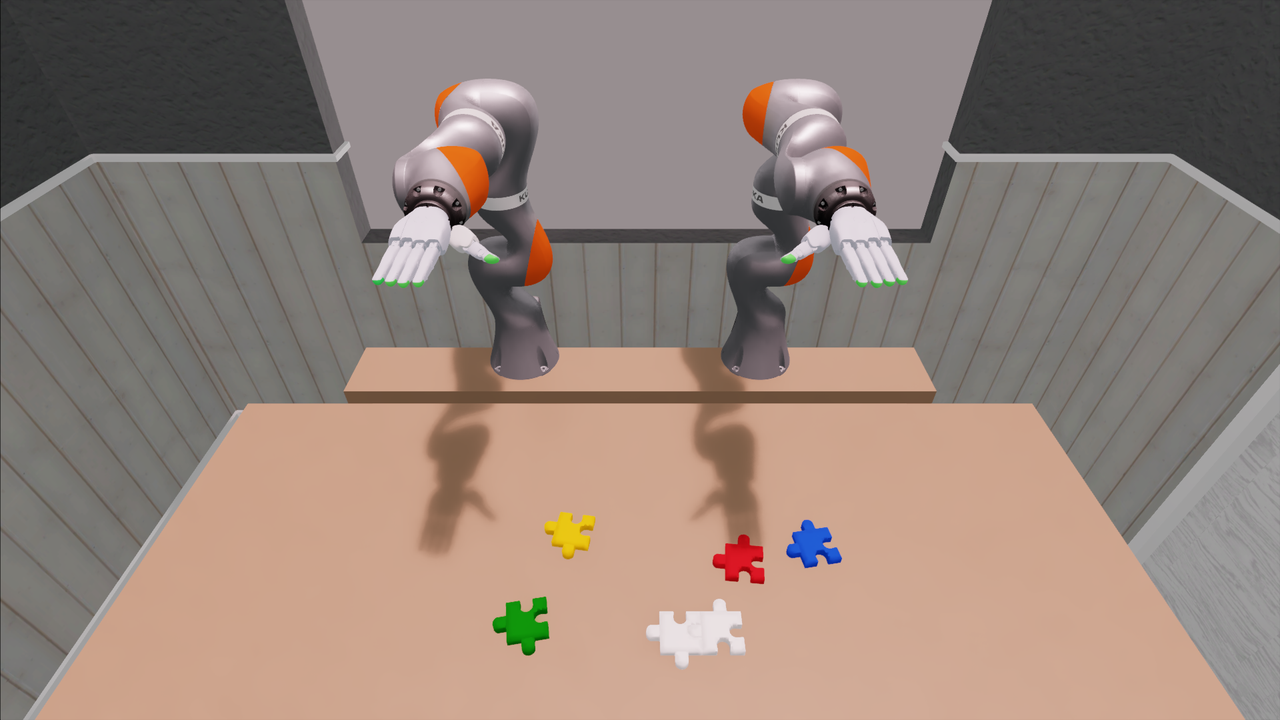} \\[2pt]
Soup Serving
& Hold the bowl beside the pot with the left hand, ladle two scoops of soup from the pot into the bowl with the right hand, carry the bowl to the front‑right area, and return the ladle.
& The pot is on the stove, the bowl is held near the pot, the ladle dips into the pot and is brought over the bowl twice, the bowl is delivered to the front‑right serving zone, and the ladle is returned.
& UR5+\allowbreak{}Shadow & \includegraphics[width=\linewidth]{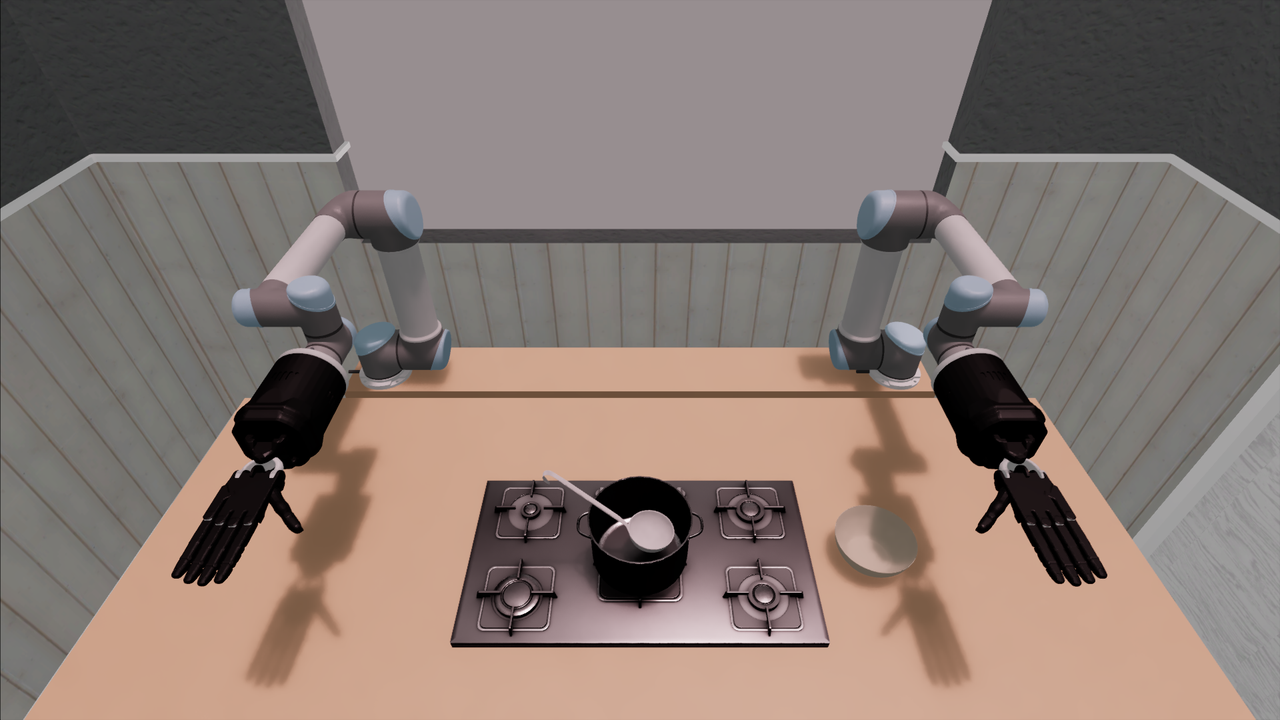} \\[2pt]
Bimanual Piano Melody
& Both hands play the piano key sequence C‑C‑G‑G‑A‑A‑G together, the left hand playing the bass notes and the right hand the treble notes.
& The left hand plays the bass melody and the right hand plays the treble melody C‑C‑G‑G‑A‑A‑G, with each note key pressed in sequence.
& xArm7+\allowbreak{}Ability & \includegraphics[width=\linewidth]{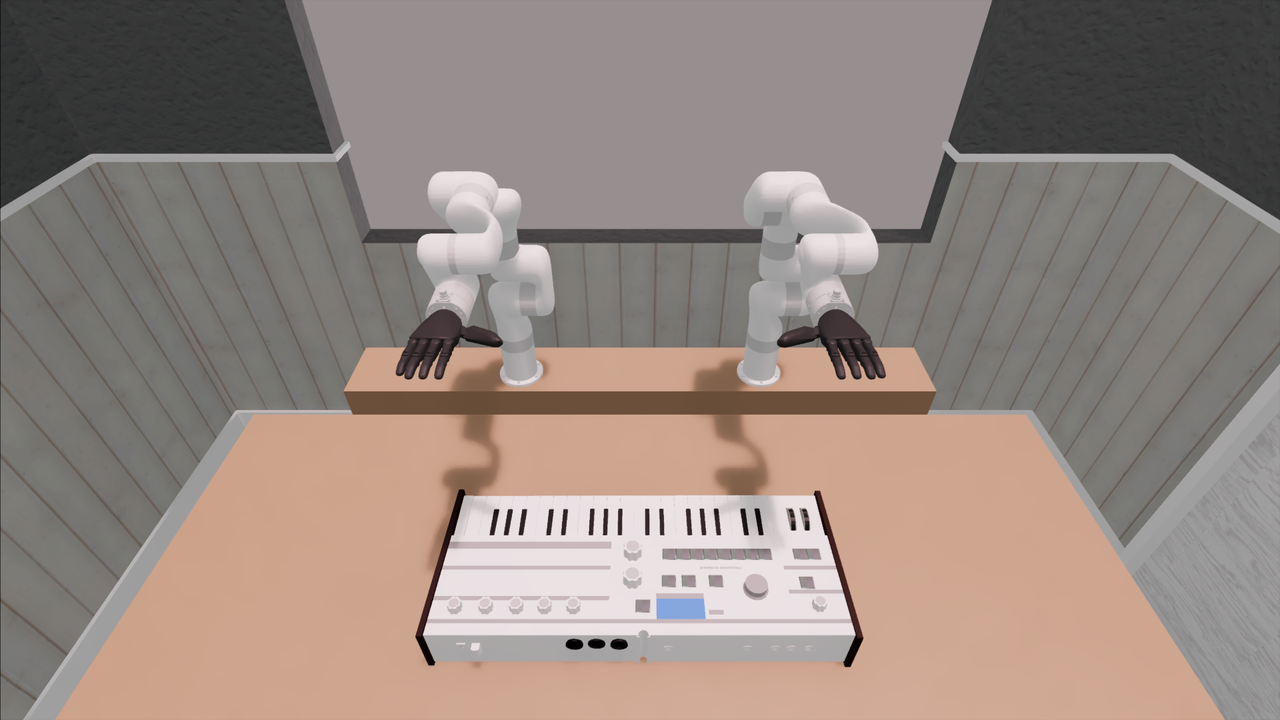} \\[2pt]
Gaming Desk Setup
& Straighten the monitor screen forward with both hands, press the ESC key, move the mouse back to the left side of the keyboard, and click the left mouse button.
& The monitor is straightened forward, the ESC key is pressed once, the mouse is returned to the left of the keyboard, and the left mouse button is clicked.
& JAKA ZU7+\allowbreak{}DexHand021 & \includegraphics[width=\linewidth]{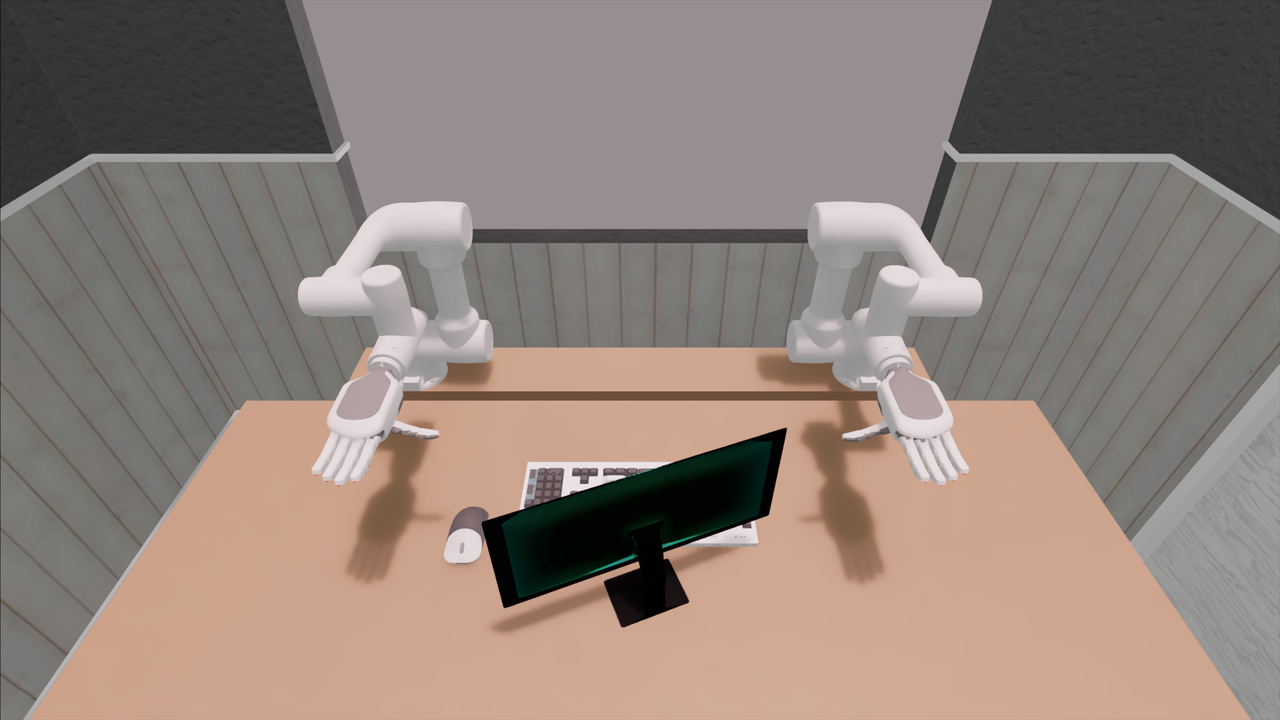} \\[2pt]
\bottomrule
\end{longtable}

}

\section{Supported Bimanual Robot Embodiments}
\label{app:robot_embodiments}

This section summarizes the 12 bimanual robot embodiments supported by Bench2Dex. Each embodiment combines a robot arm and a dexterous hand, enabling evaluation across different arm kinematics, hand morphologies, palm geometries, and finger configurations.

\setlength{\LTcapwidth}{\textwidth}
\setlength{\LTleft}{0pt}
\setlength{\LTright}{0pt}

\newcolumntype{M}[1]{>{\raggedright\arraybackslash}m{#1}}
\newcolumntype{G}[1]{>{\centering\arraybackslash}m{#1}}

\newcommand{\renderframe}[3]{%
    \begingroup
    \setlength{\fboxsep}{0pt}%
    \fcolorbox{gray!35}{white}{\includegraphics[width=#2,height=#3]{#1}}%
    \endgroup
}

\newcommand{\armrender}[1]{%
    \begin{minipage}[c][1.72cm][c]{2.32cm}
        \centering
        \renderframe{#1}{2.18cm}{1.57cm}
    \end{minipage}
}

\newcommand{\armbankrow}[3]{%
    \parbox[c][1.82cm][c]{2.72cm}{\raggedright\textbf{#1}} &
    \armrender{#2} &
    \parbox[c][1.82cm][c]{7.70cm}{\raggedright #3} \\
}

{\footnotesize
\setlength{\tabcolsep}{4.0pt}
\renewcommand{\arraystretch}{1.04}

\begin{longtable}{@{}M{2.80cm}G{2.36cm}M{7.78cm}@{}}
\caption{Supported bimanual robot embodiments in Bench2Dex.}
\label{tab:appendix_robot_embodiments}\\

\toprule
\textbf{Embodiment} & \textbf{Render} & \textbf{Description} \\
\midrule
\endfirsthead

\caption[]{Supported bimanual robot embodiments in Bench2Dex (continued).}\\
\toprule
\textbf{Embodiment} & \textbf{Render} & \textbf{Description} \\
\midrule
\endhead

\midrule
\multicolumn{3}{r}{\footnotesize Continued on next page} \\
\endfoot

\bottomrule
\endlastfoot

\armbankrow{JAKA ZU7+DexHand021}
{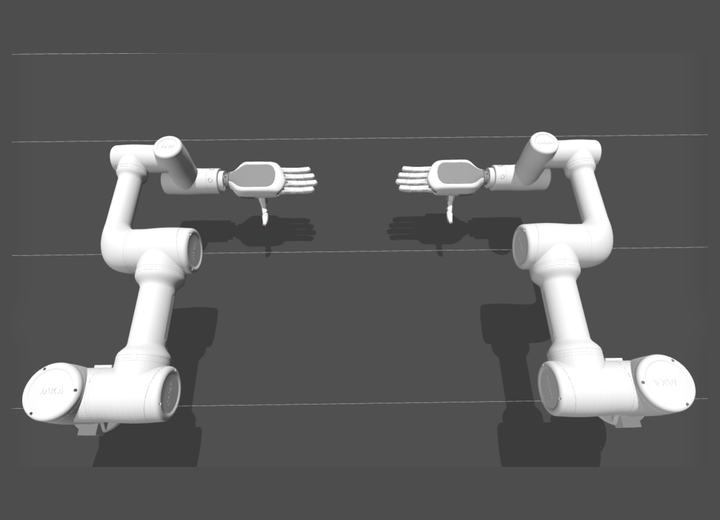}
{A metallic-gray collaborative arm with blue circular joint caps, paired with a silver dexterous hand with dense exposed mechanisms, a ribbed palm surface, and slender articulated fingers. The transparent wrist housing and mechanical finger structure make it distinctive.}

\armbankrow{IIWA7+Sharpa}
{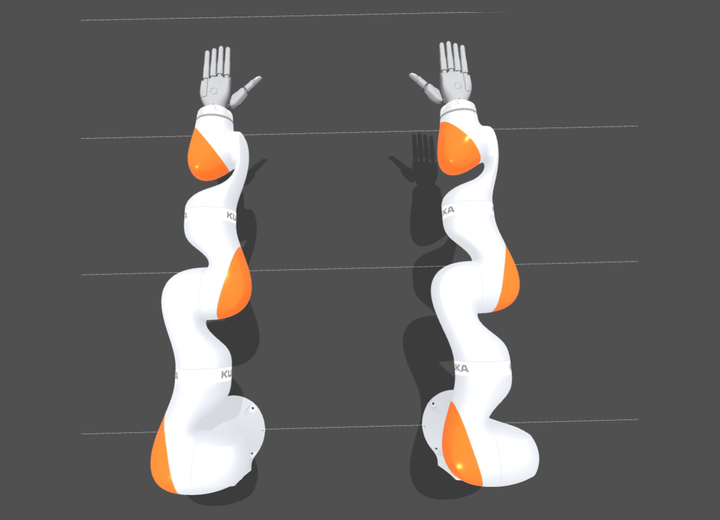}
{A white KUKA-style arm with bright orange accent panels, paired with a smooth anthropomorphic hand. The clean enclosed hand silhouette contrasts with the strong white-orange arm styling.}

\armbankrow{Panda+Allegro}
{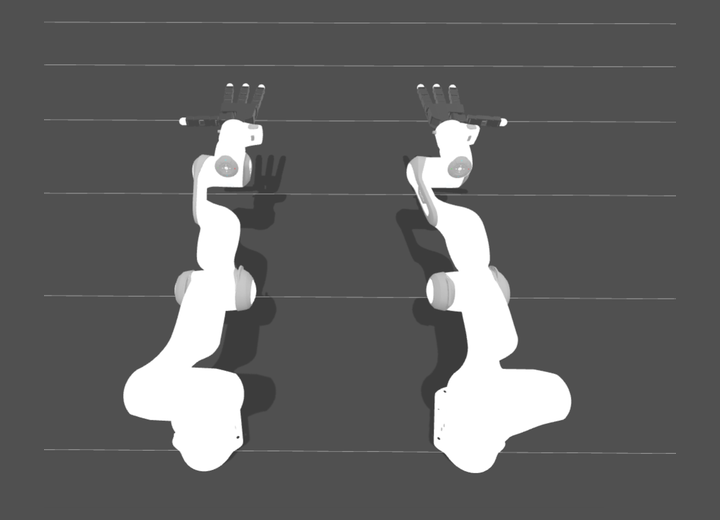}
{A light Panda-style arm paired with a compact dark modular hand. The boxy palm, rectangular segmented fingers, and bright fingertip caps create a sharp contrast with the soft industrial arm.}

\armbankrow{Panda+Orca}
{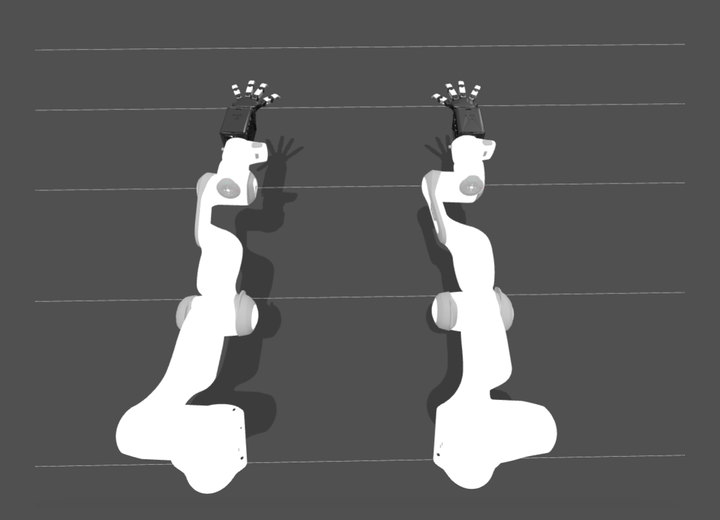}
{A light Panda-style arm attached to a dark hand with a narrow palm and strongly spread fingers. Long separated digits with bright caps create an open fan-like silhouette.}

\armbankrow{RM65+Revo2}
{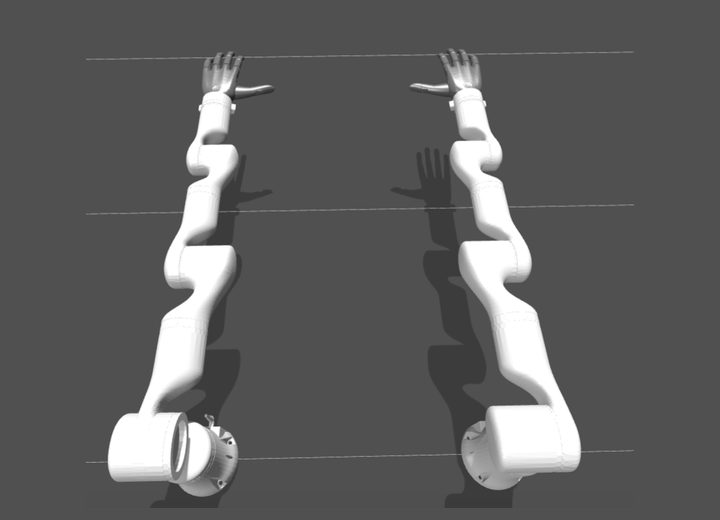}
{A smooth white arm paired with a compact anthropomorphic hand with a rounded palm shell, slim fingers, and a clean enclosed structure. The embodiment appears polished and tightly integrated.}

\armbankrow{UR5+RH56DFX}
{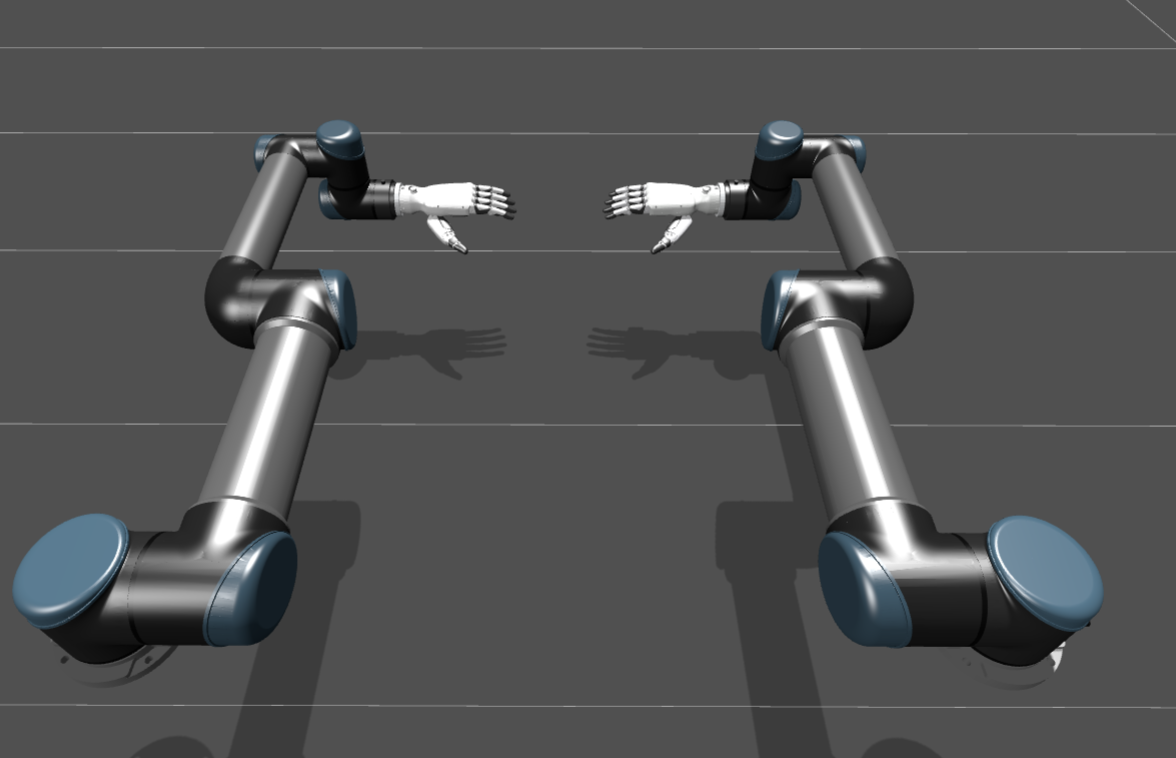}
{A metallic-gray industrial arm with rounded cylindrical links and blue joint caps, ending in a white-and-silver dexterous hand with a sculpted palm shell, dark finger pads, and a large side thumb.}

\armbankrow{UR5+RH5DG2}
{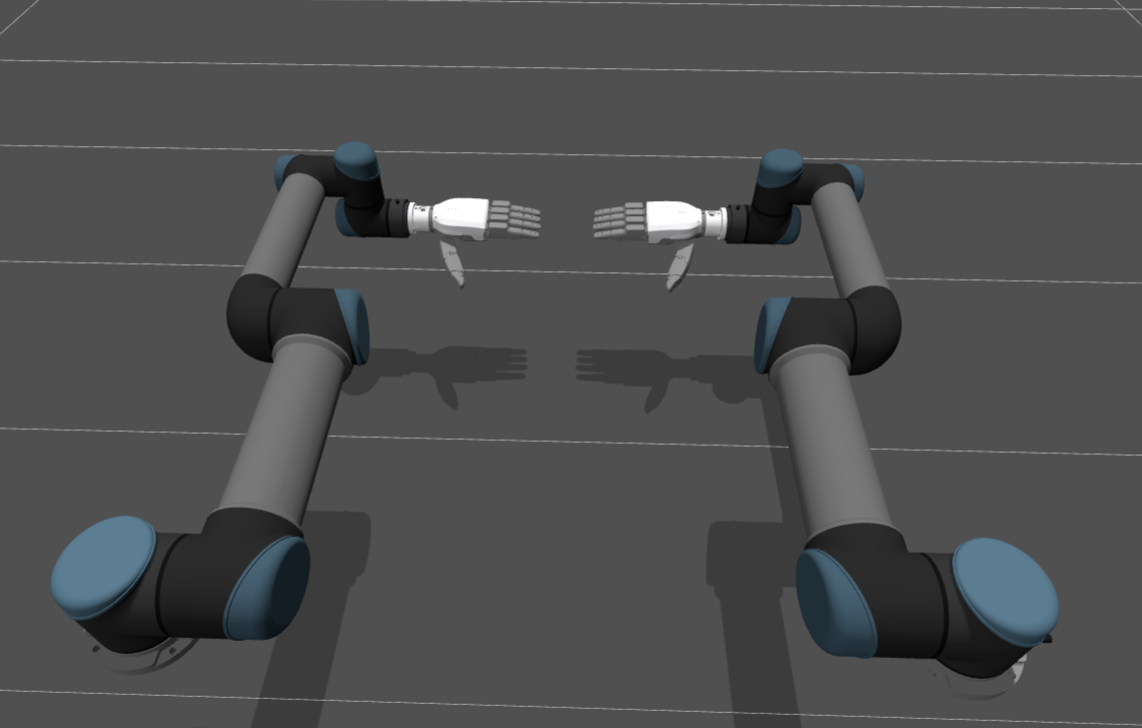}
{A metallic-gray UR5-style arm with blue circular joint caps, attached to a white-and-gray hand with a rounded palm shell, dark cylindrical finger coverings, and a thick side thumb close to the palm plane.}

\armbankrow{UR5+Schunk}
{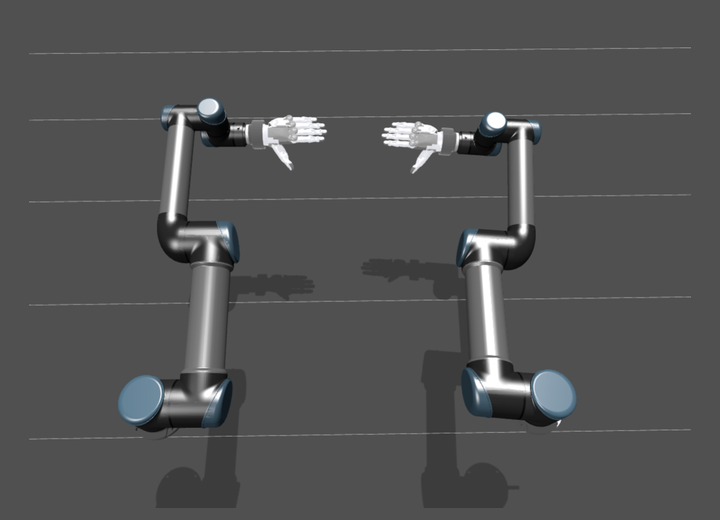}
{A metallic-gray arm with blue round joint covers, paired with a robust white industrial hand. A broad palm, thick segmented fingers, gray fingertip pads, and heavy thumb create a sturdy precision-oriented look.}

\armbankrow{UR5+Shadow}
{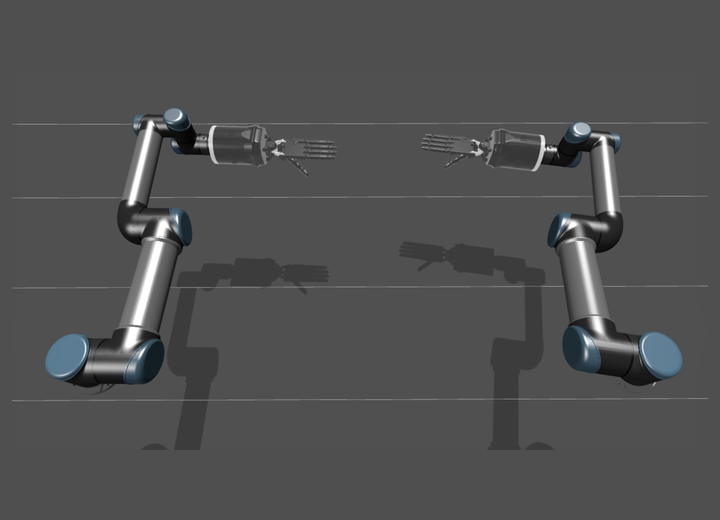}
{A metallic-gray arm with blue circular joints, attached to a dark anthropomorphic hand with slim multi-joint fingers and bright fingertip caps. The hand is lightweight and human-like relative to the heavier arm.}

\armbankrow{UR5+Wuji}
{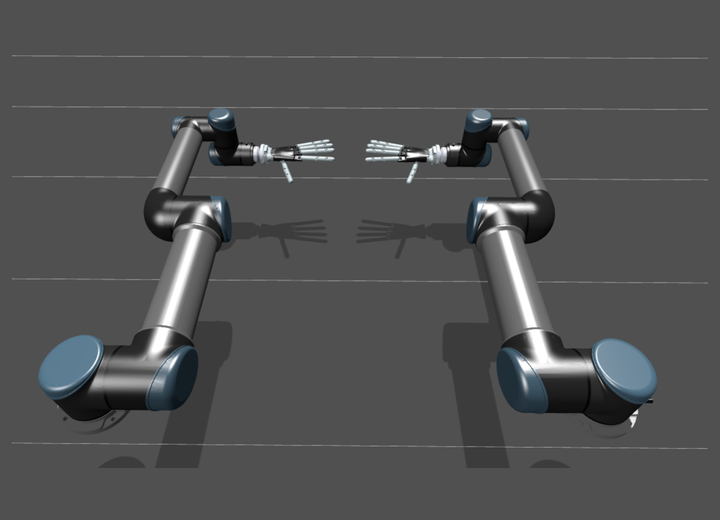}
{A metallic-gray arm with blue circular joint caps, ending in a slim dexterous hand with long narrow fingers and a thin side thumb. The hand appears lighter and more elongated than other UR5-based embodiments.}

\armbankrow{xArm7+Ability}
{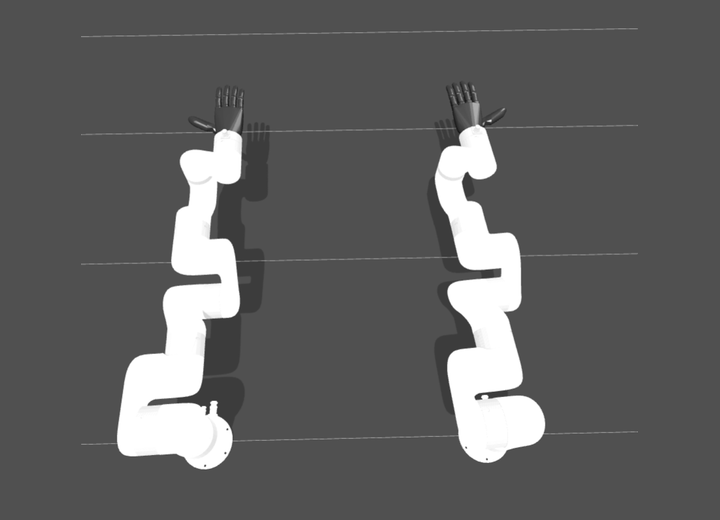}
{A white arm with smooth enclosed links, paired with a compact white hand with a rounded palm shell, dark finger coverings, and a thick side thumb. The embodiment is soft-contoured and highly enclosed.}

\armbankrow{xArm7+LEAP}
{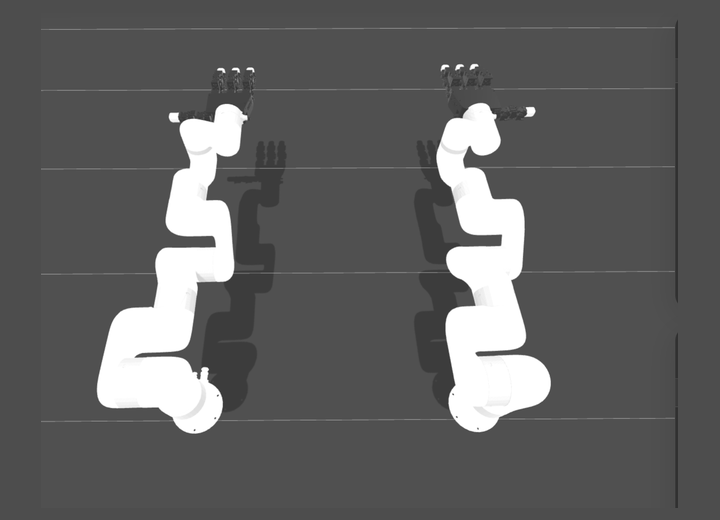}
{A white arm with smooth rounded links, attached to a compact low-profile robotic hand with a blocky palm and modular fingers. Stacked box-like segments and bright caps give it a clean geometric appearance.}

\end{longtable}
}

% Requires: \usepackage{booktabs,tabularx,longtable,array,graphicx,xcolor}
\section{Unified Tactile Surface Reconstruction}
\label{app:tacmap_surface_reconstruction}

Bench2Dex reconstructs tactile contact surfaces for the twelve bimanual hand embodiments included in the benchmark. The representation builds on image-based tactile simulation~\citep{tacto2022,taxim2021} and the geometry-consistent penetration-depth formulation of TacMap~\citep{tacmap2026}, while extending the surface-map abstraction to heterogeneous hand morphologies. For each hand, we identify the physical regions that should generate tactile evidence---elastomer pads, rubber pads, fingertip shells, force-sensor covers, or dedicated touch links---and rebuild or clean them in Blender. The cleaned meshes are then converted into surface point-and-normal assets so that the tactile signal is defined on the intended contact surface rather than on arbitrary full-link geometry.
The reconstruction follows a common convention across embodiments. Each surface mesh is expressed in the tactile sensor attach-link frame, or converted from the original URDF visual or collision geometry using the corresponding geometry origin, rotation, and mesh scale. The mesh is rasterized into a regular surface-aligned tactile grid, producing point and inward-facing normal maps at the native resolution used by the tactile registry, so that runtime rays are cast from the nominal contact surface toward the finger interior. The generated points are stored in millimeters and converted back to meters during sensing, keeping offline asset generation and runtime ray-casting consistent.

Different hands expose different mesh sharing patterns: some use a shared non-thumb surface for all non-thumb fingers, while others require independent assets per finger because the pad geometry varies. Table~\ref{tab:tacmap_surface_grouping} summarizes the resulting grouping policy. The current 12-hand setup uses a four-finger-shared grouping for nine hands and a per-finger independent grouping for three hands. Each hand places one tactile site on every finger, so the ten five-finger embodiments carry ten sites each, while the four-finger Allegro and LEAP embodiments carry eight; palm sites are excluded. In the released data, each site is sampled at the native surface-map resolution of $240\times240$ (subsampling step 1) and captured once per simulation step during replay.

\begin{table}[htbp]
\caption{Tactile surface-map grouping policy for the twelve Bench2Dex hands. ``4F'' denotes a shared non-thumb surface group.}
\label{tab:tacmap_surface_grouping}
\centering
\small
\setlength{\tabcolsep}{4pt}
\renewcommand{\arraystretch}{1.12}
\begin{tabularx}{\linewidth}{>{\raggedright\arraybackslash}p{0.24\linewidth} >{\raggedright\arraybackslash}p{0.28\linewidth} >{\raggedright\arraybackslash}X}
\toprule
Grouping policy & Hands & Surface-map asset groups \\
\midrule
Legacy side-shared thumb/4F & Sharpa & Two assets are shared across both hands: TH for thumbs and 4F for all non-thumb fingers. This preserves the original Sharpa TacMap asset convention. \\
Side-specific thumb/4F shared & RH5DG2, Shadow, Schunk, Wuji, Allegro, Orca, LEAP, DexHand021 & Four assets are used per embodiment: RTH, R4F, LTH, and L4F. The non-thumb group uses one representative surface per side. Allegro and LEAP are four-finger hands, so 4F covers index, middle, and ring. \\
Per-finger independent & RH56DFX, Ability, Revo2 & Each finger has its own surface-map group on each side, such as RTH, RIDX, RMID, RRING, RLIT and the corresponding left-hand groups. This is used when non-thumb pad meshes are not safely interchangeable. \\
\bottomrule
\end{tabularx}
\end{table}

\setlength{\LTcapwidth}{\textwidth}
\setlength{\LTleft}{0pt}
\setlength{\LTright}{0pt}

{\scriptsize
\setlength{\tabcolsep}{3pt}
\renewcommand{\arraystretch}{1.13}
\begin{longtable}{@{}
>{\raggedright\arraybackslash}p{0.105\linewidth}
>{\raggedright\arraybackslash}p{0.125\linewidth}
>{\raggedright\arraybackslash}p{0.205\linewidth}
>{\raggedright\arraybackslash}p{0.285\linewidth}
>{\raggedright\arraybackslash}p{0.205\linewidth}
@{}}
\caption{Tactile contact-surface mesh checklist used for Blender reconstruction. The mesh column lists only source filenames rather than full dataset paths.}
\label{tab:tacmap_blender_mesh_checklist}\\
\toprule
Hand & Grouping & Source links for reconstruction & Mesh source used in Blender & Notes \\
\midrule
\endfirsthead
\caption[]{Tactile contact-surface mesh checklist used for Blender reconstruction (continued).}\\
\toprule
Hand & Grouping & Source links for reconstruction & Mesh source used in Blender & Notes \\
\midrule
\endhead
\midrule
\multicolumn{5}{r}{\footnotesize Continued on next page} \\
\endfoot
\bottomrule
\endlastfoot

DexHand021 & Side-specific thumb/4F & Thumb: r\_f\_link1\_4 and l\_f\_link1\_4. 4F representative: r\_f\_link2\_4 and l\_f\_link2\_4. & r\_f\_link1\_4.STL; l\_f\_link1\_4.STL; r\_f\_link2\_4.STL; l\_f\_link2\_4.STL. & The source is the fourth link of each finger. The current 4F group uses finger 2 as the representative non-thumb surface. \\
Sharpa & Legacy side-shared thumb/4F & Thumb representative: right\_thumb\_elastomer. 4F representative: right\_index\_elastomer. & thumb\_elastomer\_surface.STL; elastomer\_surface.STL. & Preserves the original TH/4F naming. The legacy assets store outward-facing normals, which are flipped at runtime so that all hands share the inward-facing normal convention. \\
Allegro & Side-specific thumb/4F & Thumb: multi\_link\_15.0\_tip and link\_15.0\_tip. 4F representative: multi\_link\_3.0\_tip and link\_3.0\_tip. & link\_tip.obj. & All fingertips reuse the same tip mesh. The Blender export keeps the fingertip offset convention, with side-specific sensor normals. \\
Orca & Side-specific thumb/4F & Thumb: multi\_right\_thumb\_dp and left\_thumb\_dp. 4F representative: multi\_right\_index\_ip and left\_index\_ip. & right\_collision\_thumb\_dp\_skin\_mesh.stl; left\_collision\_thumb\_dp\_skin\_mesh.stl; right\_collision\_index\_ip\_skin\_mesh.stl; left\_collision\_index\_ip\_skin\_mesh.stl. & Uses collision skin geometry rather than visual body geometry. Left and right origins/rotations differ, so side-specific meshes are required. \\
Revo2 & Per-finger independent & Right and left thumb\_touch\_link, index\_touch\_link, middle\_touch\_link, ring\_touch\_link, and pinky\_touch\_link. & right\_thumb\_touch\_link.STL; right\_index\_touch\_link.STL; right\_middle\_touch\_link.STL; right\_ring\_touch\_link.STL; right\_pinky\_touch\_link.STL; left-hand equivalents. & Each touch link is a tactile surface. The non-thumb meshes differ enough that a shared 4F map would lose edge fidelity. \\
RH56DFX & Per-finger independent & Right and left thumb\_rubber\_3, index\_rubber\_2, middle\_rubber\_2, ring\_rubber\_2, and little\_rubber\_2. & right\_thumb\_rubber\_3.STL; right\_index\_rubber\_2.STL; right\_middle\_rubber\_2.STL; right\_ring\_rubber\_2.STL; right\_little\_rubber\_2.STL; left-hand equivalents. & Rubber pad dimensions differ across fingers, especially the little finger, so each finger is reconstructed separately. \\
RH5DG2 & Side-specific thumb/4F & Thumb: right\_thumb\_force\_sensor and left\_thumb\_force\_sensor. 4F representative: right\_index\_force\_sensor and left\_index\_force\_sensor. & right\_thumb\_force\_sensor.STL; left\_thumb\_force\_sensor.STL; right\_index\_force\_sensor.STL; left\_index\_force\_sensor.STL. & The force-sensor meshes are used as the contact pads. The non-thumb force-sensor geometry is treated as shareable within each side. \\
Schunk & Side-specific thumb/4F & Thumb: right\_hand\_c and left\_hand\_c. 4F representative: right\_hand\_t and left\_hand\_t. & d13.obj; d13\_left.obj; finger\_tip.obj. & Site-body overrides move the tactile sites from distal bodies to fingertip bodies. Non-thumb fingers reuse the same fingertip mesh. \\
Shadow & Side-specific thumb/4F & Thumb: thdistal and l\_thdistal. 4F representative: ffdistal and l\_ffdistal. & th\_distal\_pst.obj; f\_distal\_pst.obj. & The Shadow URDF applies a 0.001 mesh scale. The non-thumb distal mesh is shared across the four non-thumb fingers. \\
Wuji & Side-specific thumb/4F & Thumb: right\_finger1\_tip\_link and left\_finger1\_tip\_link. 4F representative: right\_finger2\_tip\_link and left\_finger2\_tip\_link. & right\_finger1\_tip\_link.STL; left\_finger1\_tip\_link.STL; right\_finger2\_tip\_link.STL; left\_finger2\_tip\_link.STL. & Tactile sites attach to the true fingertip links. Finger 2 represents the non-thumb group in the 4F-shared version. \\
Ability & Per-finger independent & Right and left thumb\_L2, index\_L2, middle\_L2, ring\_L2, and pinky\_L2. & thumb\_F2\_right.STL (right thumb); thumb\_F2\_left.STL (left thumb); idx\_F2\_Lg.STL (index, middle, ring, and left pinky); idx\_F2.STL (right pinky). & The right-hand pinky source mesh is smaller than the shared non-thumb mesh, so Ability uses per-finger surface-map groups. \\
LEAP & Side-specific thumb/4F & Thumb: thumb\_fingertip. 4F representative: fingertip. & thumb\_fingertip.obj; fingertip.obj. & LEAP is a four-finger hand. The non-thumb fingertip mesh is shared, but its URDF visual origin must be preserved unless the Blender export is already in the attach-link frame. \\
\end{longtable}
}

Table~\ref{tab:tacmap_blender_mesh_checklist} lists the source meshes used for Blender reconstruction, and Table~\ref{tab:tacmap_surface_reconstruction} visualizes the resulting unified tactile surfaces for all twelve hands.

\newcommand{\benchdextacmapsurface}[1]{%
    \begin{minipage}[c][4.25cm][c]{8.9cm}
        \centering
        \setlength{\fboxsep}{0pt}%
        \fcolorbox{gray!35}{white}{%
            \includegraphics[width=8.6cm,height=4.0cm,keepaspectratio]{#1}%
        }%
    \end{minipage}%
}

\newcommand{\benchdextacmapsurfacerow}[2]{%
    \parbox[c][4.25cm][c]{3.2cm}{\centering\textbf{#1}} &
    \benchdextacmapsurface{#2} \\
}

{\footnotesize
\setlength{\tabcolsep}{5pt}
\renewcommand{\arraystretch}{1.05}
\begin{longtable}{@{}
>{\centering\arraybackslash}m{3.3cm}
>{\centering\arraybackslash}m{9.2cm}
@{}}
\caption{Reconstructed tactile contact surfaces for the twelve Bench2Dex hands.}
\label{tab:tacmap_surface_reconstruction}\\
\toprule
\textbf{Hand} & \textbf{Reconstructed tactile surface} \\
\midrule
\endfirsthead
\caption[]{Reconstructed tactile contact surfaces for the twelve Bench2Dex hands (continued).}\\
\toprule
\textbf{Hand} & \textbf{Reconstructed tactile surface} \\
\midrule
\endhead
\midrule
\multicolumn{2}{r}{\footnotesize Continued on next page} \\
\endfoot
\bottomrule
\endlastfoot

\benchdextacmapsurfacerow{DexHand021}{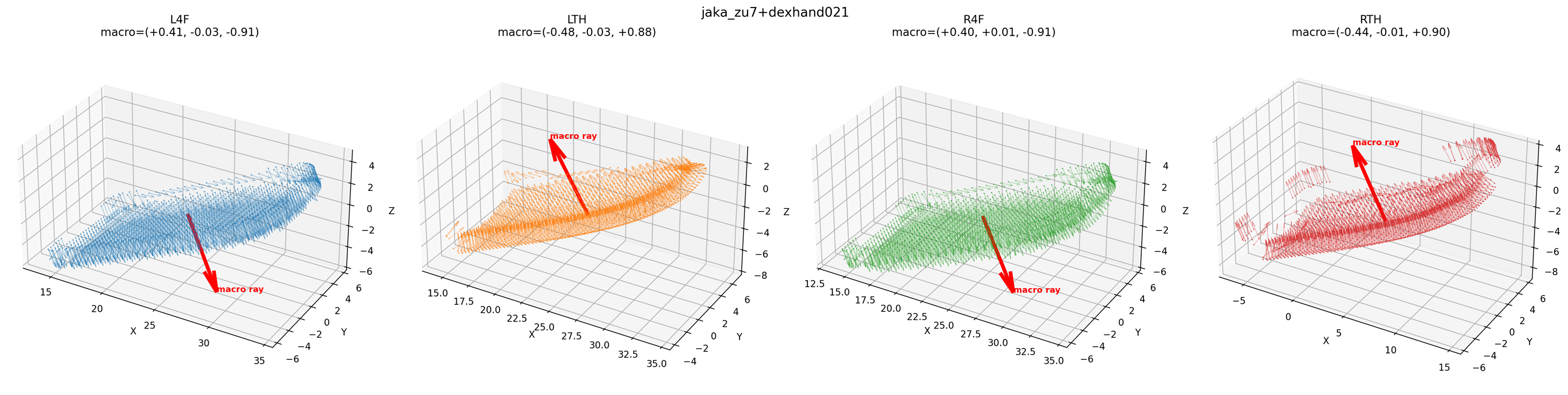}
\benchdextacmapsurfacerow{Sharpa}{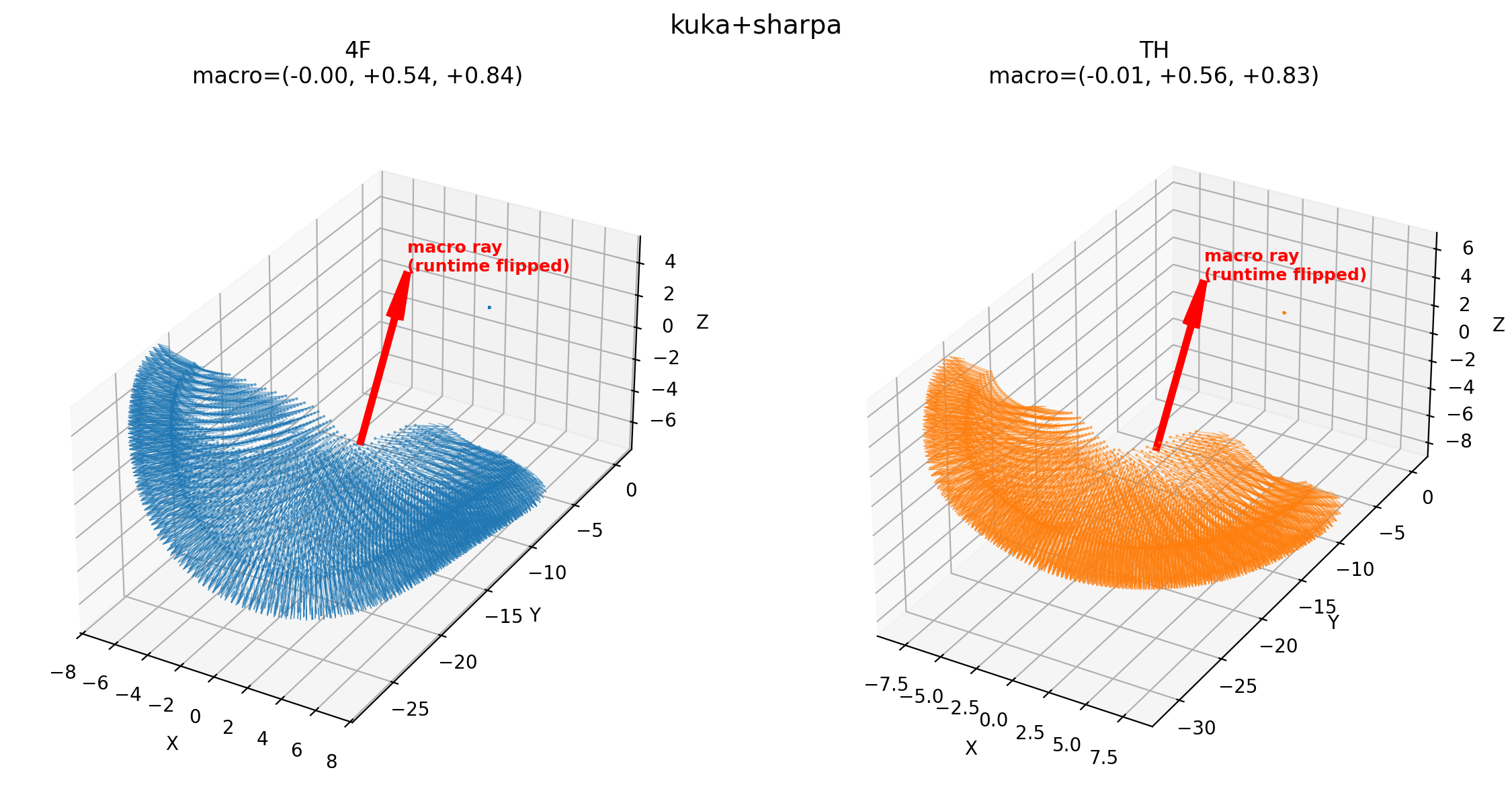}
\benchdextacmapsurfacerow{Allegro}{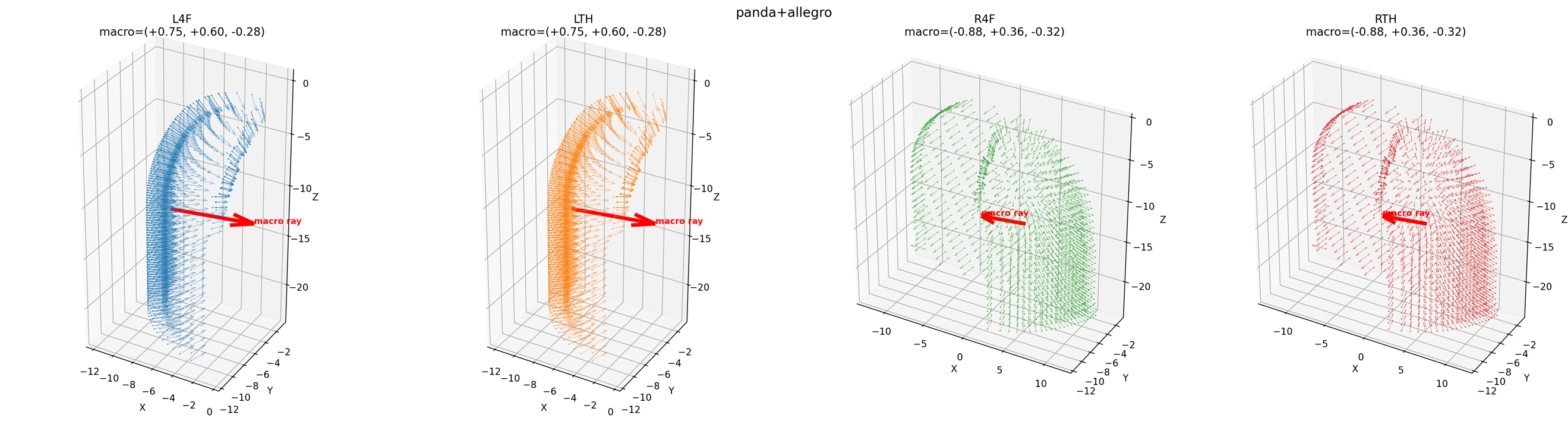}
\benchdextacmapsurfacerow{Orca}{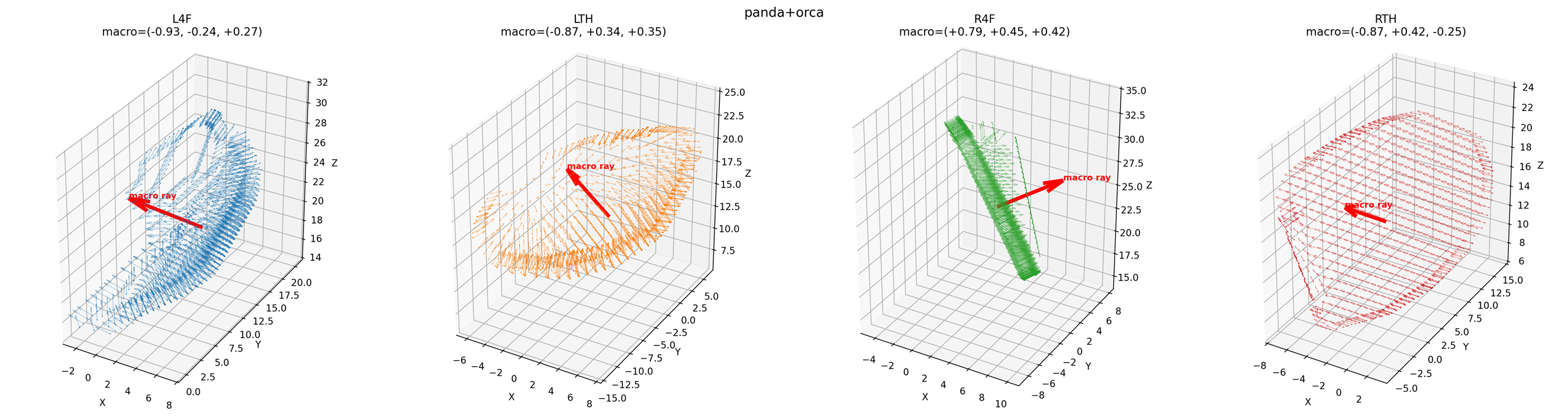}
\benchdextacmapsurfacerow{Revo2}{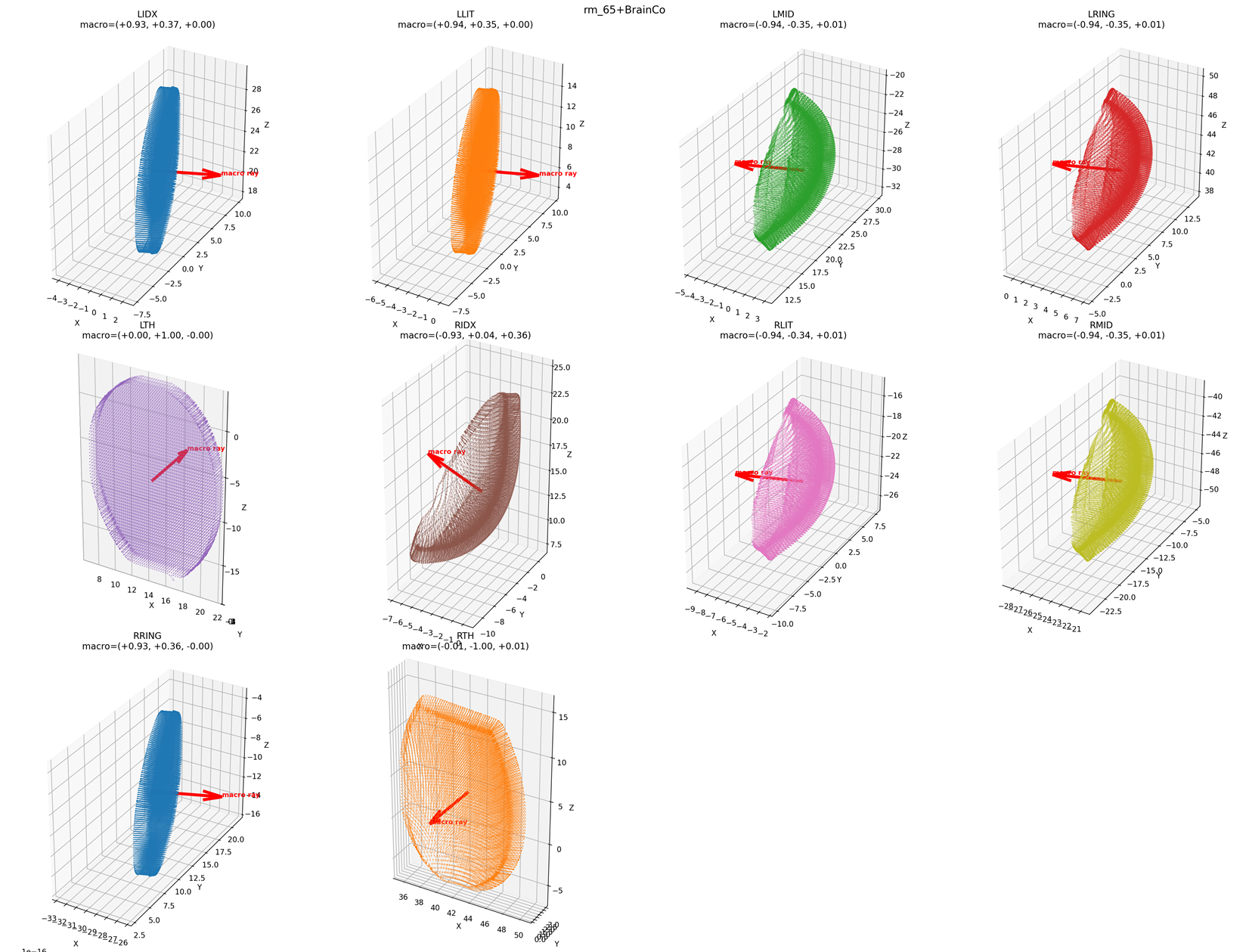}
\benchdextacmapsurfacerow{RH56DFX}{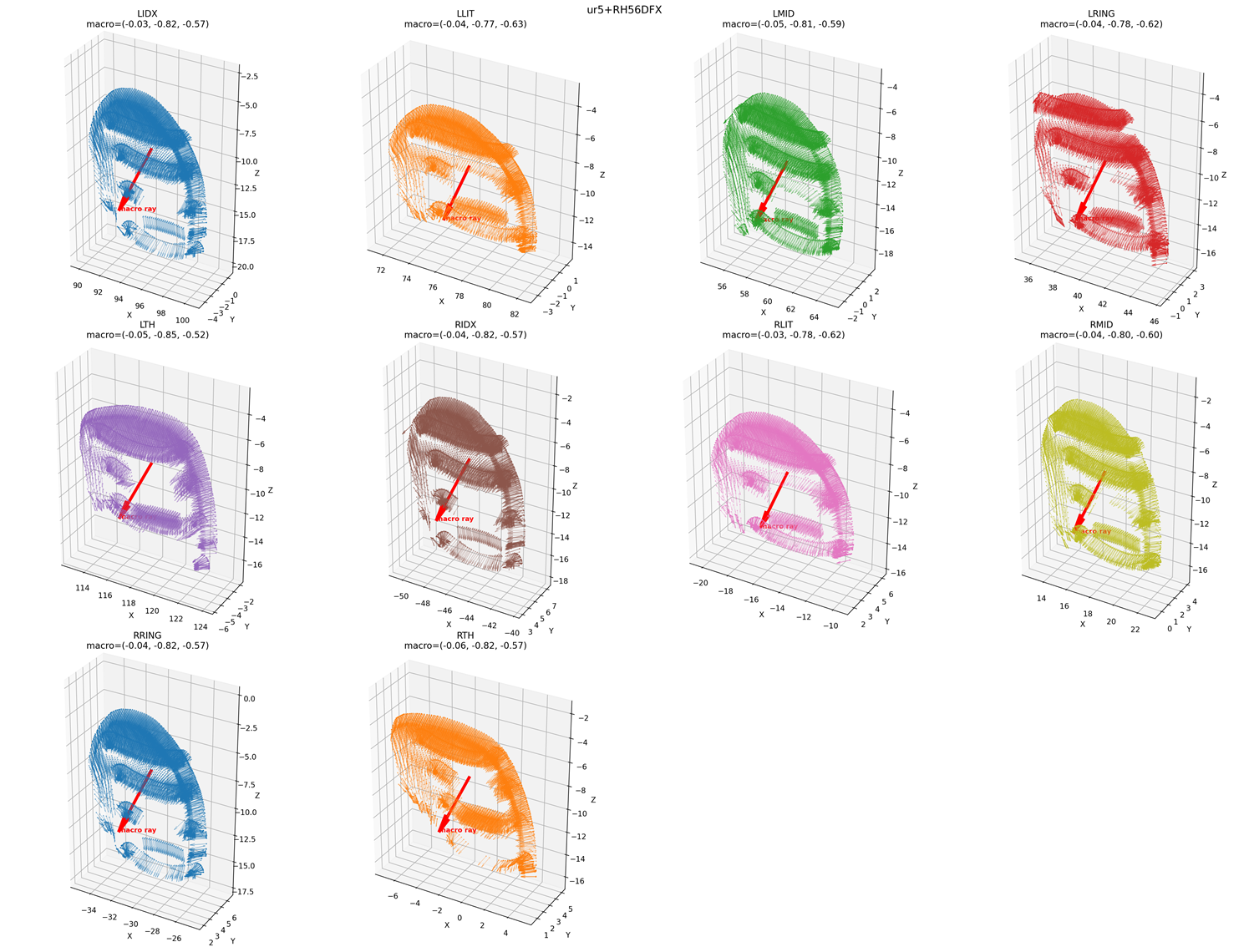}
\benchdextacmapsurfacerow{RH5DG2}{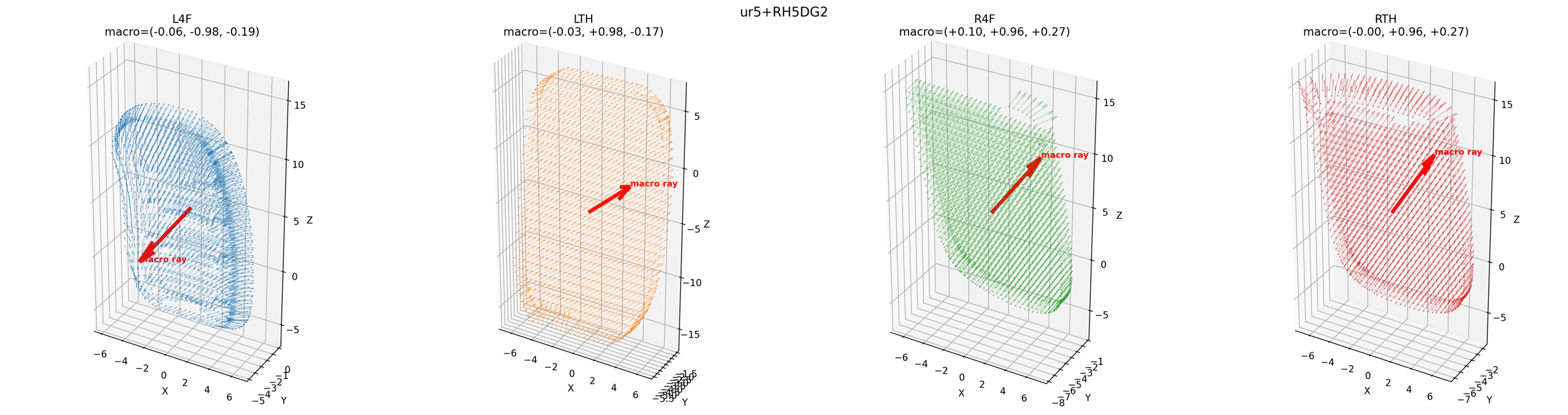}
\benchdextacmapsurfacerow{Schunk}{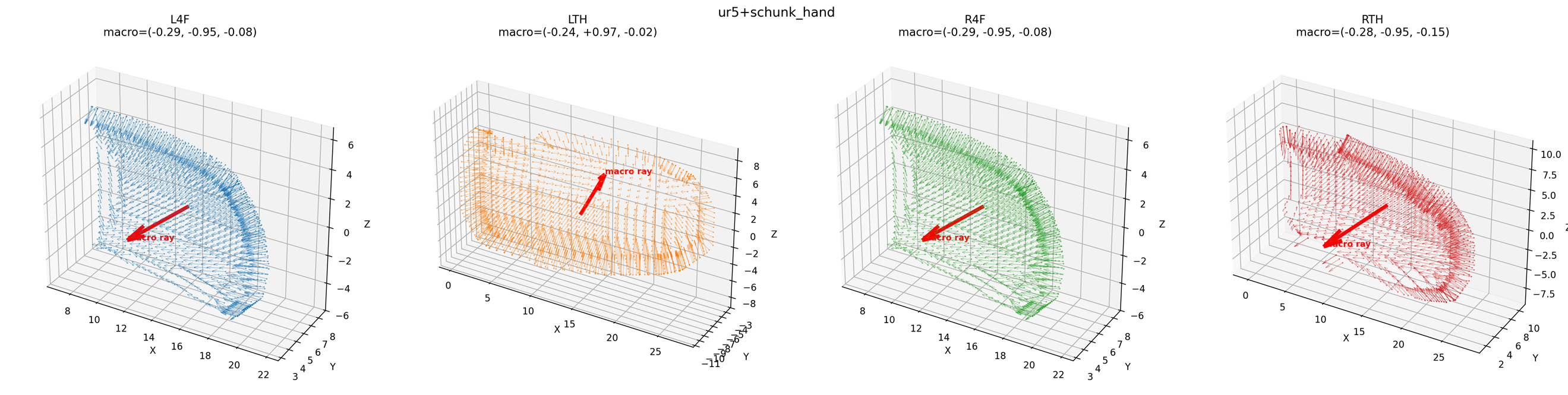}
\benchdextacmapsurfacerow{Shadow}{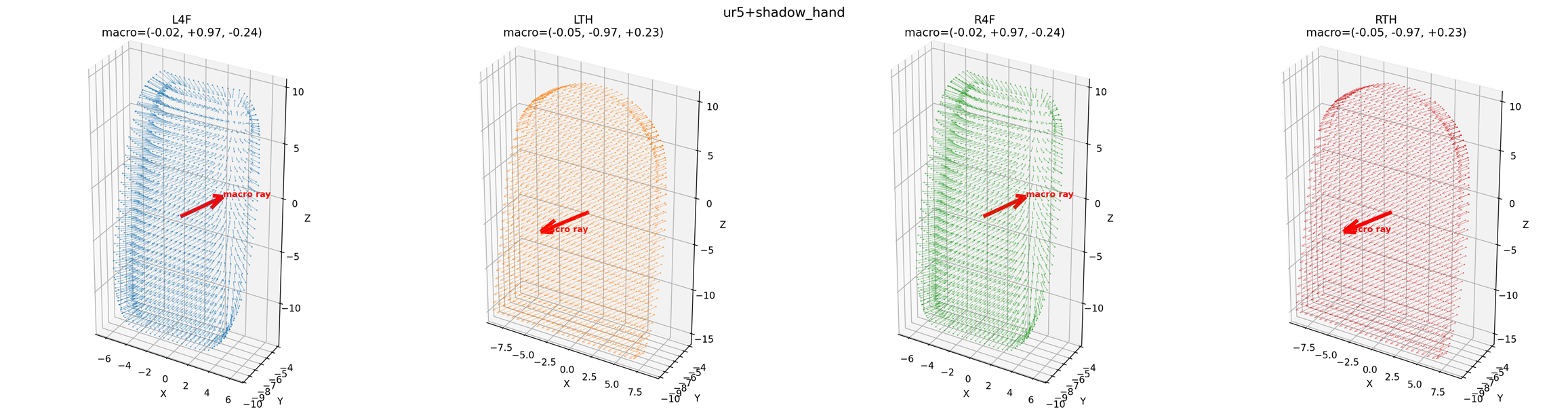}
\benchdextacmapsurfacerow{Wuji}{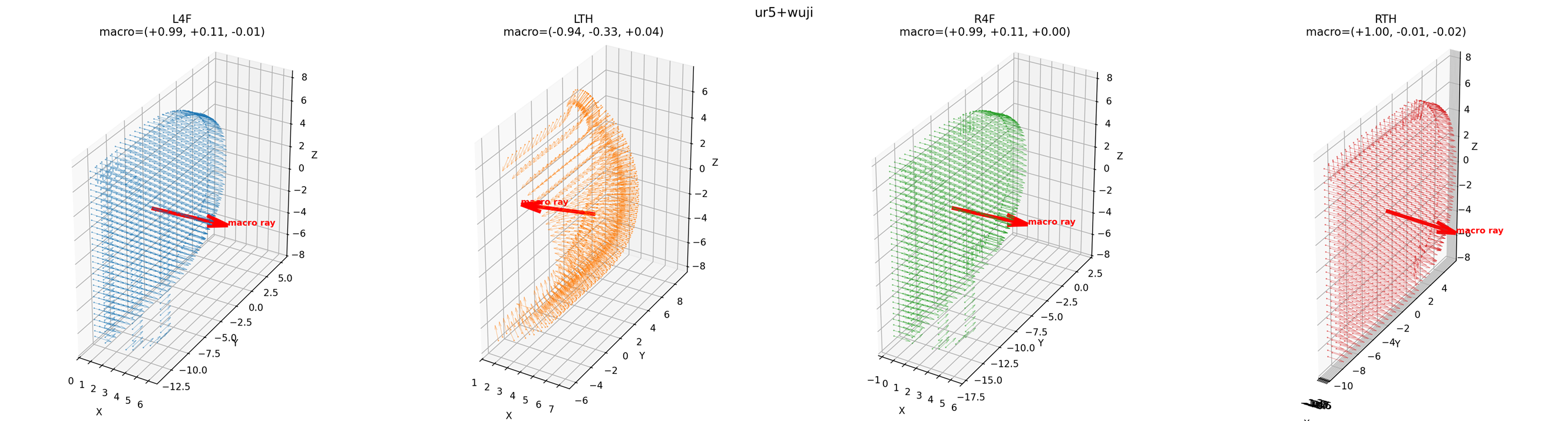}
\benchdextacmapsurfacerow{Ability}{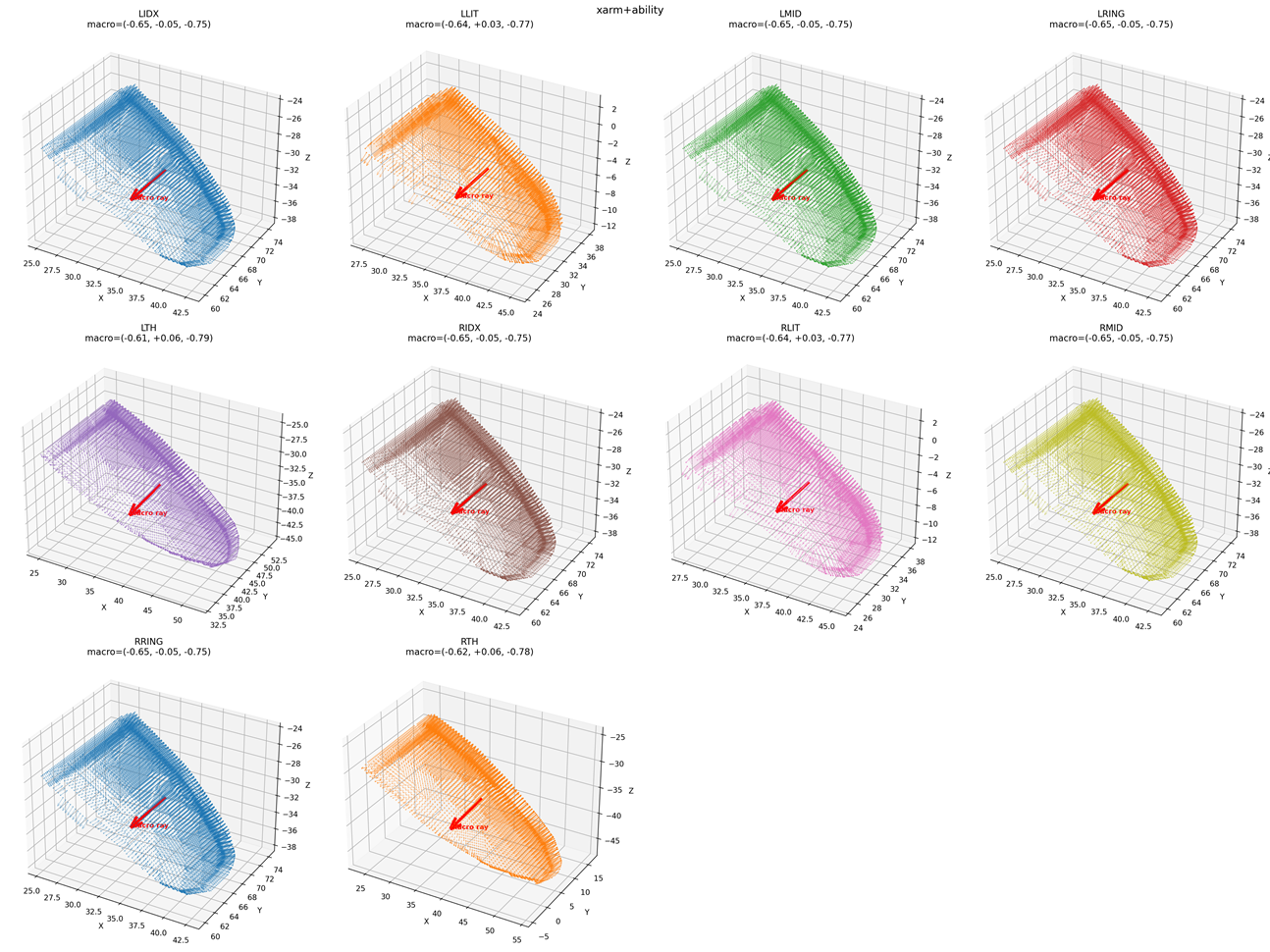}
\benchdextacmapsurfacerow{LEAP}{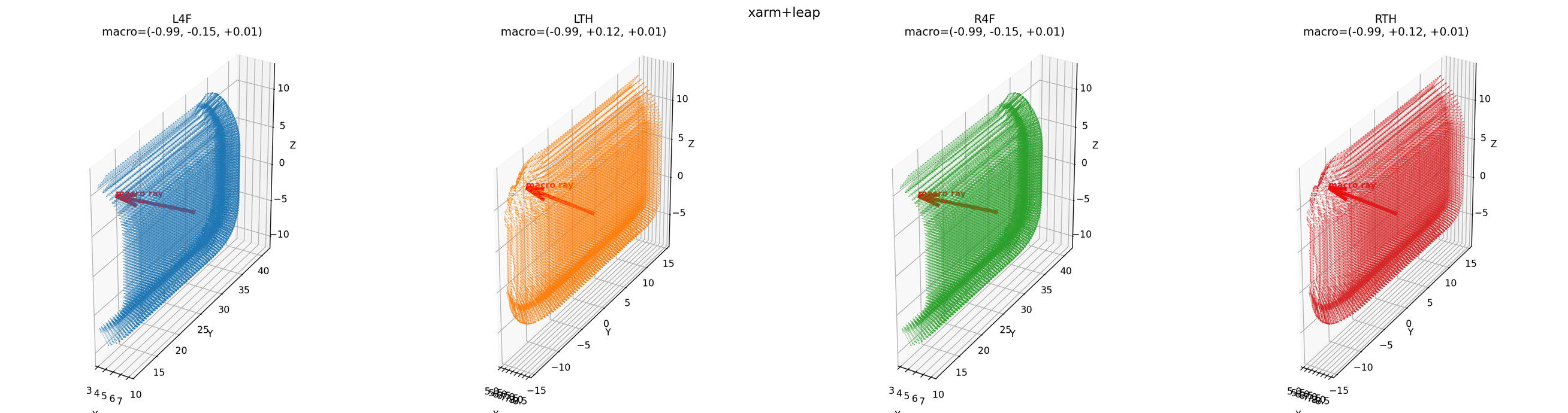}
\end{longtable}
}

\subsection{Signal Semantics and Depth Quantization}
\label{app:tacmap_quantization}

The raw value before quantization is the ray-cast distance along the inward-facing local surface normal: each ray is launched from a nominal surface point toward the finger interior, so the first-hit distance measures the depth to which an object surface has penetrated past the nominal contact surface, rather than the gap to an approaching but non-contacting object. Let $d_{s,t}(u,v)$ denote this metric distance, in meters, at tactile pixel $(u,v)$ for site $s$ and frame $t$. The contact-consistency check is a two-pass test: after the first hit, the ray origin is advanced to just short of the hit point and a reverse ray is cast; the sample is kept only when the reverse ray also intersects the target mesh, which holds once the object surface has crossed the ray's nominal surface origin. Rays are cast against the meshes of the dynamic task objects and, for articulated task objects, against their mesh-bearing rigid links, within a maximum cast range of $15$~mm; static scene geometry such as the tabletop is not among the raycast targets and produces no tactile response. Missed rays (no target hit within this range) and samples that fail the check are assigned zero distance and are additionally recorded by a binary contact-validity mask. For valid samples, the distance is first converted to millimeters,
\begin{equation}
    z_{s,t}(u,v) = 10^3 d_{s,t}(u,v),
\end{equation}
and then mapped to an 8-bit integer through the piecewise quantizer used by the released TacMap implementation~\citep{tacmap2026},
\begin{equation}
    q(z) =
    \begin{cases}
        z / 0.005, & z < 0.5, \\
        (z - 0.5) / 0.03 + 100, & z \ge 0.5,
    \end{cases}
\end{equation}
followed by clipping and casting,
\begin{equation}
    \widetilde{\mathbf{T}}_{s,t}(u,v)
    = \operatorname{uint8}\!\left(
        \operatorname{clip}\!\left(q(z_{s,t}(u,v)), 0, 255\right)
      \right).
\end{equation}
The quantized map is then spatially smoothed with a Gaussian kernel evaluated in floating point. The filtered response is rounded and cast back to an 8-bit integer before being stored in the tactile stream,
\begin{equation}
    \mathbf{T}_{s,t}
    = \operatorname{uint8}\!\left(
        \operatorname{round}\!\left(
            \mathcal{G}_{\sigma} * \widetilde{\mathbf{T}}_{s,t}
        \right)
      \right),
\end{equation}
where $\mathcal{G}_{\sigma}$ denotes Gaussian filtering and the current implementation uses $\sigma=1.5$; the kernel size is $\max(1,\lfloor 9/\mathrm{step}\rfloor)$ rounded up to an odd integer, which gives $9\times9$ at the released subsampling step of 1. The first branch allocates finer precision to small contact depths, using $0.005$ mm per integer level for $z<0.5$ mm, while the second branch uses a coarser $0.03$ mm per level for larger geometric penetration-depth values. Values of $5.15$~mm and above saturate at 255. Thus, each tactile-map pixel preserves high sensitivity near initial contact while remaining compact as a single 8-bit value in the HDF5 tactile stream.

\section{HDF5 Schema Details}
\label{app:hdf5_details}

\begin{table*}[t]
\caption{Eight data modalities recorded by Bench2Dex.}
\label{tab:modalities}
\resizebox{1\textwidth}{!}{
\begin{tabular}{l|l}
\toprule
Modality & Description \\
\midrule
RGB images & Multi-camera visual observations for policy learning and video analysis. \\
Depth maps & Geometric observations for 3D perception and reconstruction. \\
Robot states & Joint positions, velocities, efforts, and proprioceptive state information. \\
Object states & Object poses, velocities, and task-relevant physical states. \\
Tactile signals & Unified surface-aligned tactile images defined on robot hand contact surfaces. \\
2D boxes & Image-space object annotations for visual detection supervision. \\
3D boxes & Object-level 3D bounding boxes for geometric supervision. \\
Occupancy & TSDF or mesh-based occupancy labels for 3D scene understanding. \\
\bottomrule
\end{tabular}
}
\end{table*}

\begin{table*}[htbp]
\centering
\caption{Full per-episode HDF5 organization used by Bench2Dex when all eight modalities are collected. $T$ denotes the number of recorded frames, $J$ the robot joint dimension, $H \times W$ the camera resolution, and $H_t \times W_t$ the surface-aligned tactile resolution ($240\times240$ in the released data).}
\label{tab:hdf5_schema}
\vspace{2pt}
\begingroup
\scriptsize
\setlength{\tabcolsep}{3.5pt}
\renewcommand{\arraystretch}{0.94}
\newcommand{\benchhdfnode}[2]{\hspace*{#1em}\texttt{#2}}
\newcommand{\benchhdfgroup}[2]{\rowcolor{gray!10}\hspace*{#1em}\textbf{\texttt{#2}}}
\begin{tabularx}{\linewidth}{@{}>{\raggedright\arraybackslash}p{0.39\linewidth}>{\raggedright\arraybackslash}X@{}}
\toprule
\rowcolor{gray!18}
\textbf{HDF5 node} & \textbf{Stored content} \\
\midrule
\benchhdfgroup{0}{episode\_\{id\}.hdf5} & One synchronized trajectory episode. The parent directory also stores an \texttt{episode\_manifest.json} index. \\
\benchhdfnode{1}{meta/} & Task, scene, robot key, modality list, camera definitions, collection config, instruction, success flag, schema version, and dataset metadata. \\
\benchhdfnode{1}{time/} & \texttt{frame\_index}, \texttt{timestamp\_ns}, and \texttt{sim\_step} for temporal alignment across all modalities. \\
\benchhdfnode{1}{episode/} & Episode flags: \texttt{is\_first}, \texttt{is\_last}, \texttt{done}, and \texttt{success}. \\
\benchhdfnode{1}{action/} & Commanded robot action, action validity, action source, joint/action names, control mode, and action type. \\
\benchhdfnode{1}{frame\_valid, frame\_errors} & Per-frame validity flags and error messages for filtering corrupted or partially captured frames. \\
\addlinespace[1.5pt]
\benchhdfgroup{1}{cameras/\{camera\_id\}/} & Per-camera visual observations and calibration metadata. \\
\benchhdfnode{2}{rgb} & RGB images, shape $[T,H,W,3]$ when stored as dense images. \textit{Modality: RGB images}. \\
\benchhdfnode{2}{depth\_m, depth\_semantics} & Metric depth maps and depth convention, shape $[T,H,W]$. \textit{Modality: depth maps}. \\
\benchhdfnode{2}{intrinsic, extrinsic\_world\_from\_cam} & Camera intrinsics and per-frame world-from-camera extrinsics. \\
\benchhdfnode{2}{camera\_model, fisheye calibration} & Camera model, clipping range, and optional fisheye matrix, distortion coefficients, and valid mask. \\
\addlinespace[1.5pt]
\benchhdfgroup{1}{robot/} & Robot proprioception and tactile observations. \\
\benchhdfnode{2}{joint\_names, qpos, qvel, qeffort} & Joint names and per-frame joint position, velocity, and effort, shape $[T,J]$. \textit{Modality: robot states}. \\
\benchhdfnode{2}{tactile/meta} & Surface-aligned tactile metadata, including the tactile site list and sensor settings. \\
\benchhdfnode{2}{tactile/tacmap/\{site\_name\}} & Quantized surface-aligned tactile images, shape $[T,H_t,W_t]$, uint8. \textit{Modality: tactile signals}. \\
\benchhdfnode{2}{tactile/distance\_along\_normal\_m} & Group of per-site raw metric ray-depth datasets (float32, meters), one dataset per tactile site, each of shape $[T,H_t,W_t]$. \\
\benchhdfnode{2}{tactile/contact\_mask} & Group of per-site binary contact-validity datasets (bool), one dataset per tactile site, each of shape $[T,H_t,W_t]$; true where the sample passed the contact-consistency check. \\
\addlinespace[1.5pt]
\benchhdfgroup{1}{objects/\{obj\_id\}/} & Per-object physical state traces. \\
\benchhdfnode{2}{pose\_world} & Object pose in world frame, shape $[T,7]$ for position and quaternion. \textit{Modality: object states}. \\
\benchhdfnode{2}{lin\_vel\_world, ang\_vel\_world} & Object linear and angular velocity traces, shape $[T,3]$. \\
\benchhdfnode{2}{joint\_names, qpos, qvel, qeffort} & Optional articulated-object joint state fields. \\
\addlinespace[1.5pt]
\rowcolor{gray!10}
\benchhdfnode{1}{labels/box2d/\{camera\_id\}/\{obj\_id\}/} & Image-space boxes \texttt{xyxy} and visibility flags, shapes $[T,4]$ and $[T]$. \textit{Modality: 2D boxes}. \\
\benchhdfnode{1}{labels/box3d/\{obj\_id\}/} & Oriented 3D boxes: \texttt{center\_world}, \texttt{size\_lwh}, and \texttt{quat\_world}. \textit{Modality: 3D boxes}. \\
\benchhdfnode{1}{labels/occupancy\_tsdf/} & Occupancy grid state, grid shape, bounds, voxel size, method, semantics, and frame-valid flags. \textit{Modality: occupancy}. \\
\addlinespace[1.5pt]
\rowcolor{gray!10}
\benchhdfnode{1}{metrics/episode} & Episode-level task progress, success, efficiency, and safety metrics when available. \\
\benchhdfnode{1}{metrics/timeseries} & Per-frame or per-step metric traces aligned with the episode timeline. \\
\bottomrule
\end{tabularx}
\endgroup
\end{table*}

\section{Evaluation Metrics}
\label{app:evaluation_metrics}

Bench2Dex uses a unified metric suite that separates core benchmark scores from auxiliary diagnostic signals. Each evaluated task defines executable terminal conditions, stage-level subgoals, safety rules, and, when applicable, grasp or tool-use diagnostics. Let $q$, $m$, and $p$ index the task, policy, and evaluation channel, respectively, and let $N$ denote the number of recorded episodes in a $(q,m,p)$ cell unless stated otherwise. Episode-level quantities are first summarized within each cell, and benchmark-level episode metrics are equal-weight means across tasks. For metrics defined as means over all $N$ episodes, this task-macro mean equals the corresponding pooled episode mean because every reported cell contains 50 episodes. Unless noted otherwise, the benchmark uses the reach-and-stop protocol; fixed-horizon evaluation is treated as a distinct protocol and identified explicitly in the evaluation summary.

\subsection{Primary Metrics}

\paragraph{Metric set.}
The core reported metric set is
\begin{align}
\mathcal{M}_{\mathrm{core}}
=
\{
&\mathrm{SR},\ \mathrm{LSCR},\ \overline{t}_{\mathrm{succ}}, \nonumber\\
&\mathrm{SafeSR},\ \mathrm{HVR},\ \mathrm{DropR},\ \mathrm{HSR},\ \mathrm{VSR}_{\mathrm{step}},\ \mathrm{RR}
\}.
\end{align}
SR is the primary completion outcome; LSCR and SafeSR provide complementary views of progress and safety, while the remaining quantities characterize success-conditional efficiency, violations, and robustness. SR can be accompanied by a central 95\% confidence interval computed from 10,000 percentile-bootstrap resamples of the corresponding episode group with a fixed seed of 0.

\paragraph{Reach-and-stop success.}
In the reported evaluations, the evaluator is updated once per executed physics step. Let $C_{e,j}\in\{0,1\}$ be the terminal predicate at evaluator update $j$, $\delta t_{e,j}$ the elapsed simulation time since the preceding update, $n_e$ the number of executed updates, and $\Delta_e$ the configured dwell time. The accumulated dwell time is
\begin{align}
h_{e,0}&=0, &
h_{e,j}&=C_{e,j}\bigl(h_{e,j-1}+\delta t_{e,j}\bigr).
\end{align}
Define $j_e^{\mathrm{stable}}=\min\{j\leq n_e:h_{e,j}\geq\Delta_e\}$ when this set is nonempty, $S_e=\mathbf{1}[j_e^{\mathrm{stable}}\ \text{exists}]$, and $t_e^{\mathrm{stable}}=\sum_{r=1}^{j_e^{\mathrm{stable}}}\delta t_{e,r}$. The default dwell time is $0.5$~s. Under reach-and-stop, an episode terminates at stable success; otherwise it ends at the evaluation horizon or when evaluation cannot proceed. A recorded episode that terminates because of a policy or runtime evaluation error remains in the denominator and is counted as unsuccessful. The stable success rate is
\begin{align}
\mathrm{SR}=\frac{1}{N}\sum_{e=1}^{N} S_e.
\end{align}
This protocol avoids counting transient contacts or unstable placements as successful episodes.

\paragraph{Latched stage progress.}
Binary success is insufficient for long-horizon manipulation because a policy may complete early stages but fail later. Each task specifies a set of stages $\mathcal{G}_e=\{g_{e,k}\}_{k=1}^{K_e}$ with optional dependency edges. Let $L_{e,k}$ be one if the predicate of stage $k$ is true at some update $j\leq n_e$ and all dependencies were latched earlier or become valid in the same-step fixed-point closure over the task's acyclic dependency graph; otherwise let it be zero. Once set, $L_{e,k}$ remains one. The primary progress metric is
\begin{align}
\mathrm{LSCR}_e
=
\frac{1}{K_e}
\sum_{k=1}^{K_e}
L_{e,k}.
\end{align}
The benchmark defines aggregate LSCR as the mean of $\mathrm{LSCR}_e$ over all evaluated episodes. It measures ever-reached, dependency-valid milestones rather than final-state satisfaction; terminal-state stage status is retained as a diagnostic below.

\paragraph{Completion time.}
Completion time is reported as the success-conditional mean
\begin{align}
\overline{t}_{\mathrm{succ}}
=
\frac{\sum_{e=1}^{N} S_e t^{\mathrm{stable}}_e}{\sum_{e=1}^{N} S_e},
\end{align}
when at least one episode succeeds; otherwise it is undefined. Because this quantity conditions on success, it should be interpreted together with SR and the number of successful episodes. Expert-normalized speed and step-based execution counts are auxiliary diagnostics rather than primary ranking fields.

\paragraph{Safety.}
Safety is evaluated over executed physics steps. Let $D_e$ and $H_e$ denote whether episode $e$ contains a task-defined drop or tracked-object high-speed violation, respectively. The hard-violation indicator is $B_e=D_e\lor H_e$ by default; a joint-limit rule is included only when the task explicitly promotes it to the core safety criterion. The safe-success indicator is $S_e(1-B_e)$. Accordingly,
\begin{align}
\mathrm{SafeSR} &= \frac{1}{N}\sum_{e=1}^{N} S_e(1-B_e), \\
\mathrm{HVR} &= \frac{1}{N}\sum_{e=1}^{N} B_e.
\end{align}
The associated rates are $\mathrm{DropR}=N^{-1}\sum_e D_e$ and $\mathrm{HSR}=N^{-1}\sum_e H_e$. These events are not mutually exclusive, so their rates need not sum to HVR. If $V_{e,j}$ indicates any task-level violation at executed step $j$, the pooled violation-step rate is
\begin{align}
\mathrm{VSR}_{\mathrm{step}}
=
\frac{\sum_e\sum_{j=1}^{n_e} V_{e,j}}{\sum_e n_e}.
\end{align}
This is an exposure-normalized rate over executed steps and is interpreted together with episode-level HVR. High-speed events are severe-motion proxies, not contact-force or collision measurements. Joint-limit observations are treated as auxiliary diagnostics unless a task explicitly defines them as safety violations.

\paragraph{Robustness.}
For generalized evaluation, let $\mathcal{P}$ comprise the None, Equi., Inv., and Full evaluation channels. The benchmark provides each channel's SR, confidence interval, and sample count. For a shifted channel $p\in\mathcal{P}\setminus\{\mathrm{None}\}$, the relative ratio is
\begin{align}
\mathrm{RR}_{p}
=
\frac{\mathrm{SR}_{p}}
       {\mathrm{SR}_{\mathrm{None}}},
\end{align}
when $\mathrm{SR}_{\mathrm{None}}>0$. Let $\mathcal{P}_{+}=\mathcal{P}\setminus\{\mathrm{None}\}$ and $N_p$ be the number of episodes in channel $p$. The aggregate used by the benchmark is
\begin{align}
\mathrm{SR}_{\mathrm{shift}}
&=\frac{\sum_{p\in\mathcal{P}_{+}}N_p\mathrm{SR}_p}{\sum_{p\in\mathcal{P}_{+}}N_p}, &
\mathrm{RR}_{\mathrm{shift}}
&=\frac{\mathrm{SR}_{\mathrm{shift}}}{\mathrm{SR}_{\mathrm{None}}}.
\end{align}
RR is reported together with the absolute channel-wise SRs as a supporting robustness descriptor. These are grouped, unpaired comparisons rather than paired causal estimates.

\subsection{Diagnostic Metrics}

Diagnostic metrics are not used as primary ranking fields. They expose failure modes and execution characteristics that are not uniformly applicable or comparable across all tasks and embodiments. A quantity that is undefined for a task is marked as unavailable rather than assigned a value of zero.

\paragraph{Completion diagnostics.}
Let $I_e=\max_{j\leq n_e} C_{e,j}$ indicate whether the terminal predicate is satisfied at least once, without imposing the dwell-time requirement, and let $F_e=C_{e,n_e}$ indicate whether it is satisfied at the rollout's actual termination observation. The corresponding ever-instantaneous and terminal-state success rates are
\begin{align}
\mathrm{ISR}_{\mathrm{ever}} &= \frac{1}{N}\sum_{e=1}^{N} I_e, &
\mathrm{SR}_{\mathrm{final}} &= \frac{1}{N}\sum_{e=1}^{N} F_e.
\end{align}
For stage-level diagnosis, let $A_{e,k}(n_e)$ indicate that the dependencies of stage $k$ have been satisfied by termination. The terminal-state stage completion rate is
\begin{align}
\mathrm{CSCR}_e
=
\frac{1}{K_e}
\sum_{k=1}^{K_e}
\mathbf{1}\!\left[g_{e,k}(n_e)=1 \land A_{e,k}(n_e)=1\right].
\end{align}
Current and latched dependency-chain depths further distinguish terminal-state progress from progress reached at any earlier point. LSCR remains the primary stage-completion measure.

\paragraph{Efficiency and execution diagnostics.}
For a successful episode from task $q$ with a configured expert reference, task efficiency is the benchmark-specific speed ratio
\begin{align}
\mathrm{TE}_e = \frac{t^{\mathrm{expert}}_q}{t^{\mathrm{stable}}_e},
\qquad
t^{\mathrm{expert}}_q = H^{\mathrm{expert}}_q\,\delta t_q,
\end{align}
where $H^{\mathrm{expert}}_q$ is the task-level reference step count and $\delta t_q$ is its simulation integration interval. TE is unavailable for failures or tasks without a reference. This ratio is unbounded and is comparable only under matched simulation timing, initial-state distribution, terminal condition, dwell time, and evaluation horizon. Additional execution measures include the numbers of physics steps, policy-rate control steps, and policy queries required to reach stable success; the latter two differ for chunked policies.

\paragraph{Safety diagnostics.}
Safety diagnostics include violation-event density, per-type violation counts, hard joint-limit observations, finger-joint saturation, active-step rates, and maximum observed joint-limit excess. These quantities characterize kinematic constraint violations and control behavior, and they remain separate from the primary task-safety definition unless included in a task's safety criterion.

\paragraph{Grasp diagnostics.}
For contact-rich tasks, Bench2Dex defines a kinematic grasp stability index (GSI). Since reliable hand--object contact geometry and hand forward kinematics are not uniformly available for every supported embodiment, GSI is a proxy based on object lift, stable holding, object motion, hold duration, and slip events. For a configured tracked object $o\in\mathcal{O}^{\mathrm{grasp}}_e$ at step $j$, the canonical score is
\begin{align}
G_{o,j}
=
L_{o,j}\,H_{o,j}\,M_{o,j}\,Q_{o,j}\,A_{o,j},
\end{align}
where all components lie in $[0,1]$. Specifically, $L$ indicates lift relative to the initial object height; $H$ takes values $1$, $0.35$, $0.15$, or $0$ for held, lifted-and-stable, lifted-only, or other states; $M=\exp[-\tfrac{1}{2}(\|v\|/\sigma_v+\|\omega\|/\sigma_\omega)]$; $Q=0.75+0.25\min(1,d/d_{\min})$ when held and $1$ otherwise; and $A=1-\min(1,s/s_{\max})$. The thresholds and scales follow the task configuration or benchmark defaults. The episode-level grasp diagnostics are
\begin{align}
\mathrm{GSI}_e &= \max_{o\in\mathcal{O}^{\mathrm{grasp}}_e}\max_{j\leq n_e}G_{o,j},\\
\overline{\mathrm{GSI}}_e
&= \operatorname{mean}_{o,j: G_{o,j}>0} G_{o,j}.
\end{align}
The former records peak grasp quality, whereas the latter summarizes positive-score object--step observations.

\paragraph{Tool-use and motion diagnostics.}
Tool-use diagnostics are enabled only when stages specify tool IDs or tool equivalence classes. For episode $e$, tool selection accuracy measures whether the first held object selected for an eligible stage belongs to its allowed tool set:
\begin{align}
\mathrm{TSA}_e
=
\frac{\#\ \mathrm{correct\ stage\ tool\ selections}}
       {\#\ \mathrm{tool\ annotated\ stages}}.
\end{align}
An annotated stage with no selected tool is counted as incorrect. Tool switch success rate measures whether an attempted switch releases the previous tool stably and grasps the next tool stably before timeout:
\begin{align}
\mathrm{TSSR}_e
=
\frac{\#\ \mathrm{successful\ tool\ switches}}
       {\#\ \mathrm{attempted\ tool\ switches}}.
\end{align}
TSSR is unavailable for episodes without a switch attempt. Aggregate TSA and TSSR are means over episodes for which the corresponding quantity is defined, and TSSR is accompanied by total attempted and successful switch counts. Robot-motion diagnostics summarize joint velocity, acceleration, jerk, and effort by first computing the root mean square over available steps and joints within each episode and then averaging episode values. They are interpreted only under a shared embodiment and timing configuration. Execution summaries additionally provide the mean rollout length and the number of episodes that terminate because evaluation cannot proceed.

\subsection{Reproducibility Metadata}

Each evaluation summary is accompanied by the protocol required to interpret and reproduce its scores: policy identity, base seed and seed-derivation rule, number of episodes, evaluation-failure count, termination rule, maximum evaluation horizon, and dwell time. For randomized evaluation, the resolved episode seed and sampled generalization parameters are retained for each rollout. Together, these quantities support reproducible audit and reconstruction under a matched simulator, backend, task, and perturbation configuration.

\section{Policy Training Configurations}
\label{app:policy_training}

This section discloses the training configurations of the four policies evaluated
in Table~\ref{tab:policy_generalization_results}: ACT, DP, $\pi_{0.5}$, and
GR00T N1.5. ACT and DP are trained from scratch on Bench2Dex demonstrations;
$\pi_{0.5}$ and GR00T N1.5 are fine-tuned from their respective public
checkpoints. For $\pi_{0.5}$ and GR00T N1.5, whose pretrained action
heads assume lower-dimensional action spaces than the bimanual-dexterous
embodiments require, we retain the pretrained weights and randomly initialize
the additional output dimensions to match the target action space. Unless stated
otherwise, each policy is trained on a single task--embodiment setting and
consumes the synchronized multi-view RGB and proprioceptive observations
recorded in the unified HDF5 episodes. The default hyperparameters reported
below are those of the released training launchers
(\texttt{policy/<name>/train.sh}); per-task overrides, when applied, are logged
with the corresponding checkpoint.

\paragraph{ACT.}
We train ACT from scratch using its conditional variational autoencoder (cVAE)
formulation. The encoder and decoder are Transformer networks with a hidden
dimension of $512$ and a feed-forward dimension of $3{,}200$. Inputs are four
camera views (both wrist and both stereo cameras) encoded by a shared ResNet-18
backbone together with the normalized proprioceptive state, whose dimension is
sized to each embodiment's active degrees of freedom. The policy predicts an
action chunk of length $T_{\mathrm{pred}}=30$ from a single observation
($T_o=1$) and is supervised with an $\ell_1$ reconstruction loss regularized by
a KL term of weight $10$. We optimize with AdamW at a learning rate of
$1\times10^{-5}$ and a batch size of $32$ for up to $6{,}000$ epochs on a single
RTX~4090 GPU. At inference the policy re-plans at every step and fuses
overlapping chunk predictions by temporal ensembling.

\paragraph{Diffusion Policy (DP).}
We train DP from scratch using the image-conditioned 1D U-Net variant of
Diffusion Policy. Multi-view RGB observations are resized to $216\times288$ and
encoded independently by ResNet-18 visual backbones, whose features are fused
into the global conditioning vector of the denoiser. The denoiser is a 1D
conditional U-Net with feature widths $(256,512,1024)$, a $128$-dimensional
diffusion-step embedding, and kernel size $5$. At each control cycle the policy
conditions on the most recent $T_o=3$ observations, denoises a
$T_{\mathrm{pred}}=8$-step action trajectory, and executes the first
$T_{\mathrm{exec}}=6$ actions before re-planning from fresh observations,
yielding receding-horizon closed-loop control. Training adopts the DDPM
($\epsilon$-prediction) objective over $100$ diffusion timesteps under a
squared-cosine noise schedule, and $100$ denoising steps are taken at inference.
We optimize with AdamW at a peak learning rate of $1\times10^{-4}$, a global
batch size of $512$, $500$-step linear warmup followed by cosine decay, and an
EMA of the policy weights with decay $0.9999$. Each task-specific policy is
trained for up to $300$ epochs on eight H100 GPUs.

\paragraph{$\pi_{0.5}$.}
We fully fine-tune the pretrained $\pi_{0.5}$ vision--language--action model
using the openpi implementation, initializing from the public
\texttt{pi05\_base} checkpoint. The released
\texttt{pi05\_base\_dex2bench\_full} configuration trains the non-LoRA
PaliGemma-2B vision--language backbone and Gemma-300M action expert jointly.
The state and action dimensions are selected from the active degrees of freedom
of each embodiment. Compatible pretrained parameters are retained, while
state/action projection layers whose shapes differ from the pretrained
checkpoint remain randomly initialized. The policy conditions on four RGB
views (both stereo and both wrist cameras), normalized proprioception, and the
task instruction. It predicts an action chunk of
$T_{\mathrm{pred}}=20$ steps with a maximum token length of $280$; all
$T_{\mathrm{exec}}=20$ actions are executed before the policy re-observes and
re-plans. We optimize all trainable parameters with AdamW using a cosine
learning-rate schedule with $100$ warmup steps, a peak learning rate of
$1\times10^{-4}$, and decay to $1\times10^{-6}$ over $2{,}000$ steps. The
released launcher uses a global batch size of $256$, $32$ data-loading workers,
\texttt{fsdp\_devices}$=1$, and a total of $2{,}000$ training steps. EMA is
disabled.

\paragraph{GR00T N1.5.}
We fine-tune the pretrained GR00T N1.5 generalist policy on the demonstrations
of each evaluation task using full-parameter fine-tuning of the backbone,
without LoRA adapters. The policy receives four temporally synchronized RGB
camera views, the embodiment proprioceptive state, and the task instruction as a
per-task language prompt. To accommodate the heterogeneous bimanual-dexterous
embodiments, we pad the state vector to a common $64$-dimensional representation
and extend the pretrained action head from its native maximum dimensionality to
$64$ dimensions; action channels beyond the pretrained head are newly
initialized and optimized jointly with the pretrained parameters, while padded
action dimensions are masked from the action objective, preserving a single
policy interface across embodiments. GR00T N1.5 uses a flow-matching action head
that, from a single observation ($T_o=1$), predicts a
$T_{\mathrm{pred}}=16$-step action chunk; at inference the chunk is generated
with $4$ flow-matching denoising steps and executed in full
($T_{\mathrm{exec}}=16$) before the policy re-observes and re-plans. We optimize
with AdamW at a peak learning rate of $1\times10^{-4}$ with $5\%$ linear warmup
followed by cosine decay and a global batch size of $64$, training in bfloat16
mixed precision for $20{,}000$ steps on a single H100 GPU.

\section{Generalization Configuration}
\label{app:generalization_config}

\paragraph{Scene background.}
Scene background randomization changes the visual surroundings while leaving the tabletop task unchanged. The benchmark includes 60 iTHOR/USD indoor scenes, partitioned into 50 seen and 10 unseen scenes. Each scene has a calibrated yaw and translation offset, and, when this factor is resampled, each episode samples exactly one scene from its designated partition; the nominal background is not substituted. The geometry of the sampled background scene is used only for visual rendering and is excluded from collision and contact simulation. This factor therefore evaluates object grounding under unseen scene backgrounds without introducing room-level physical interactions.

\paragraph{Tabletop texture.}
Tabletop texture randomization changes the visual material of the support surface while preserving its geometry and contact behavior. The texture collection contains 11,824 materials across carpet, fabric, flooring, leather, metal, rust, stone, and wood, with 9,436 seen and 2,388 unseen materials assigned by a deterministic 80/20 split. When this factor is resampled, each episode receives a sampled texture, preventing a fixed tabletop texture from becoming a localization shortcut.

\paragraph{Lighting conditions.}
Lighting randomization modifies the lights embedded in the sampled USD room. The distant light samples intensity in $[1000,1500]$, each color channel in $[0.4,1.0]$, pitch in $[-30^\circ,-5^\circ]$, yaw in $[-90^\circ,90^\circ]$, and angular size in $[0.5^\circ,1.0^\circ]$. The dome light samples intensity in $[150,350]$, color between $[0.8,0.8,0.6]$ and $[1,1,1]$, and exposure in $[-0.5,0.5]$. These ranges are shared across splits and probe sensitivity to illumination, shadows, contrast, and exposure.

\paragraph{Object pose.}
Object pose randomization changes task-relevant initial poses within valid bounds. For explicitly positioned objects, $x$ and $y$ are independently perturbed by up to $\pm2$~cm and yaw by up to $\pm10^\circ$. If a sampled pose violates collision or workspace constraints, the perturbation magnitude is progressively reduced; the nominal pose is retained only when no valid perturbed pose is found. Objects without a fixed pose are instead resampled within task-defined zones subject to collision constraints. Task-specific geometric constraints may fix an object's initial pose or restrict its perturbation range.

\paragraph{Camera pose.}
Camera pose randomization perturbs external and wrist-mounted camera viewpoints. World-mounted cameras sample translation and look-at-target offsets of $\pm5$~mm per axis, distance offsets of $\pm3$~cm, and roll--pitch--yaw offsets of $\pm1^\circ$ per axis. Wrist cameras sample link-frame translation offsets of $\pm5$~mm per axis and rotation offsets of $\pm1^\circ$ per axis. Stereo cameras share one perturbation sample to preserve their relative pose. The seen and unseen camera profiles use the same numerical ranges, so this factor evaluates calibration and viewpoint perturbations rather than a disjoint camera-hardware split.

\paragraph{Distractor objects.}
Distractor object randomization adds one to three dynamic distractors per episode. The distractor set contains 68 assets, partitioned into 52 seen and 16 unseen assets; each selected object is scaled by a factor in $[0.8,1.2]$. Objects that duplicate a task-relevant object or are semantically confusable with one are excluded before sampling. Placement satisfies collision constraints, a 5~cm table-edge margin, and a central exclusion region $x\in[-0.25,0.25]$~m and $y\in[-0.20,0.35]$~m. This procedure evaluates grounding under occlusion, ambiguity, crowding, and incidental contact without deliberately blocking the primary workspace.

\paragraph{Table height.}
Table height randomization samples a vertical offset in $[-0.05,0.05]$~m around each task's nominal tabletop height (default $0.75$~m). The robot mount remains at its nominal height, while on-table object placement, evaluator height references, and dependent collision checks follow the shifted surface. The resulting change in robot-to-table geometry tests adaptation of reaching, grasping, and contact heights.

\begin{table*}[t]
\caption{Composition and perturbation strengths of the four Bench2Dex evaluation channels. Entries prefixed by \emph{From anchor} replay the resolved parameters of the matched anchor episode. Task-level object overrides may narrow or disable pose perturbations.}
\label{tab:bench2dex_generalization_channels}
\centering
\scriptsize
\setlength{\tabcolsep}{3pt}
\renewcommand{\arraystretch}{1.16}
\begin{tabularx}{\textwidth}{>{\raggedright\arraybackslash}p{0.08\textwidth} >{\raggedright\arraybackslash}p{0.15\textwidth} >{\raggedright\arraybackslash}X >{\raggedright\arraybackslash}X >{\raggedright\arraybackslash}p{0.15\textwidth}}
\toprule
Channel & Scene construction & Invariance factors: scene background, tabletop texture, lighting conditions, distractor objects, camera pose & Equivariance factors: object pose, table height & Evaluation purpose \\
\midrule
\texttt{none} & Matched anchor; all factors are replayed exactly. & \emph{From anchor}: scene background, tabletop texture, lighting conditions, distractor objects, and camera pose. & \emph{From anchor}: object poses and table height. & Establishes the matched baseline without resampling. \\
\texttt{equi\_only} & Matched anchor; only equivariance factors are resampled. & \emph{From anchor}: scene background, tabletop texture, lighting conditions, distractor objects, and camera pose. & \emph{Resampled}: explicit object $x$--$y$: $\pm2$~cm; yaw: $\pm10^\circ$; objects without fixed pose: task zones; table: $\pm5$~cm. & Tests adaptation to task-relevant geometry while controlling visual context. \\
\texttt{inv\_only} & Matched anchor; only invariance factors are resampled from the unseen split. & \emph{Resampled from unseen split}: scene background: 10 iTHOR scenes; tabletop texture: 2,388 materials; distant/dome lighting: ranges above; distractor objects: 1--3 objects from 16 assets at $0.8$--$1.2\times$ scale; world-mounted camera translation/look-at-target offsets: $\pm5$~mm, distance: $\pm3$~cm, RPY: $\pm1^\circ$; wrist-mounted camera translation: $\pm5$~mm, RPY: $\pm1^\circ$. & \emph{From anchor}: object poses and table height. & Tests nuisance robustness under matched task geometry. \\
\texttt{inv\_equi} & Independent full-scene sample; no anchor is used. & \emph{Resampled from unseen split}: the same perturbations to scene background, tabletop texture, lighting conditions, distractor objects, and camera pose as \texttt{inv\_only}. & \emph{Resampled}: the same perturbations to object pose and table height as \texttt{equi\_only}. & Tests simultaneous perceptual invariance and geometric adaptation. \\
\bottomrule
\end{tabularx}
\end{table*}

\clearpage
\section{Collected Tactile Observations}
\label{app:collected_tactile_observations}

The following figures present representative tactile observations collected across 12 robot--hand embodiments. Each image is shown individually and without cropping. Each figure visualizes the tactile maps of one replayed frame, with one heatmap per tactile site; brighter pixels encode larger quantized penetration depth (values $0$--$255$, Appendix~\ref{app:tacmap_quantization}).

\begin{figure*}[htbp]
    \centering
    \includegraphics[width=0.6\textwidth]{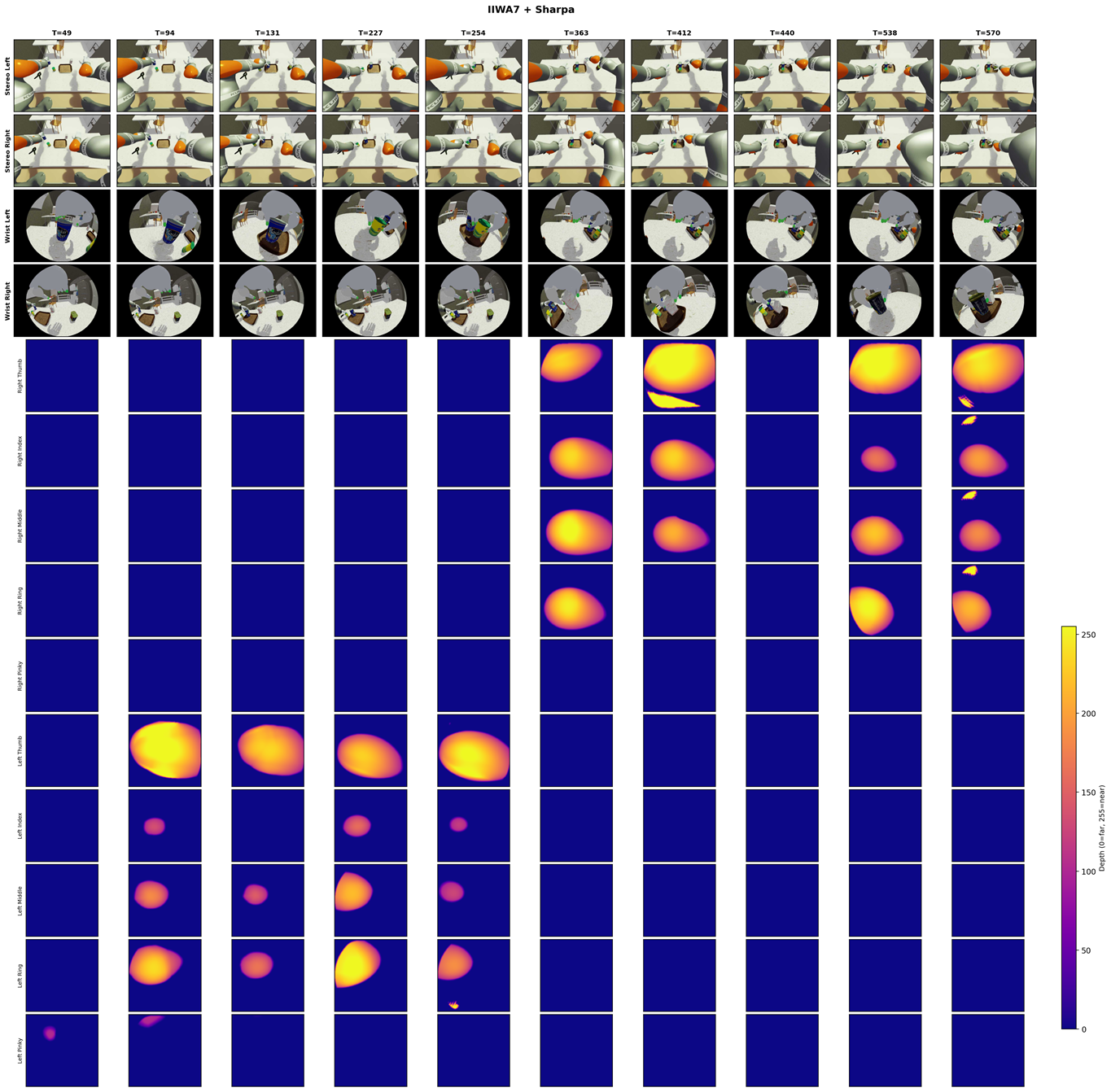}
    \caption{Visualization of tactile data collected with the IIWA7+Sharpa embodiment.}
    \label{fig:tactile_iiwa7_sharpa_26_000032}
\end{figure*}

\begin{figure*}[htbp]
    \centering
    \includegraphics[width=0.6\textwidth]{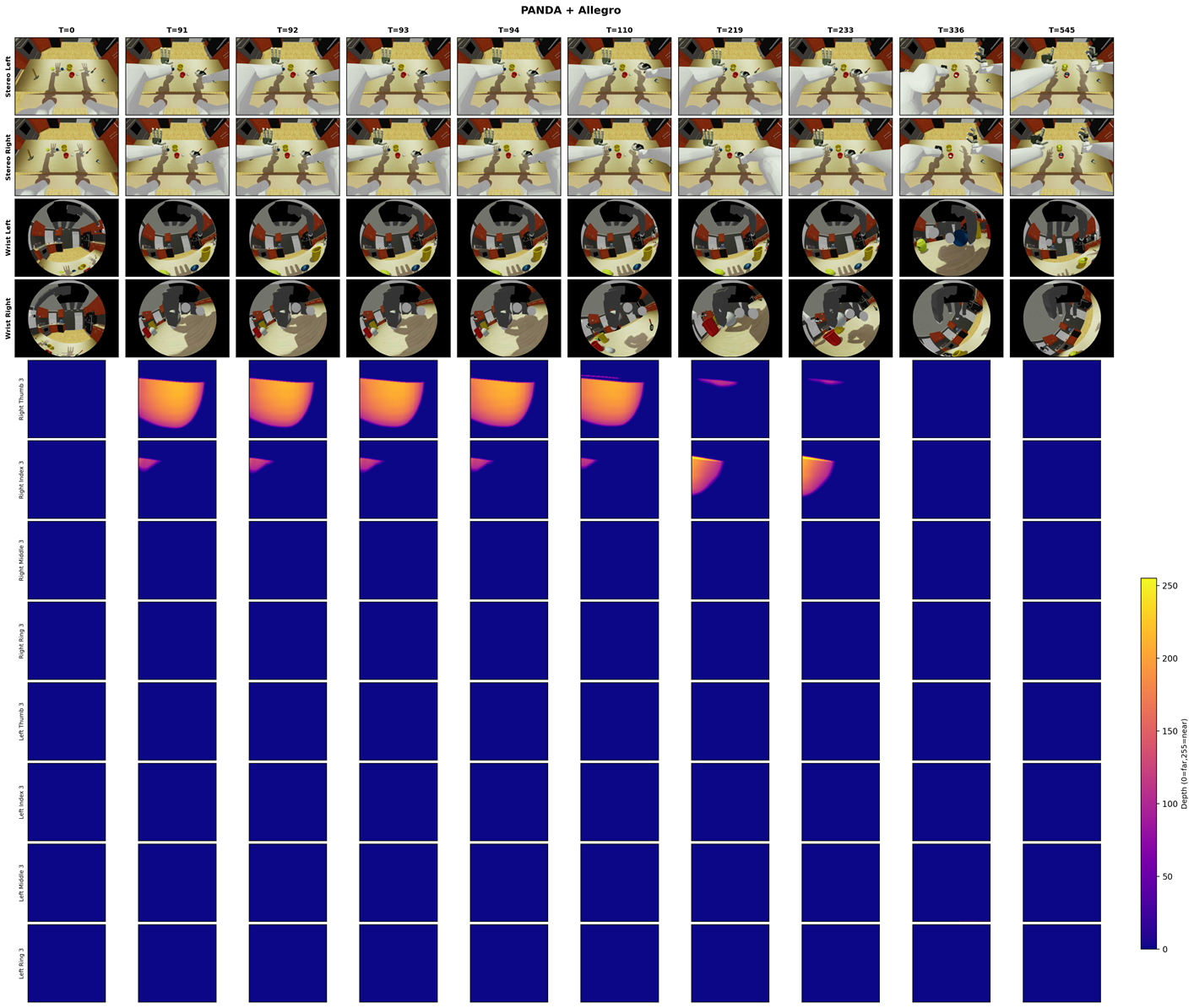}
    \caption{Visualization of tactile data collected with the Panda+Allegro embodiment.}
    \label{fig:tactile_panda_allegro_64_000002}
\end{figure*}

\begin{figure*}[htbp]
    \centering
    \includegraphics[width=0.6\textwidth]{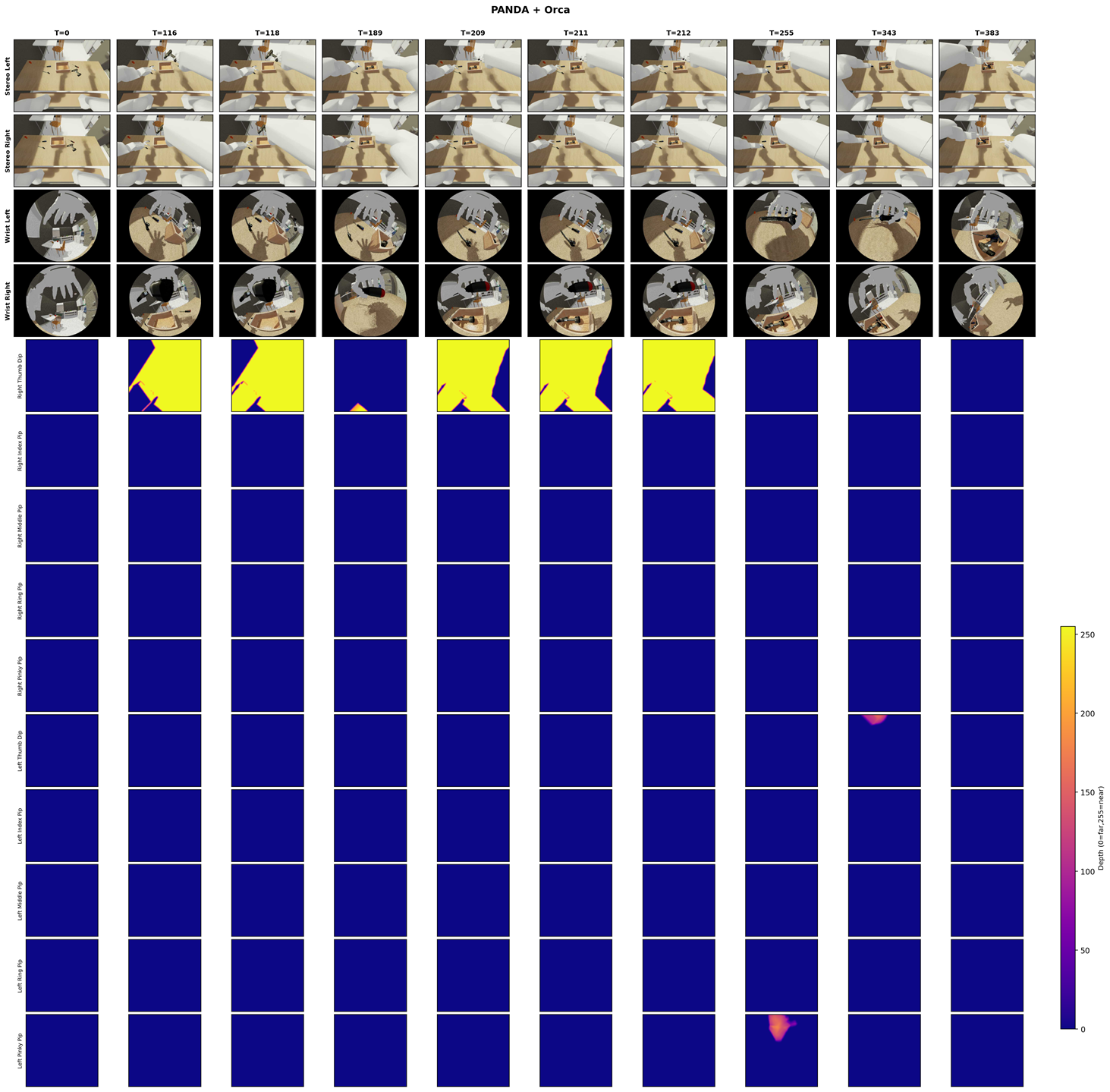}
    \caption{Visualization of tactile data collected with the Panda+Orca embodiment.}
    \label{fig:tactile_panda_orca_22_000007}
\end{figure*}

\begin{figure*}[htbp]
    \centering
    \includegraphics[width=0.6\textwidth]{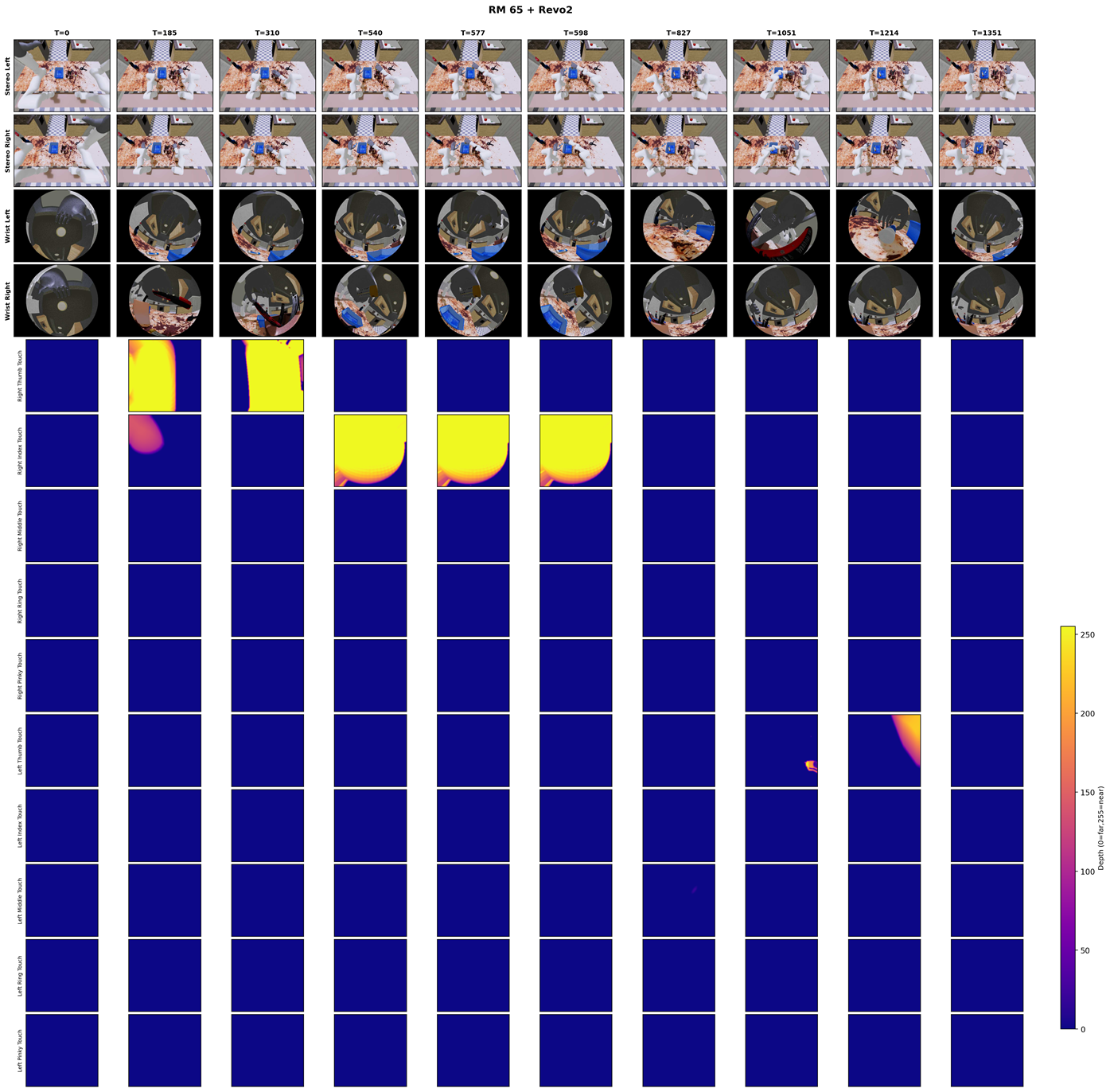}
    \caption{Visualization of tactile data collected with the RM65+Revo2 embodiment.}
    \label{fig:tactile_rm_65_revo2_24_000003}
\end{figure*}

\begin{figure*}[htbp]
    \centering
    \includegraphics[width=0.6\textwidth]{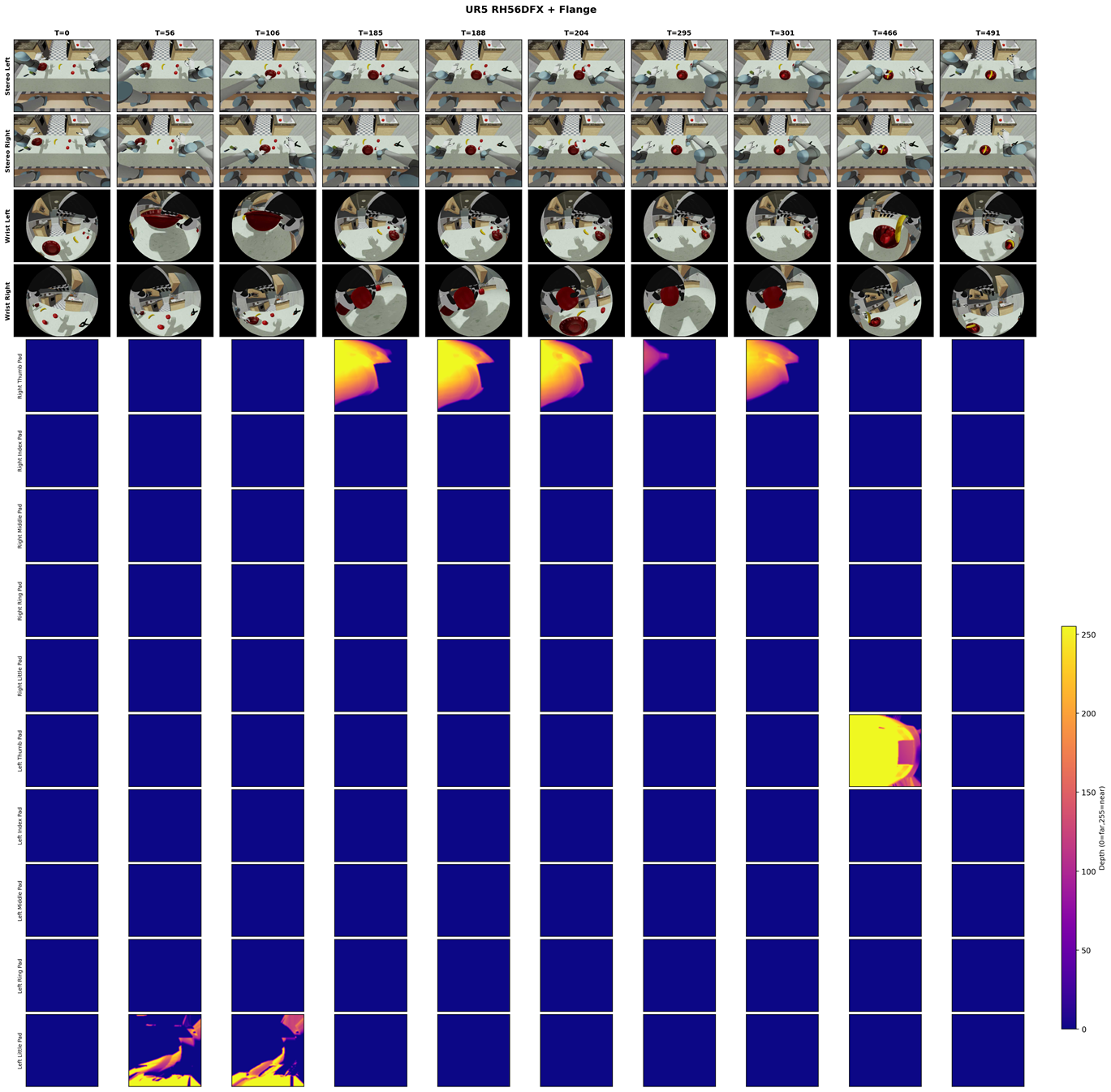}
    \caption{Visualization of tactile data collected with the UR5+RH56DFX embodiment.}
    \label{fig:tactile_ur5_rh56dfx_flange_06_000003}
\end{figure*}

\begin{figure*}[htbp]
    \centering
    \includegraphics[width=0.6\textwidth]{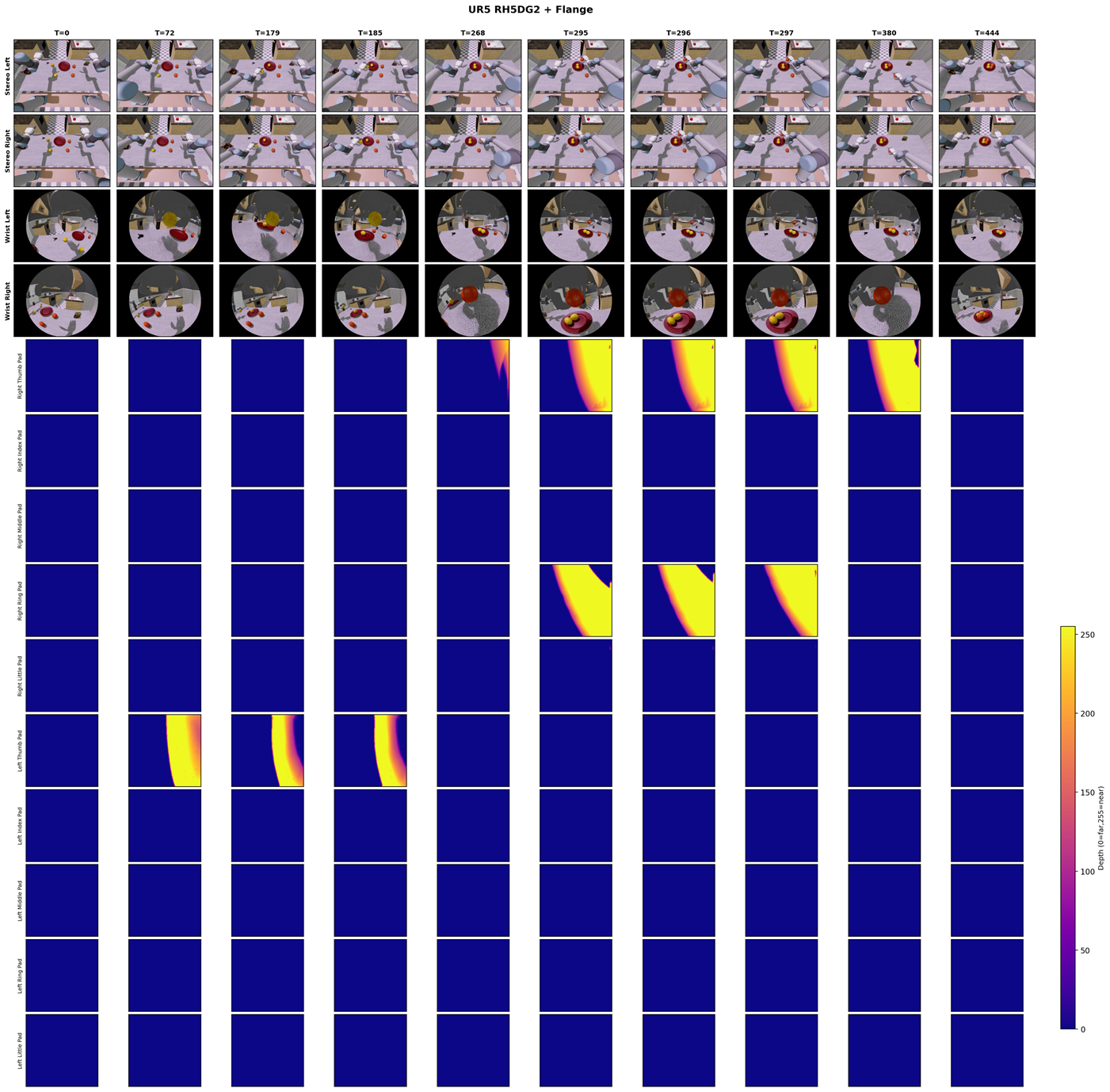}
    \caption{Visualization of tactile data collected with the UR5+RH5DG2 embodiment.}
    \label{fig:tactile_ur5_rh5dg2_flange_07_000003}
\end{figure*}

\begin{figure*}[htbp]
    \centering
    \includegraphics[width=0.6\textwidth]{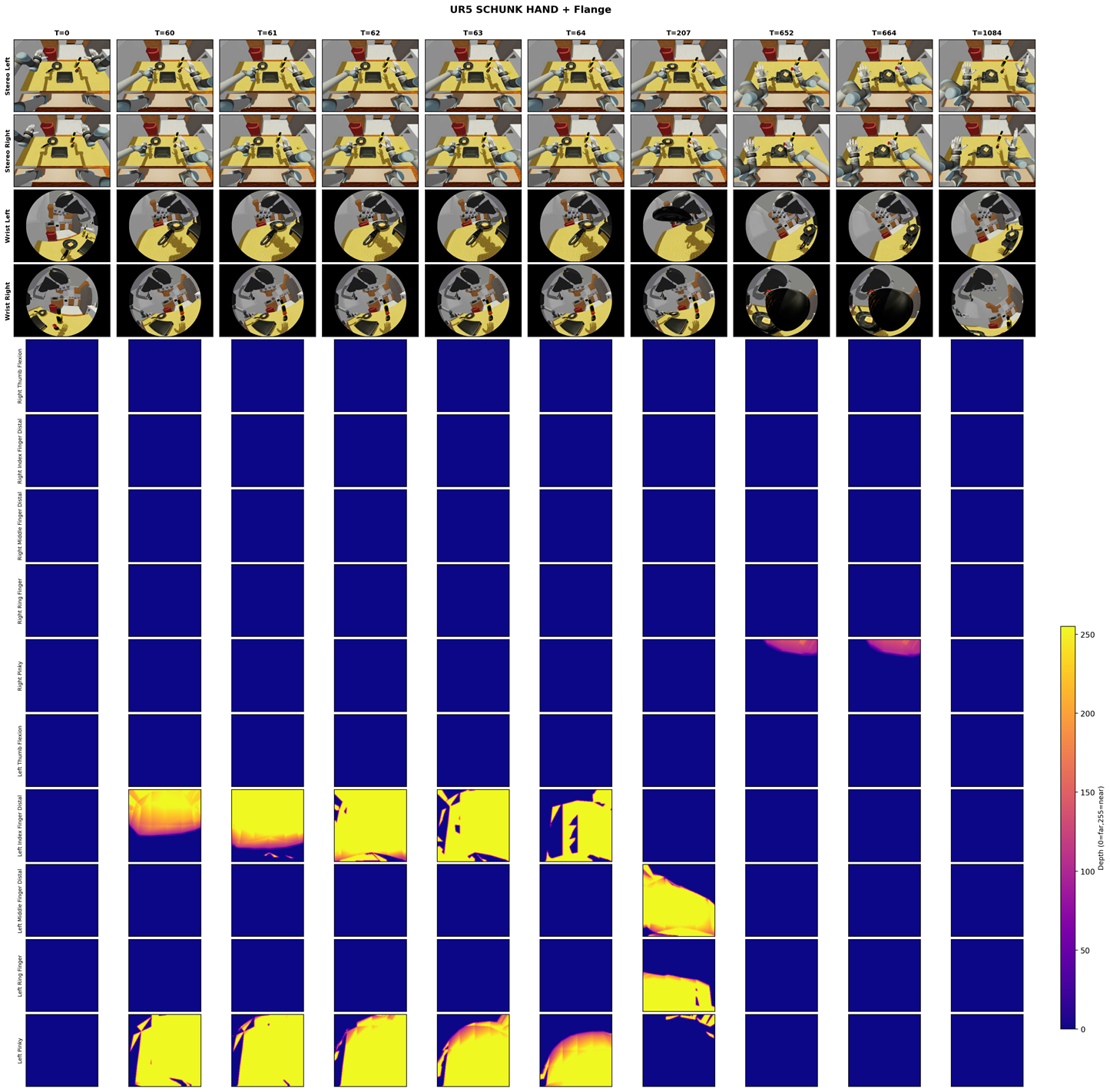}
    \caption{Visualization of tactile data collected with the UR5+Schunk embodiment.}
    \label{fig:tactile_ur5_schunk_hand_flange_08_000006}
\end{figure*}

\begin{figure*}[htbp]
    \centering
    \includegraphics[width=0.6\textwidth]{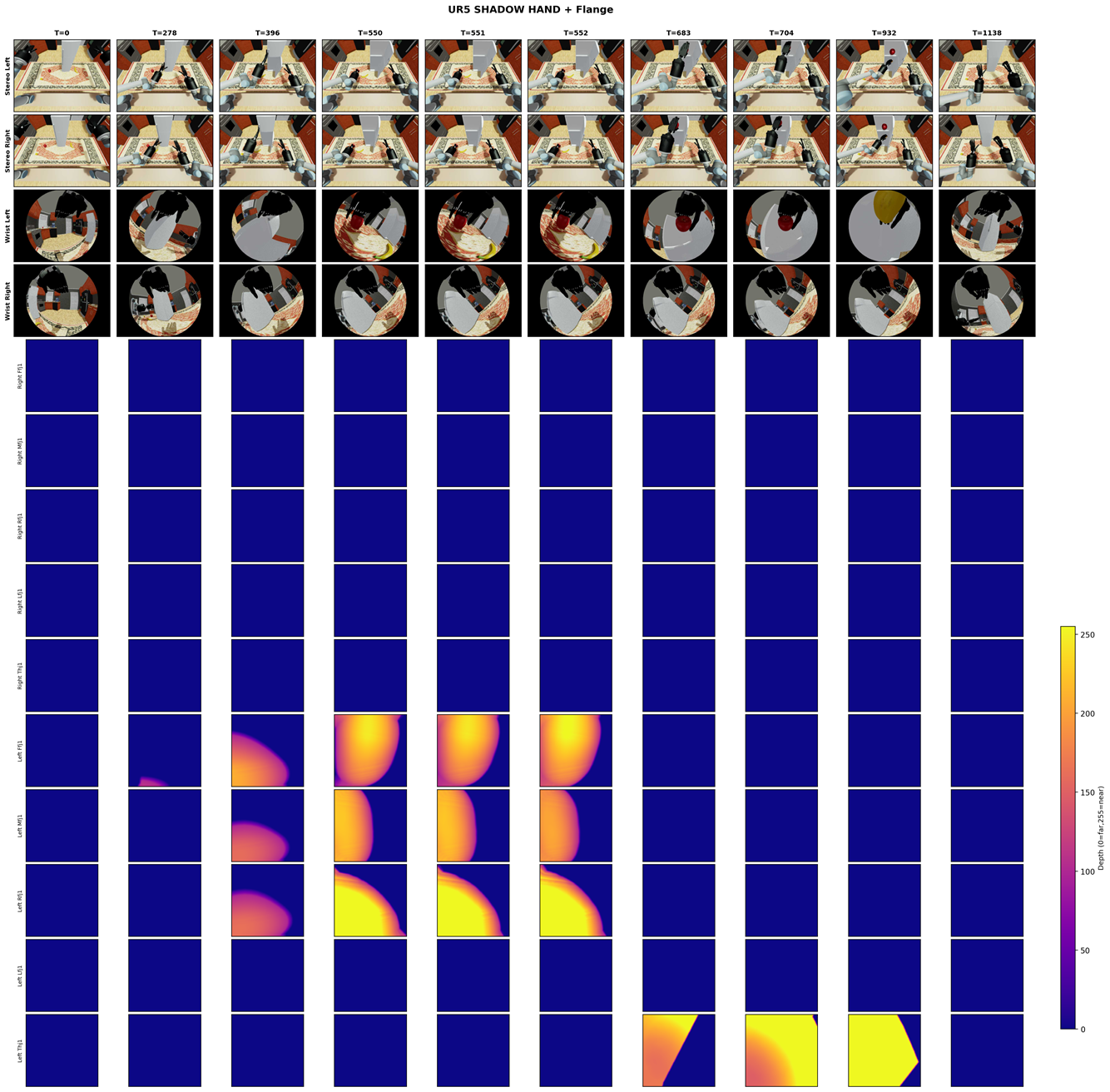}
    \caption{Visualization of tactile data collected with the UR5+Shadow embodiment.}
    \label{fig:tactile_ur5_shadow_hand_flange_43_000002}
\end{figure*}

\begin{figure*}[htbp]
    \centering
    \includegraphics[width=0.6\textwidth]{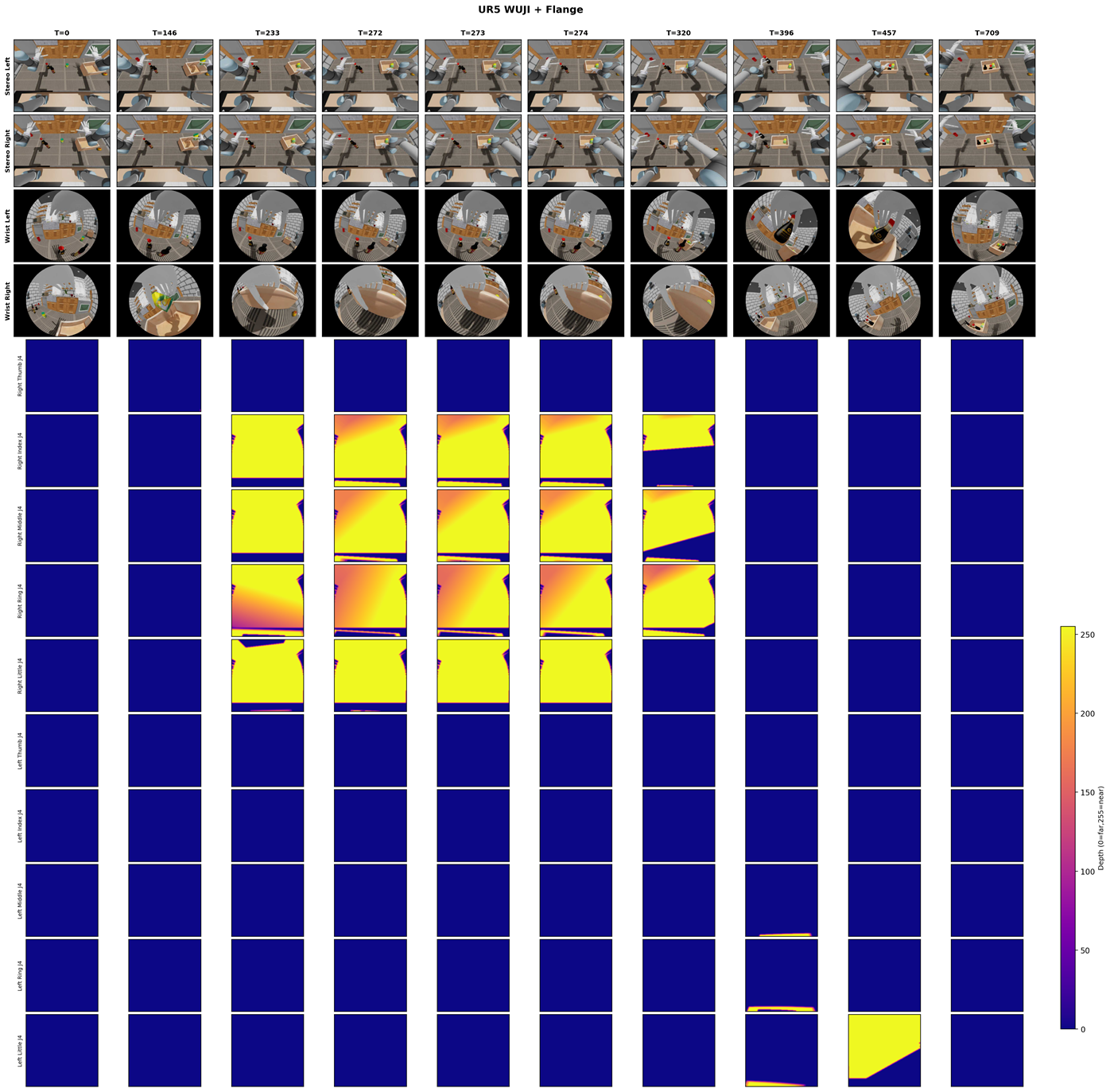}
    \caption{Visualization of tactile data collected with the UR5+Wuji embodiment.}
    \label{fig:tactile_ur5_wuji_flange_21_000001}
\end{figure*}

\begin{figure*}[htbp]
    \centering
    \includegraphics[width=0.6\textwidth]{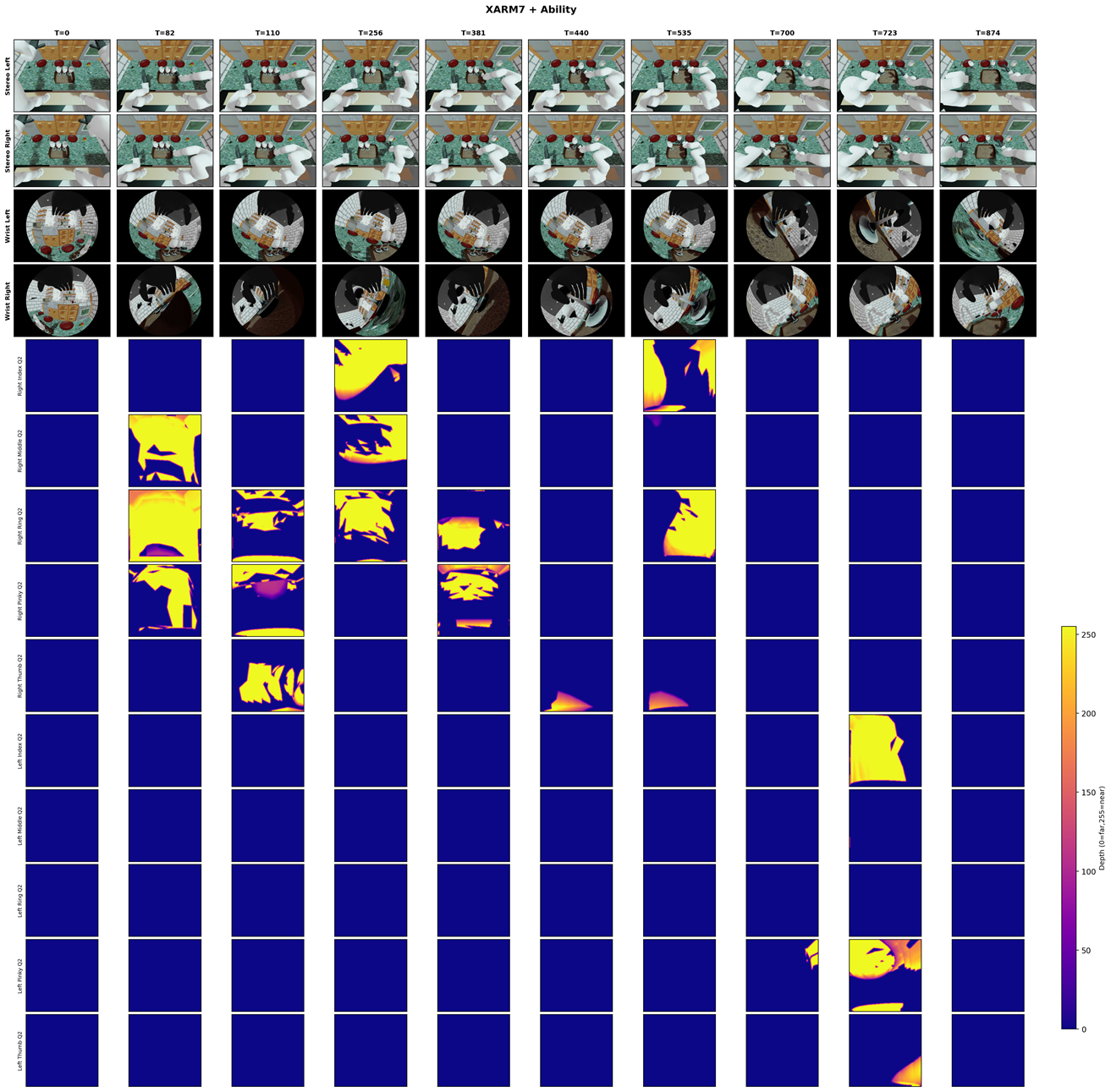}
    \caption{Visualization of tactile data collected with the xArm7+Ability embodiment.}
    \label{fig:tactile_xarm7_ability_03_000001}
\end{figure*}

\begin{figure*}[htbp]
    \centering
    \includegraphics[width=0.6\textwidth]{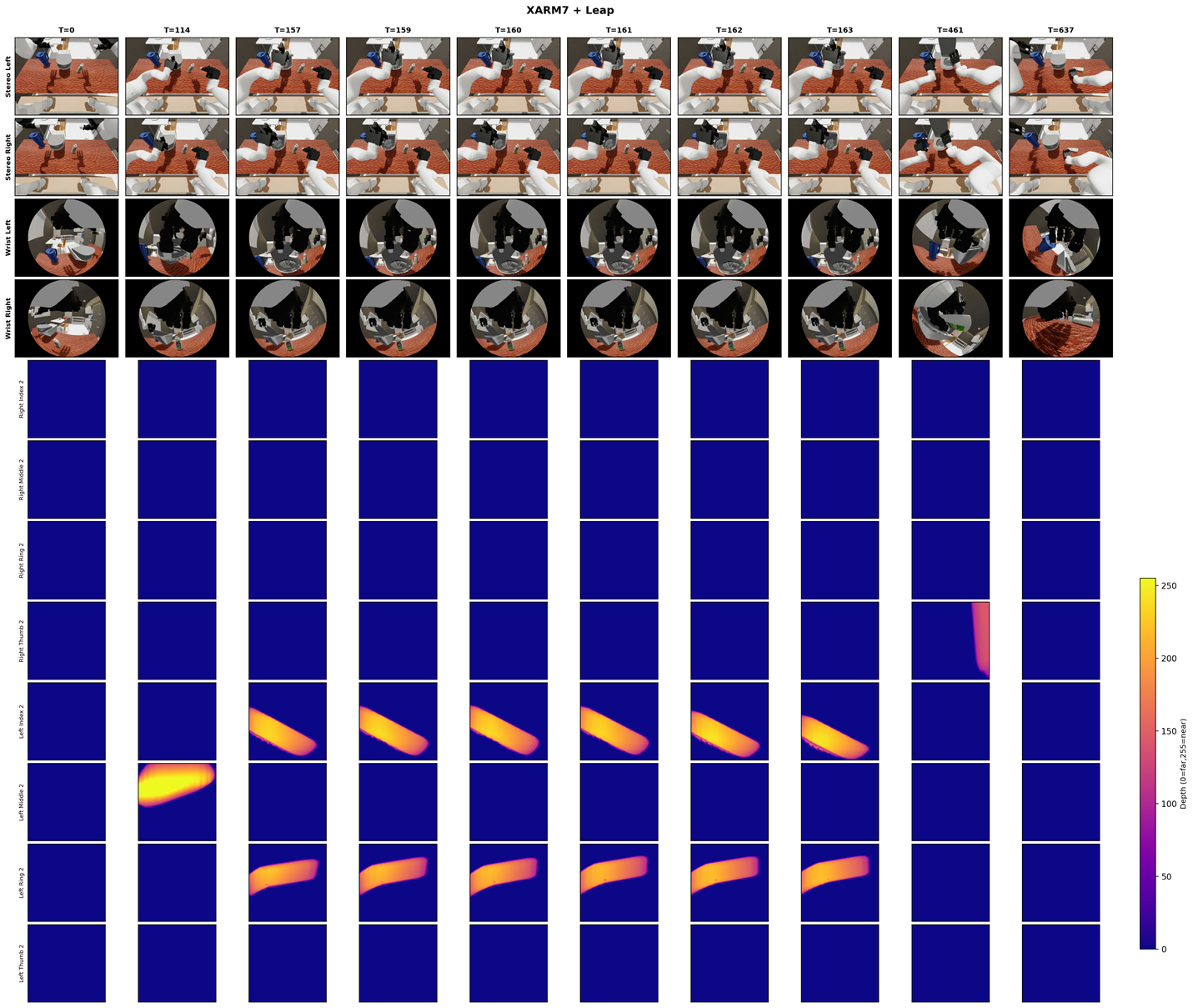}
    \caption{Visualization of tactile data collected with the xArm7+LEAP embodiment.}
    \label{fig:tactile_xarm7_leap_51_000007}
\end{figure*}

\begin{figure*}[htbp]
    \centering
    \includegraphics[width=0.6\textwidth]{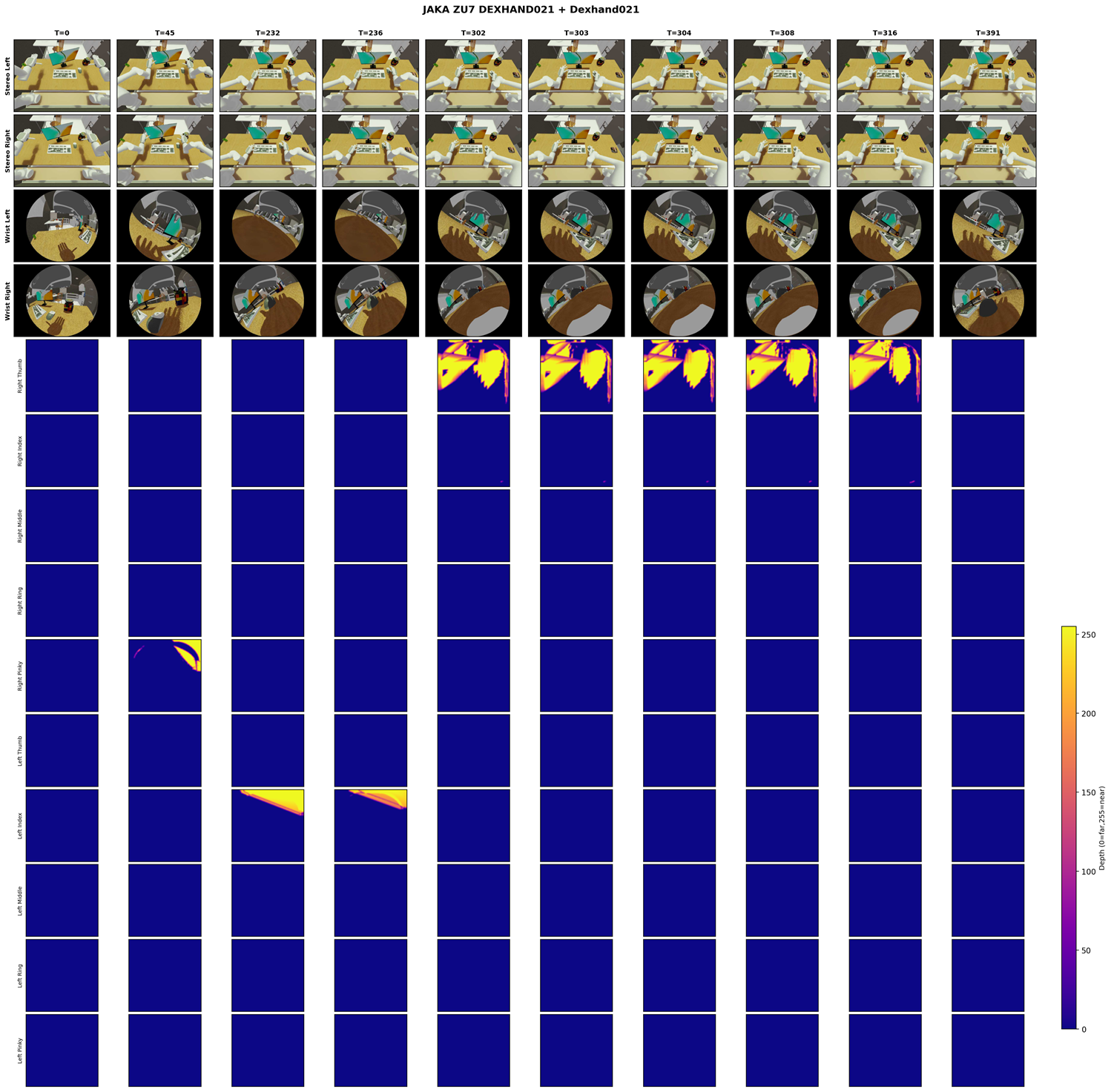}
    \caption{Visualization of tactile data collected with the JAKA ZU7+DexHand021 embodiment.}
\label{fig:tactile_jaka_zu7_dexhand021_flange_80_000007}
\end{figure*}

\clearpage

% \section{Background Assets and Desktop Texture Assets}
% \label{app:background_desktop_assets}

% To support appearance-level domain randomization, we build a large asset pool for both scene backgrounds and desktop surface textures. The background pool contains approximately 3K HDRI panoramas sourced from open HDRI libraries and 57 USDA-format indoor scene assets, covering diverse illumination conditions, room layouts, and environmental appearances. The desktop texture pool contains approximately 13K image-based surface materials spanning wood, fabric, leather, metal, marble, stone, and composite finishes, collected from open texture repositories. During data collection and evaluation, backgrounds and textures are sampled according to the generalization configuration described in Appendix~\ref{app:generalization_config}, enabling Bench2Dex to generate visually diverse manipulation scenes and to evaluate policy robustness under appearance shifts.

% \begin{figure}[htbp]
%     \centering
%     \includegraphics[width=1\linewidth]{figures/scene_texture.png}
%     \caption{Partial display of background assets and desktop texture assets used in Bench2Dex.}
%     \label{fig:background_desktop_assets}
% \end{figure}

\end{document}